\documentclass{article}

\PassOptionsToPackage{numbers,sort&compress}{natbib}
\usepackage[preprint]{neurips_2026}

\usepackage[utf8]{inputenc}
\usepackage[T1]{fontenc}
\usepackage[colorlinks=true,linkcolor=black,citecolor=black,urlcolor=black,hypertexnames=false]{hyperref}
\usepackage{url}
\usepackage{amsmath,amssymb,mathtools}
\usepackage{amsfonts}
\usepackage{graphicx}
\usepackage{booktabs}
\usepackage{array}
\usepackage{tabularx}
\usepackage{multirow}
\usepackage{makecell}
\usepackage{nicefrac}
\usepackage{microtype}
\usepackage{xcolor}
\usepackage{wrapfig}
\usepackage{xspace}
\usepackage{placeins}
\usepackage{enumitem}
\usepackage{algorithm}
\usepackage{algorithmic}

\graphicspath{{./}}

\definecolor{revisionblue}{RGB}{0,0,0}
\definecolor{orbitgray}{RGB}{245,246,248}
\definecolor{orbitdark}{RGB}{0,0,0}
\definecolor{deeporange}{RGB}{0,0,0}
\definecolor{taskprotocol}{RGB}{0,0,0}
\definecolor{taskcoverage}{RGB}{0,0,0}
\definecolor{taskquality}{RGB}{0,0,0}
\definecolor{tasksupport}{RGB}{0,0,0}
\definecolor{taskpareto}{RGB}{0,0,0}
\definecolor{taskfinal}{RGB}{0,0,0}
\definecolor{taskorbit}{RGB}{0,0,0}
\definecolor{taskview}{RGB}{0,0,0}
\definecolor{taskrelwork}{RGB}{0,0,0}
\definecolor{taskimpl}{RGB}{0,0,0}
\definecolor{taskmedian}{RGB}{0,0,0}
\definecolor{taskdnear}{RGB}{0,0,0}
\definecolor{taskidentity}{RGB}{0,0,0}
\definecolor{taskrobust}{RGB}{0,0,0}
\definecolor{taskrotabl}{RGB}{0,0,0}
\definecolor{taskfinalpolish}{RGB}{0,0,0}
\definecolor{taskboundary}{RGB}{0,0,0}
\definecolor{taskanchor}{RGB}{0,0,0}
\definecolor{taskfitstats}{RGB}{0,0,0}
\definecolor{taskconst}{RGB}{0,0,0}
\definecolor{taskclaim}{RGB}{0,0,0}
\definecolor{taskappcite}{RGB}{0,0,0}
\definecolor{taskqualapp}{RGB}{0,0,0}
\definecolor{taskmedapp}{RGB}{0,0,0}
\definecolor{taskcompapp}{RGB}{0,0,0}
\definecolor{taskcleanapp}{RGB}{0,0,0}

\newcommand{\ours}{OrbitForge\xspace}
\newcommand{\Rzero}{R_0}
\newcommand{\Rone}{R_1}
\newcommand{\tthreebench}{T\textsuperscript{3}Bench\xspace}
\newcommand{\Gzero}{G_0}
\newcommand{\Gone}{G_1}
\newcommand{\Gmed}{G_{\mathrm{med}}}
\newcommand{\Vzero}{V_0}
\newcommand{\Vhat}{\widehat{V}_0}
\newcommand{\orbit}{\mathcal{C}}
\newcommand{\renderer}{\mathcal{P}}
\newcommand{\recon}{\mathcal{R}}
\newcommand{\medianop}{\mathcal{M}}
\newcommand{\completion}{\mathcal{K}}

\newcommand{\taskorbit}[1]{#1}
\newcommand{\taskview}[1]{#1}
\newcommand{\taskrelwork}[1]{#1}
\newcommand{\taskimpl}[1]{#1}
\newcommand{\taskmedian}[1]{#1}
\newcommand{\taskdnear}[1]{#1}

\newcommand{\taskrobust}[1]{#1}
\newcommand{\taskrotabl}[1]{#1}

\newcommand{\taskboundary}[1]{#1}
\newcommand{\taskanchor}[1]{#1}
\newcommand{\taskfitstats}[1]{#1}
\newcommand{\taskconst}[1]{#1}
\newcommand{\taskclaim}[1]{#1}
\newcommand{\taskappcite}[1]{#1}
\newcommand{\taskqualapp}[1]{#1}
\newcommand{\taskmedapp}[1]{#1}
\newcommand{\taskcompapp}[1]{#1}
\newcommand{\taskcleanapp}[1]{#1}

\title{\ours: Text-to-3D Scene Generation via Reconstruction-Anchored Video Synthesis}

\author{Chenrui Fan \quad Paolo Favaro\\
Computer Vision Group, Institute of Computer Science\\
University of Bern, Switzerland\\
\texttt{\{chenrui.fan,paolo.favaro\}@unibe.ch}}
\date{}

\begin{document}

\maketitle

\begin{abstract}
Generic text-to-video models can be used as rich open-world scene priors. Despite the high quality of today's generated videos, they do not directly yield reliable 3D assets: camera motion is difficult to control, view coverage is partial, and frames often contain inconsistencies across time. We introduce \ours, an adapter built from frozen video priors and per-prompt Gaussian Splatting reconstruction optimization that converts a single text-generated video into a canonical closed-orbit 3D Gaussian Splatting scene. We use 3D reconstruction as an anchor to improve the 3D consistency of the generated video. We obtain a preliminary 3D reconstruction from a first generated video via Deformable Gaussian Splatting with a robust MedianGS proxy. We render views from a prescribed orbit to detect missing viewpoints. \ours uses the text-to-video model to complete only the missing views, and reconstructs the completed orbit into a final Gaussian Splatting scene. This design requires no task-specific video or multiview fine-tuning, avoids per-prompt score-distillation optimization, and does not progressively generate views one step at a time. We further argue that this setting demands coverage-aware evaluation: local smoothness alone rewards methods that never attempt a full orbit. \taskclaim{On a frozen 300-prompt \tthreebench{}-derived audit, \ours{} reconstruction attains a 359.0-degree measured median span, raises originally unsupported-bin Q10 ImageReward from 8.07 to 16.36 relative to MedianGS-only reconstruction, while remaining competitive with VideoMV on the coverage-quality.}
\end{abstract}

\section{Introduction}

Text-driven 3D generation has made remarkable progress by borrowing the visual knowledge of large 2D and video diffusion models. Yet there remains a sharp gap between a beautifully generated image or video and a usable 3D scene. \taskclaim{A 3D scene is not merely a collection of plausible views: it must live in a coordinate system, remain renderable from requested cameras, keep the depicted object and scene recognizable around the orbit, and expose the back, sides, ground, and surrounding environment rather than only the most photogenic front view.}

Existing approaches address this challenge from one of three directions. Score-distillation methods optimize a 3D representation per prompt using a frozen image prior, often producing renderable object-centric assets but requiring long optimization and suffering from over-saturation, Janus artifacts, or weak scene context \citep{poole2022dreamfusion,wang2023score,lin2023magic3d,wang2023prolificdreamer}. Specialized multiview generators learn a camera-conditioned prior and produce a fixed set of views that is convenient for reconstruction, but the camera format, elevation, and object-centric assumptions are usually built into the model \citep{shi2023mvdream,voleti2024sv3d,zuo2024videomv,li2025coser}. Local novel-view methods can synthesize high-quality short trajectories from one or a few images, but their strength is local view completion, not necessarily closed 360-degree scene generation from an unconstrained text-generated video \citep{gao2024cat3d,yu2024viewcrafter,liu2026reconx}.

\begin{figure}[t]
\centering
\includegraphics[width=\linewidth]{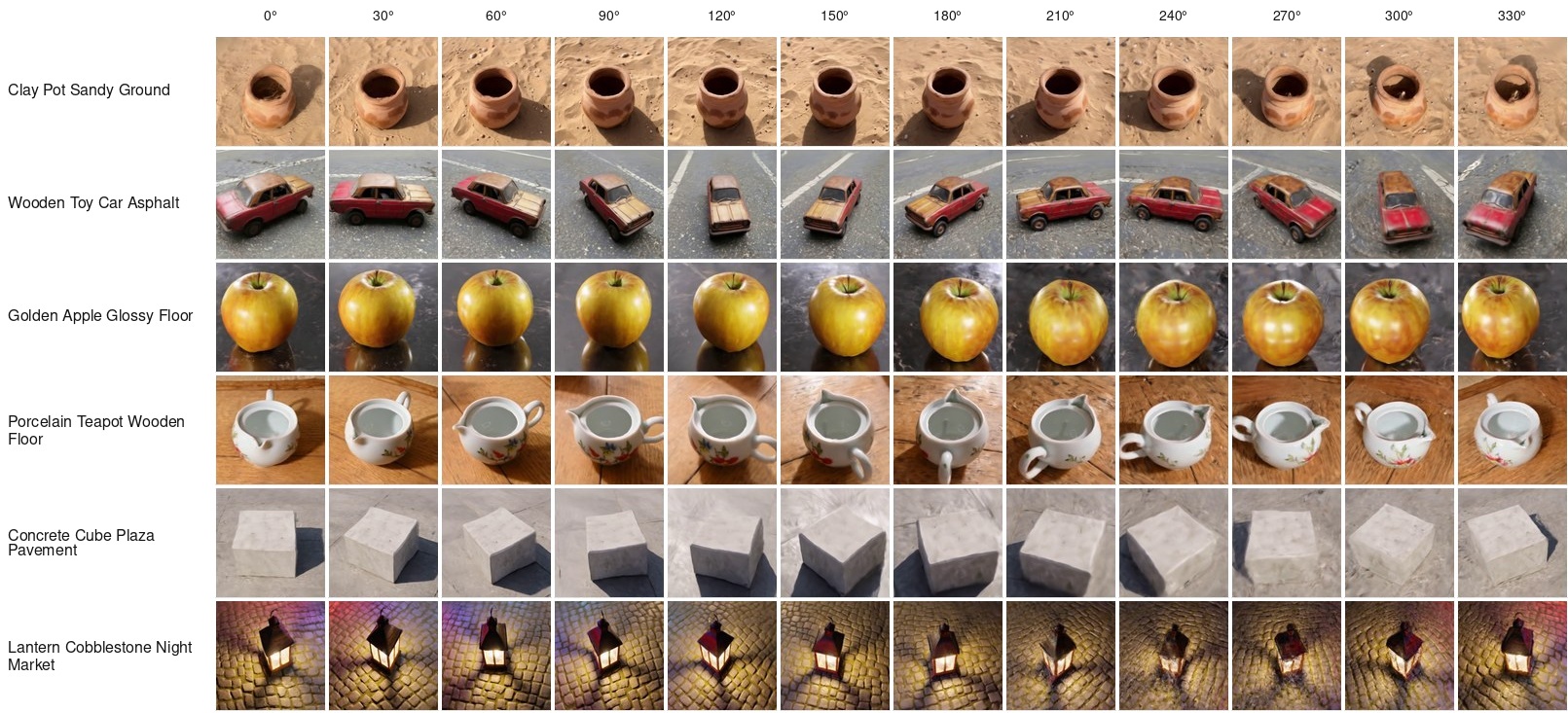}
\caption{Example outputs from \ours across six prompts. Each row renders the final Gaussian Splatting scene on the same canonical orbit at 30-degree intervals, showing complete orbit coverage while preserving object identity, ground contact, and surrounding scene context.}
\label{fig:orbitforge-effect-main}
\end{figure}

We study a complementary question: \emph{can a frozen generic text-to-video model be used as a closed-orbit 3D scene generator without training a new multiview prior or running per-prompt Score Distillation Sampling (SDS) optimization?} Generic text-to-video models are attractive because they already produce scene-level realism: objects rest on floors, reflect light, interact with backgrounds, and inherit open-world semantic detail from language prompts. However, their videos are not calibrated multiview captures. Camera motion is implicit and often covers only a partial arc. Object shape can deform, textures can be repainted, and backgrounds can drift. Directly fitting 3D Gaussians to such frames turns these contradictions into floaters, blurred geometry, stretched surfaces, and view-dependent artifacts.

The key observation behind \ours is that a failed first reconstruction is still useful. Even when it is imperfect, it has performed a crucial act: it has organized the generated frames into a shared 3D coordinate system. Once this preliminary reconstruction is rendered along a prescribed canonical orbit, the problem becomes structured. Some views are supported by the source trajectory; other views are visibly unsupported. The method can then complete the unsupported orbit interval by querying the text-to-video model again, while keeping the supported arc fixed, and reconstruct again under known cameras (Figure~\ref{fig:orbitforge-effect-main} shows several outputs generated by \ours).

Therefore, this paper makes a precise claim. \ours is not designed to dominate every local aesthetic metric, and it should not be evaluated as if a narrow view-completion trajectory and a complete scene orbit were the same task. Its target is prompt-controllable, scene-level, closed-orbit 3D generation from generic text-to-video priors without task-specific generative-model training or per-prompt SDS optimization, while still using per-prompt Gaussian Splatting reconstruction optimization. A fair evaluation must measure coverage before rewarding local smoothness.

Our contributions are fourfold. First, we formulate closed-orbit 3D scene generation from generic text-generated videos, a setting that leverages frozen text-to-video priors while explicitly handling uncalibrated cameras, partial coverage, and generated-video inconsistency. Second, we propose \ours, a reconstruction-render-completion adapter that does not train a new generative prior and uses an imperfect first reconstruction as an anchor rather than as a final output. Third, we introduce MedianGS static scene consolidation and coverage-aware endpoint-window orbit completion, which convert a partial generated-video trajectory into a complete canonical-orbit reconstruction signal without progressive view expansion. Fourth, we introduce a coverage-aware evaluation protocol--measured angular span, monotone progress, closed-orbit success, lower-tail quality, and diagnostic appearance consistency--so that methods are compared on the task they actually solve rather than only on adjacent-frame smoothness.

\begin{figure}[t]
\centering
\includegraphics[width=\linewidth,height=0.70\textheight,keepaspectratio]{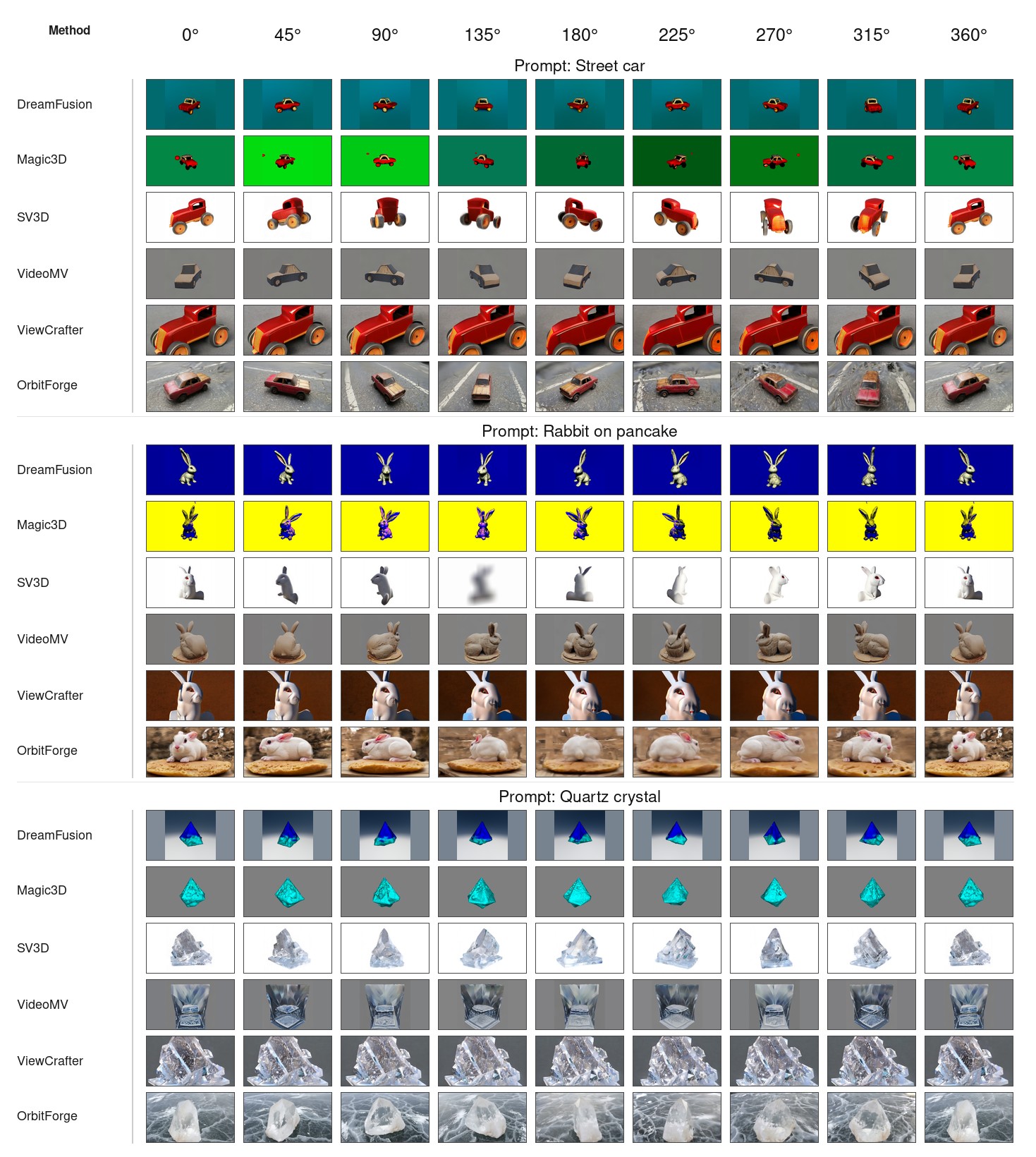}
\caption{Qualitative comparison on three scene prompts and nine uniformly sampled orbit views. SDS-style text-to-3D methods can produce clean isolated turntables, while specialized multiview methods often simplify the surrounding scene. ViewCrafter preserves strong local image quality over a short trajectory, but does not expose a comparable closed orbit. \ours targets a complete scene-level canonical orbit, retaining more of the ground, background, and object-environment relationship across the full view range.}
\label{fig:qualitative-main}
\end{figure}

\section{Positioning and Related Work}

We summarize the positioning of \ours in Table~\ref{tab:positioning}.
\paragraph{Score-distillation text-to-3D.} DreamFusion optimizes a NeRF using a pretrained text-to-image diffusion prior through probability density distillation \citep{poole2022dreamfusion}; SJC back-propagates diffusion scores through a differentiable renderer \citep{wang2023score}; Magic3D improves resolution and efficiency through a coarse-to-fine mesh optimization pipeline \citep{lin2023magic3d}; and later methods such as ProlificDreamer and JSDDreamer revisit the distillation objective to improve diversity and stability \citep{wang2023prolificdreamer,do2025text}. These methods usually keep the diffusion prior frozen, but they still perform per-prompt SDS-style 3D optimization. \ours instead samples a generic video prior and reconstructs from generated observations, avoiding per-prompt SDS optimization while retaining the scene-level appearance of modern video models.

\paragraph{Specialized multiview generation.} MVDream, SV3D, VideoMV, and CoSER show that diffusion and video architectures can be adapted into strong multiview generators \citep{shi2023mvdream,voleti2024sv3d,zuo2024videomv,li2025coser}. Their outputs are valuable for object reconstruction because the camera grid is known and the views are designed to be mutually consistent. The trade-off is specialization: such systems typically assume a fixed multiview format, an object-centered orbit, a canonical elevation, or task-specific multiview training. \ours takes the opposite route. It does not fine-tune the video model into a multiview generator; it uses a generic text-to-video model as an open-world scene prior, then constructs the camera system after generation.

\paragraph{\taskrelwork{Video priors as 3D or orbit generators.}} \taskrelwork{A growing line of work uses video diffusion more directly for 3D or camera-orbit generation. ViVid-1-to-3 reformulates novel-view synthesis as generating a scanning video toward a target view \citep{kwak2023vivid}; V3D fine-tunes a video diffusion model to generate 360-degree object-orbit frames from a single image before reconstruction \citep{chen2024v3d}; Generative Gaussian Splatting integrates a Gaussian Splatting feature representation inside a latent video diffusion model for 3D scene generation \citep{schwarz2025ggs}; and G4Splat uses geometry-guided generative completion for sparse-view or unposed-video reconstruction \citep{ni2025g4splat}. \ours is complementary: it does not train a new video-to-3D generator or assume real sparse views. Instead, it wraps a frozen generic text-to-video prior with reconstruction, canonical-orbit fitting, coverage-aware completion, and a second reconstruction.}

\paragraph{Video diffusion for novel view synthesis.} CAT3D, ViewCrafter, and ReconX use generative priors to synthesize novel views from sparse reliable observations \citep{gao2024cat3d,yu2024viewcrafter,liu2026reconx}. ViewCrafter is especially important because it can produce high-quality local trajectories by progressively extending 3D clues. Our problem differs in the starting point and the output protocol. We do not start from a real calibrated image set; we start from a generated video whose own frames may be inconsistent. We do not only require a locally plausible trajectory; we require a closed canonical orbit suitable for 3D reconstruction and repeated rendering.

\paragraph{Generated videos as 3D priors.} Recent work such as WonderVerse demonstrates that video generation priors can support immersive 3D scene generation and extension \citep{feng2025wonderverse}. This direction motivates our central design principle: text-to-video models should not be treated merely as video renderers, but as scalable scene priors. \ours contributes a simple interface between such priors and 3D reconstruction: a first reconstruction builds coordinates, canonical rendering exposes coverage, video completion fills the missing interval, and a second reconstruction closes the orbit.

\paragraph{Gaussian Splatting and deformable reconstruction.} 3D Gaussian Splatting provides an explicit representation with efficient differentiable rendering \citep{kerbl20233d}. Deformable 3D Gaussians extend this representation to monocular dynamic scenes by learning a canonical Gaussian Splatting state and time-dependent deformation field \citep{yang2024deformable}. We use DeformableGS in a nonstandard way. The goal is not to preserve a real dynamic scene, but to absorb the pseudo-dynamics of a generated video--shape drift, texture repainting, and background inconsistency--and then compress the result into a static structural proxy.

\section{Method}

\subsection{Overview}

\begin{figure}[t]
\centering
\includegraphics[width=\linewidth]{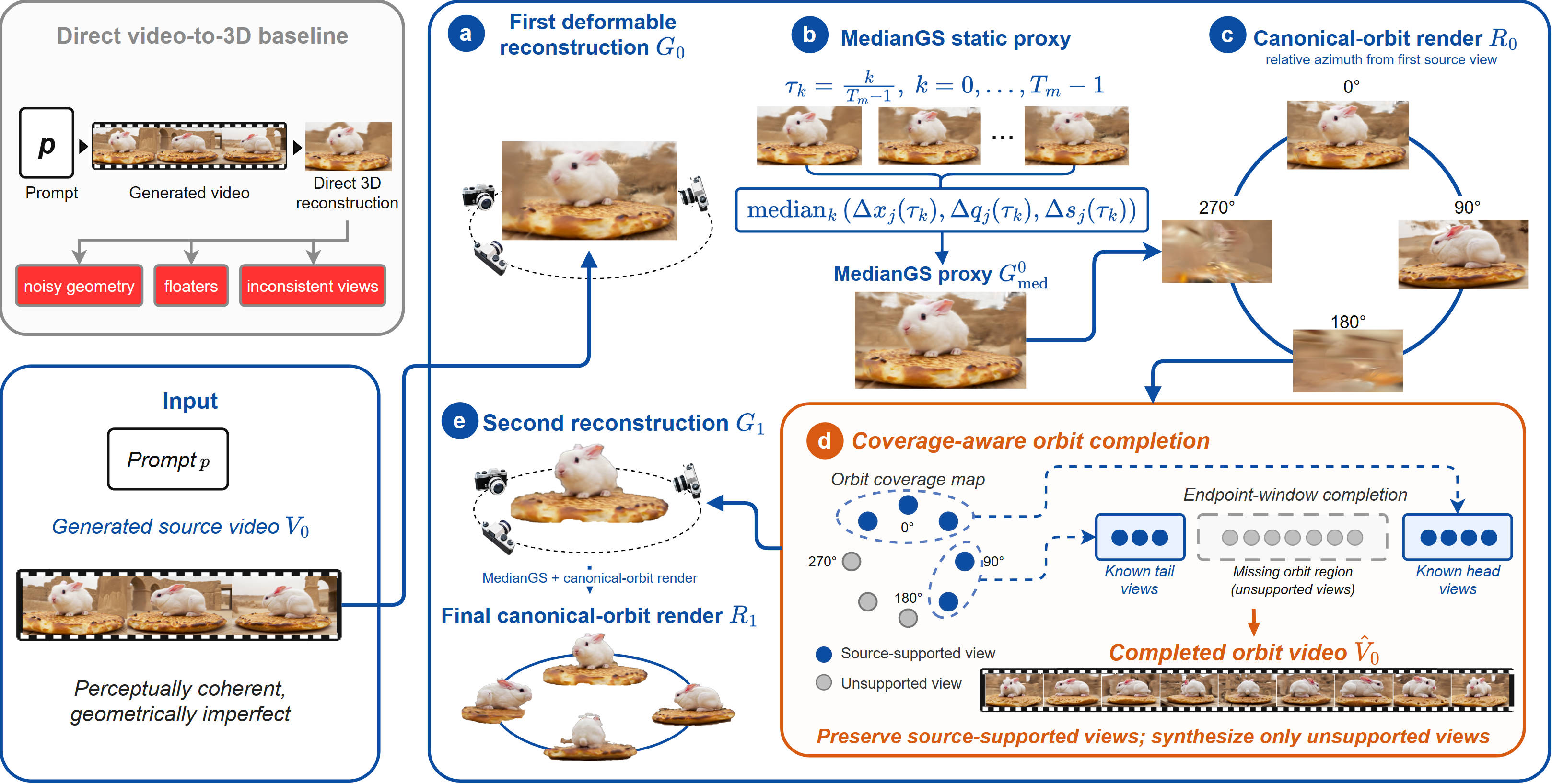}
\caption{Canonical-orbit reconstruction--completion loop. A frozen text-to-video model first supplies an uncalibrated source video. The first reconstruction organizes its frames into a shared coordinate system rather than serving as the final asset. MedianGS static scene consolidation turns this deformable reconstruction into a stable canonical-orbit render, whose supported arc and unsupported gap define the coverage-completion problem. The completed orbit video is then reconstructed under the prescribed canonical cameras to produce the final Gaussian Splatting scene and render.}
\label{fig:pipeline}
\end{figure}

Given a text prompt $p$, \ours produces a Gaussian Splatting reconstruction that renders views of a static scene on a prescribed 360-degree orbit. The pipeline is
\begin{equation}
  p \xrightarrow{\Phi} \Vzero
  \xrightarrow{\recon} \Gzero
  \xrightarrow{\medianop} \Gmed^0
  \xrightarrow{\renderer(\orbit)} \Rzero
  \xrightarrow{\completion} \Vhat
  \xrightarrow{\recon} \Gone
  \xrightarrow{\medianop,\renderer(\orbit)} \Rone .
  \label{eq:pipeline}
\end{equation}
Here $\Phi$ is a frozen text-to-video model, $\Vzero=\{I^0_t\}_{t=1}^{T}$ is the initial video, $\Gzero$ is the first Deformable Gaussian Splatting reconstruction, $\Gmed^0$ is its static MedianGS proxy, $\Rzero$ is the first canonical-orbit rendering, $\Vhat$ is the coverage-completed orbit video, $\Gone$ is the second reconstruction, and $\Rone$ is the final canonical-orbit rendering. The main 3D output is $\Gone$ through its static proxy and render $\Rone$. Figure~\ref{fig:pipeline} visualizes this loop as a sequence of preliminary static 3D reconstruction, canonical-orbit support analysis, coverage-aware completion, and final reconstruction.

For the \tthreebench{}-derived audit, the source-video prompt is not the raw short benchmark string alone. Appendix~\ref{app:prompt-protocol} describes how we preserve the original semantic target while expanding the video-model instruction with stable scene support, rigid-object behavior, and explicit orbit-camera control. This wrapper is part of the \ours{} source-video protocol rather than a prompt-normalized baseline comparison; baselines use their native prompt conventions and output protocols.

\begin{table}[t]
\caption{Positioning of \ours. The claim is not that every baseline fine-tunes a model or that every baseline lacks a full orbit; rather, each family solves a related but different problem. \ours targets a specific gap: converting generic text-generated videos into scene-level closed-orbit Gaussian Splatting reconstructions with no task-specific multiview fine-tuning, no per-prompt SDS optimization, and no progressive one-view-at-a-time expansion.}
\label{tab:positioning}
\centering
\small
\resizebox{\linewidth}{!}{%
\begin{tabular}{@{}lccccp{5.8cm}@{}}
\toprule
Family & Examples & Per-prompt SDS & Task-specific MV prior & Local/progressive expansion & What \ours adds \\
\midrule
SDS text-to-3D & DreamFusion, SJC, Magic3D & Yes & No & No & Uses video scene priors and avoids per-prompt score-distillation optimization. \\
Object/multiview priors & SV3D, MVDream, VideoMV, CoSER & No & Usually yes & No & Does not require a trained multiview format and preserves scene context beyond isolated assets. \\
Local NVS/video completion & CAT3D, ViewCrafter, ReconX & No & Varies & Often yes & Completes a closed canonical orbit rather than only a locally plausible view path. \\
Generated-video 3D scenes & WonderVerse-style pipelines & No & No & Varies & Uses the first reconstruction as an explicit anchor and coverage signal. \\
\textbf{\ours} & \textbf{OrbitForge} & \textbf{No} & \textbf{No} & \textbf{No} & \textbf{Frozen video priors + MedianGS + coverage-aware closed-orbit reconstruction.} \\
\bottomrule
\end{tabular}}
\end{table}

The method boundary is important. As shown in Figure~\ref{fig:pipeline}, \ours is not a new video model, not a multiview fine-tuning recipe, and not an SDS method. All generative priors are frozen. The only optimized representations are Gaussian Splatting reconstructions fitted to generated observations. Optional video cleanup can be applied after $\Rone$, but it is reported separately and is not required for the main 3D claim.

\subsection{Preliminary 3D reconstruction as an anchor}

The initial generated video is first assigned cameras by an estimator $\mathcal{E}$,
\begin{equation}
  \widehat{\mathcal{C}}_0 = \mathcal{E}(\Vzero)=\{\hat c_t\}_{t=1}^{T},
\end{equation}
and reconstructed with a Deformable Gaussian Splatting backend,
\begin{equation}
  \Gzero = \recon(\Vzero,\widehat{\mathcal{C}}_0).
\end{equation}
Let the canonical Gaussian Splatting parameters be
$\Theta=\{(x_j,q_j,s_j,\alpha_j,h_j)\}_{j=1}^{M}$, where $x_j$ is center, $q_j$ rotation, $s_j$ scale, $\alpha_j$ opacity, and $h_j$ spherical-harmonic appearance. A deformation field predicts geometry offsets conditioned on a normalized video time $\tau_t$:
\begin{equation}
  (\Delta x_j(\tau_t),\Delta q_j(\tau_t),\Delta s_j(\tau_t)) = F_\phi(x_j,\tau_t).
\end{equation}
The frame-level fit is
\begin{equation}
  I^0_t \approx
  \operatorname{Render}\left(\{x_j+\Delta x_j(\tau_t),q_j+\Delta q_j(\tau_t),s_j+\Delta s_j(\tau_t),\alpha_j,h_j\}_{j=1}^{M},\hat c_t\right).
  \label{eq:deformable-fit}
\end{equation}
The deformation field absorbs inconsistencies that a static model would be forced to bake into geometry. This does not mean that the scene is truly dynamic; it means that the reconstruction backend has enough flexibility to organize weak generated-video evidence without immediately destroying the coordinate system.

\subsection{MedianGS static scene consolidation}

A deformable reconstruction should not be rendered as a canonical orbit with a changing time code: in generated videos, time absorbs nonphysical pseudo-dynamics such as object drift and texture repainting. We therefore convert $\Gzero$ into a static MedianGS proxy. We evaluate the deformation field at $T_m$ uniformly spaced time codes $\tau_k=k/(T_m-1)$ and aggregate only geometry offsets by coordinate-wise medians \taskmedian{in the renderer's deformation-parameter space}:
\begin{equation}
  \overline{\Delta x}_j=\operatorname{median}_{k}\Delta x_j(\tau_k),\qquad
  \overline{\Delta q}_j=\operatorname{median}_{k}\Delta q_j(\tau_k),\qquad
  \overline{\Delta s}_j=\operatorname{median}_{k}\Delta s_j(\tau_k).
  \label{eq:median}
\end{equation}
Opacity and appearance are copied from the canonical Gaussian Splatting state. \taskmedian{The rotation aggregation in eq.~\eqref{eq:median} is not an intrinsic geodesic median on $\mathrm{SO}(3)$; it is a robust parameter-space aggregation for the additive rotation offsets used by the DeformableGS renderer, followed by normalization.} The static render state is
\begin{equation}
  \widetilde{x}_j=x_j+\overline{\Delta x}_j,
  \qquad
  \widetilde{q}_j=\frac{q_j+\overline{\Delta q}_j}{\|q_j+\overline{\Delta q}_j\|_2},
  \qquad
  \widetilde{s}_j=\max(|s_j+\overline{\Delta s}_j|,\epsilon_s),
  \qquad
  \widetilde{\ell}_j=\log \widetilde{s}_j,
\end{equation}
which defines
\begin{equation}
  \Gmed=\{\widetilde{x}_j,\widetilde{q}_j,\widetilde{\ell}_j,\alpha_j,h_j\}_{j=1}^{M}.
\end{equation}
Here $\widetilde{\ell}_j$ is the stored log-scale parameter used by the baked static Gaussian Splatting state.
The median is robust to abnormal offsets and avoids choosing an arbitrary temporal slice. It extracts the majority deformation state shared by the generated video and produces the structural proxy used for orbit rendering. \taskmedian{This heuristic is intentionally tied to the reconstruction parameterization; replacing it with a chordal or geodesic rotation statistic is a straightforward implementation variant, but is not required for the coverage-completion argument.}

\subsection{Coverage-Aware Closed-Orbit Completion}

\taskorbit{Before completion, \ours converts the irregular source-video trajectory into a deterministic canonical orbit. Source-frame cameras estimated from $\Vzero$ provide camera centers and viewing rays. We fit a stable look-at point from the source rays, estimate an orbit plane from the source camera centers, fit a circle in that plane, and then sample a fixed-radius 360-degree orbit whose cameras all look at the fitted scene center. Thus the orbit is derived from the generated video's recovered motion, but the completed sequence is rendered and reconstructed in a regular camera system.} \taskappcite{Appendix~\ref{app:canonical-orbit} details the VGGT camera estimation, orbit-plane and circle fitting, fixed-elevation look-at cameras, source-support mask construction, and endpoint-window assembly. Table~\ref{tab:vggt-orbit-robustness} reports that all 300 \ours{} prompts produced valid 361-view canonical orbits, with median 121 successful source cameras, median $128.7^\circ$ source-covered span, and median circle-fit RMSE 0.0096 scene units. Figures~\ref{fig:app-camera-fit} and~\ref{fig:app-support-mask} visualize the fitted orbit and known/unknown support mask; Figures~\ref{fig:app-completion-layout}--\ref{fig:app-stride} and Table~\ref{tab:completion-summary} ablate endpoint placement, angle alignment, and anchor stride.}

\taskorbit{We define the canonical orbit $\orbit=\{c_i\}_{i=0}^{N-1}$ with $N=361$ cameras corresponding to $0^\circ,1^\circ,\ldots,360^\circ$, where the endpoint duplicates the first view for loop closure. Rendering the MedianGS proxy gives}
\begin{equation}
  \Rzero = \renderer(\Gmed^0,\orbit)=\{R_{0,i}\}_{i=0}^{N-1}.
\end{equation}
\taskorbit{The source cameras are projected onto the fitted orbit plane and assigned to canonical indices by their wrapped relative azimuth from the first source camera:}
\begin{equation}
  \begin{aligned}
  \Delta\theta_t
    &=\operatorname{wrap}_{[0,360^\circ)}(\theta_t-\theta_{\mathrm{start}}),\\
  \mu(t)
    &=\operatorname{clip}\left(
      \operatorname{round}\left(\frac{\Delta\theta_t}{360^\circ}(N-1)\right),
      0,N-1\right),\\
  S_{\mathrm{obs}}
    &=\{\mu(t)\mid t=1,\ldots,T\}.
  \end{aligned}
\end{equation}
Here $\theta_{\mathrm{start}}$ is the fitted-plane azimuth of the first source camera.
\taskorbit{We order these correspondences by source-frame order and preserve the wrapped canonical interval between the first and last mapped targets. This gives a source-covered arc $S_{\mathrm{arc}}$ and an unsupported complement}
\begin{equation}
  U_{\mathrm{gap}}=\{0,\ldots,N-1\}\setminus S_{\mathrm{arc}}.
\end{equation}
\taskorbit{The completion sequence is arranged as an endpoint-window sequence: known tail $\rightarrow$ unknown gap $\rightarrow$ known head. The video model therefore completes a missing middle interval between two orbit endpoints rather than extrapolating an open sequence. Frames in $S_{\mathrm{arc}}$ are copied from $\Rzero$; frames in $U_{\mathrm{gap}}$ are synthesized by a frozen completion model:}
\begin{equation}
  \widetilde{V}_0 = \completion(\Rzero,S_{\mathrm{arc}},U_{\mathrm{gap}},p),\qquad
  \widehat I^0_i =
  \begin{cases}
    R_{0,i}, & i\in S_{\mathrm{arc}},\\
    \widetilde I^0_i, & i\in U_{\mathrm{gap}}.
  \end{cases}
  \label{eq:coverage-assembly}
\end{equation}
\taskorbit{This stage is not progressive view expansion. The unsupported interval is completed in a global endpoint-window pass under the canonical camera ordering. Errors can still be hallucinated by the video prior, but they are explicitly confined to completion-supplied support rather than confused with source-observed evidence.}

\subsection{Second Reconstruction and Output}

The completed orbit video $\Vhat=\{\widehat I^0_i\}_{i=0}^{N-1}$ supplies one frame per canonical camera. The second reconstruction preserves this camera system:
\begin{equation}
  \Gone=\recon(\Vhat,\orbit),\qquad
  \Rone=\renderer(\medianop(\Gone),\orbit).
\end{equation}
Preserving canonical cameras is essential. Re-estimating cameras from the completed video can yield smooth but geometrically misaligned reconstructions, because the completed frames are already indexed by the orbit we intend to render. The final output is the completed Gaussian Splatting scene $\Gone$ through its MedianGS proxy and canonical-orbit render $\Rone$.

\section{Evaluation: The Closed-Orbit Protocol}

\taskview{We freeze the evaluation set before computing any new closed-orbit metrics. The main quantitative protocol uses a 300-prompt common audit set derived from the \tthreebench~\citep{he2023t} prompt categories, pooling single-object, object-in-context, and multi-object/action prompts. Each method in the coverage audit has a resolved output for every prompt in this common set. The method list is DreamFusion, Latent-NeRF, Magic3D, SJC, SV3D, VideoMV, ViewCrafter, MedianGS-only $\Rzero$, and \ours{} $\Rone$. Coverage is reported for all methods, while the lower-tail full-orbit quality ranking is restricted to methods that expose or satisfy a comparable closed-orbit protocol.} \taskappcite{Appendix~\ref{app:additional-quant} expands this audit with the ViewCrafter local-trajectory measurement, coverage-qualified lower-tail table, coverage-quality plot, full support-stratified diagnostics, and anchor-to-view DINOv2 appearance consistency.}

\taskview{For methods with native videos, the evaluation samples frames uniformly from the available sequence. For methods with renderable orbit images, it uses the rendered sequence directly. The coverage audit uses the prescribed camera grid when that grid is part of the method output. ViewCrafter is evaluated with a declared local camera trajectory, so we audit that trajectory directly rather than assigning coverage from frame count alone. Quality metrics are computed on uniformly sampled views. Automatic metrics are treated as diagnostics rather than definitive perceptual judgments.}

\subsection{Coverage First, Smoothness Second}

A full-orbit method should not be evaluated only by adjacent-frame smoothness. A video that barely moves, or a method that generates a narrow local trajectory, can obtain excellent adjacent LPIPS or DINO scores while failing to cover the back side of the object. Conversely, a method that genuinely traverses 360 degrees must pass through more severe appearance changes and may look worse under local smoothness metrics. \taskclaim{We therefore separate evaluation into three axes: measured orbit coverage, full-orbit quality, and scene/context consistency.}

\paragraph{\taskclaim{Measured orbit coverage.}} Let $\{\theta_i\}$ be azimuths prescribed by the method's orbit grid or estimated from a rendered sequence, then sorted circularly. We define \taskclaim{measured} azimuth span as
\begin{equation}
  A = 360^\circ - \max_i \left(\theta_{i+1}-\theta_i\right),
  \label{eq:span}
\end{equation}
where the gap is computed on the circle. This penalizes trajectories that remain locally smooth but cover only a small arc. We also compute monotone progress $M$, the fraction of adjacent camera steps that move in the dominant orbit direction. \taskdnear{Because the frozen 300-prompt \tthreebench{}-derived audit contains deterministic camera-grid rows and a separate camera-estimated ViewCrafter row, we define the reported closed-orbit success gate only from the two measured geometric quantities, $A>330^\circ$ and $M>0.9$. Feature near-loop distances such as loop DINO are kept as secondary diagnostics rather than hidden success gates.} \taskview{Table~\ref{tab:closed-orbit-coverage-audit} reports the measured coverage audit. ViewCrafter's declared 25-frame trajectory spans only $18.9^\circ$ in horizontal azimuth, so it is retained as a local-view baseline in coverage and qualitative comparisons but not ranked as a full-orbit method.}

\paragraph{Full-orbit quality.} \taskview{We report prompt alignment and single-view reward over sampled views, but only after orbit coverage has been established. Otherwise, a local novel-view method can obtain a high lower-tail score by staying near the input view and avoiding the hard backside or completion-gap views. Table~\ref{tab:closed-orbit-coverage-audit} therefore separates the coverage audit from the coverage-qualified lower-tail ranking. ViewCrafter's local-view ImageReward is reported separately in Appendix~\ref{app:viewcrafter-local-diagnostic}; it is not a full-orbit ranking row.}

\paragraph{Scene/context consistency.} We distinguish isolated object turntables from scene-level orbits. Ground, background, support surface, reflections, and object-environment relationships are part of the output protocol for \ours. This is why the qualitative comparison in Figure~\ref{fig:qualitative-main} is not merely aesthetic: it shows whether a method preserves a scene around the object. \taskanchor{Appendix~\ref{app:anchor-appearance-consistency} adds an anchor-to-view DINOv2 appearance-consistency diagnostic for the key coverage-qualified video methods.}

\begin{table}[t]
\caption{\taskview{Coverage first, quality second on the frozen 300-prompt \tthreebench{}-derived audit set. Left: measured closed-orbit coverage, including ViewCrafter's declared local trajectory. \taskdnear{Closed success is the explicit geometric gate $A>330^\circ$ and $M>0.9$; feature loop distances are reported only as secondary diagnostics.} Right: coverage-qualified lower-tail quality, reported only for methods with an explicit or measured full-orbit protocol. ViewCrafter is excluded from the full-orbit quality ranking because its measured trajectory covers a local $18.9^\circ$ arc rather than a comparable closed orbit.}}
\label{tab:closed-orbit-coverage-audit}
\centering
\scriptsize
\begin{minipage}[t]{0.46\linewidth}
\centering
(a) Closed-orbit coverage
\resizebox{\linewidth}{!}{%
\begin{tabular}{lrrrrr}
\toprule
Method & N & Views & $A$ $\uparrow$ & $M$ $\uparrow$ & Closed $\uparrow$ \\
\midrule
DreamFusion~\citep{poole2022dreamfusion} & 300 & 120 & 357.0 & 1.00 & 100\% \\
Latent-NeRF~\citep{metzer2023latent} & 300 & 120 & 357.0 & 1.00 & 100\% \\
Magic3D~\citep{lin2023magic3d} & 300 & 120 & 357.0 & 1.00 & 100\% \\
SJC~\citep{wang2023score} & 300 & 120 & 357.0 & 1.00 & 100\% \\
SV3D~\citep{voleti2024sv3d} & 300 & 21 & 342.9 & 1.00 & 100\% \\
VideoMV~\citep{zuo2024videomv} & 300 & 24 & 345.0 & 1.00 & 100\% \\
ViewCrafter~\citep{yu2024viewcrafter} & 300 & 25 & 18.9 & 0.54 & 0\% \\
MedianGS-only $\Rzero$ (ours) & 300 & 361 & 359.0 & 1.00 & 100\% \\
\ours{} $\Rone$ & 300 & 361 & 359.0 & 1.00 & 100\% \\
\bottomrule
\end{tabular}}
\end{minipage}\hfill
\begin{minipage}[t]{0.53\linewidth}
\centering
(b) Coverage-qualified quality
\resizebox{\linewidth}{!}{%
\begin{tabular}{lrrrr}
\toprule
Method & N & Views & Q10 IR $\uparrow$ & Mean IR $\uparrow$ \\
\midrule
DreamFusion~\citep{poole2022dreamfusion} & 300 & 12 & $7.29{\pm}6.91$ & 10.05 \\
Latent-NeRF~\citep{metzer2023latent} & 300 & 12 & $12.78{\pm}10.99$ & 18.11 \\
Magic3D~\citep{lin2023magic3d} & 300 & 12 & $6.37{\pm}6.21$ & 8.37 \\
SJC~\citep{wang2023score} & 300 & 12 & $7.23{\pm}5.50$ & 12.14 \\
SV3D~\citep{voleti2024sv3d} & 300 & 12 & $13.41{\pm}16.97$ & 24.13 \\
VideoMV~\citep{zuo2024videomv} & 300 & 12 & $18.74{\pm}18.10$ & 30.52 \\
MedianGS-only $\Rzero$ (ours) & 300 & 12 & $8.51{\pm}8.98$ & 23.72 \\
\ours{} $\Rone$ & 300 & 12 & $16.79{\pm}16.10$ & 28.52 \\
\bottomrule
\end{tabular}}
\end{minipage}
\end{table}

\taskview{The appendix coverage-quality audit in Figure~\ref{fig:coverage-pareto} joins measured azimuth span with lower-tail ImageReward for coverage-qualified full-orbit outputs. It highlights the main effect of completion: the weak-view tail rises from MedianGS-only $\Rzero$ to \ours{} $\Rone$ while retaining the largest measured span; ViewCrafter is kept outside this frontier because its measured trajectory is local.}

\subsection{R0 versus R1: Why Completion Is Necessary}

\begin{figure}[t]
\centering
\includegraphics[width=\linewidth]{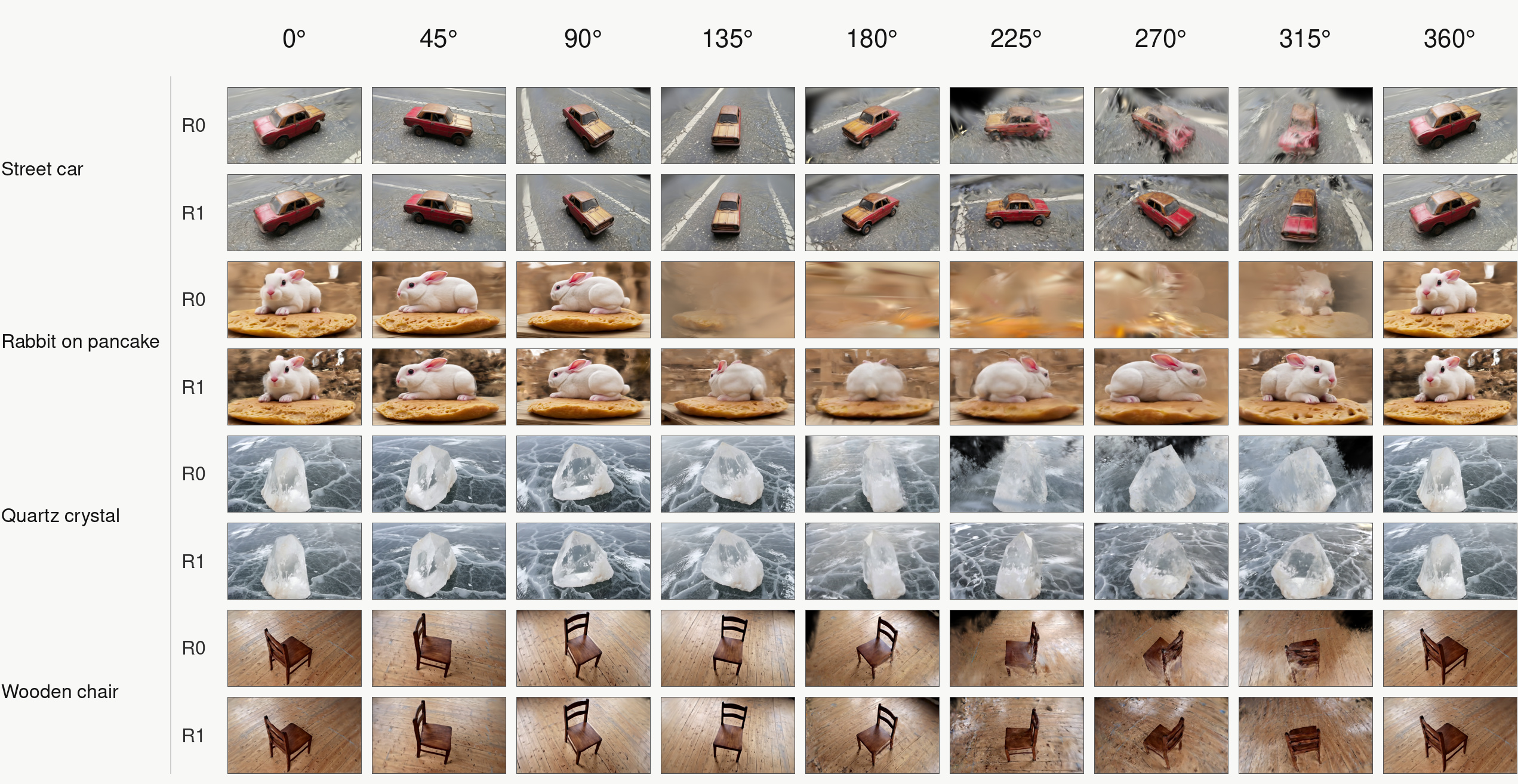}
\caption{\taskorbit{$\Rzero$ versus $\Rone$ on the same canonical cameras. The first MedianGS render organizes the source video into an orbit but remains weak in originally unsupported views. Coverage-aware completion supplies those missing views before the second reconstruction, producing a fuller scene-level orbit while retaining the same canonical camera system.}}
\label{fig:r0-r1-main}
\end{figure}

\taskorbit{The key ablation is whether the first-stage MedianGS proxy is sufficient. It is not: $\Rzero$ inherits partial source support, so unsupported views show stretched background, missing backsides, floaters, and texture smear, whereas $\Rone$ is reconstructed from a completed canonical sequence with an input frame for every canonical camera. Figure~\ref{fig:r0-r1-main} shows this support change.} \taskappcite{Appendix~\ref{app:qualitative-results} provides additional examples, while Tables~\ref{tab:r0-r1-support} and~\ref{tab:support-stratified-quality-full} report the full input-availability and regional ImageReward statistics.}

\taskorbit{Adjacent-frame metrics can therefore be misleading: $\Rone$ may have higher adjacent pixel difference because it traverses completed views with more content change. The support-stratified audit in Appendix~\ref{app:additional-quant} shows the originally unsupported gap-region Q10 rising from 8.07 to 16.36, isolating completion's effect on the hard views.} \taskappcite{Appendices~\ref{app:reconstruction-staticization}--\ref{app:camera-continuity} give the first-reconstruction, completion-design, and camera-continuity ablations supporting the final MedianGS, endpoint-window, anchor-stride, and canonical-camera settings.}

\section{Discussion}

\taskview{\paragraph{Limitations.} The completed backside is synthesized support: if the completion prior changes the object or fails on transparent, thin, or reflective structures, the second reconstruction can solidify that error. The Gaussian Splatting scene can still contain floaters, blur, and background debris, and local view quality lags behind specialized local-view methods.} \taskappcite{Representative difficult views are shown in Figure~\ref{fig:failure-modes}; Appendices~\ref{app:optional-cleanup}--\ref{app:related-work-boundary} give optional cleanup boundaries, reproducibility constants, and related-work positioning.}

\section{Conclusion}

OrbitForge converts a single text-generated video into a closed-orbit 3D Gaussian Splatting scene by using the first reconstruction as a scaffold, completing unsupported views, and reconstructing again under fixed cameras. It improves weak orbit regions while inheriting limitations from the source video, camera estimator, and completion prior.

\clearpage
\appendix
\setcounter{topnumber}{5}
\setcounter{bottomnumber}{2}
\setcounter{totalnumber}{7}
\renewcommand{\topfraction}{0.95}
\renewcommand{\bottomfraction}{0.85}
\renewcommand{\textfraction}{0.05}
\renewcommand{\floatpagefraction}{0.80}
\setlength{\textfloatsep}{8pt plus 2pt minus 2pt}
\setlength{\floatsep}{8pt plus 2pt minus 2pt}
\setlength{\intextsep}{8pt plus 2pt minus 2pt}
\renewcommand{\theHfigure}{\Alph{section}.\arabic{figure}}
\renewcommand{\theHtable}{\Alph{section}.\arabic{table}}
\renewcommand{\theHalgorithm}{\Alph{section}.\arabic{algorithm}}

\section{Prompt Protocol for Text-to-Video Source Generation}
\label{app:prompt-protocol}

The frozen audit is indexed by the original \tthreebench{}~\citep{he2023t} prompts: 100 single-object prompts, 100 object-in-context prompts, and 100 multi-object or action prompts. These short prompts are suitable semantic targets, but they are under-specified for a generic text-to-video model whose camera path, framing, background stability, and object motion are all language controlled. \ours therefore separates the benchmark item from the source-video instruction. The original prompt remains the semantic target for the audit, while the source-video prompt adds only the information needed to obtain a reconstructible orbit: stable support surface, static lighting, rigid subject behavior, and controlled camera motion.

The conversion has two stages. First, a text-to-text prompt enricher attempts to turn the short object description into a concise 3D scene prompt. For the \tthreebench{} source-video runs, automatic enrichment is enabled as a proposal mechanism: we use a T5-family model, FLAN-T5-small, with the instruction to keep the main subject unchanged while adding a plausible stable location, visible ground or support-surface details, and static lighting. The generated text is accepted only when it preserves the subject evidence and introduces concrete environment terms such as ground, floor, surface, pavement, garden, stone, or table. If the text model is unavailable or returns a vague restatement of the object, a deterministic object-category rule supplies the support surface; for example, household ceramics are placed on wooden or ceramic tabletops, toys on wooden surfaces, small outdoor objects on pavement or soil, jewelry on velvet display pads, and instruments on a matte stage floor. This fallback is part of the fixed prompt protocol rather than per-example hand editing.

Second, the enriched subject and location are wrapped into a video-model instruction with an explicit camera and rigidity protocol. The camera is requested as a single continuous, constant-radius 360-degree side orbit with a fixed long lens, no zoom, no dolly, no speed variation, low-to-medium viewing height, and constant elevation. The prompt also asks the subject to remain locked to its support surface: no rolling, sliding, vibration, topology change, deformation, texture repainting, or view-dependent reinterpretation. These clauses are important because the video prior is otherwise free to create a visually plausible clip whose apparent object identity, contact points, or camera elevation drift over time.

\begin{table}[t]
\centering
\small
\resizebox{\linewidth}{!}{%
\begin{tabular}{p{0.19\linewidth}p{0.32\linewidth}p{0.39\linewidth}}
\toprule
Stage & Example content & Purpose \\
\midrule
Original benchmark item & ``A pair of polka-dotted sneakers'' & Defines the semantic target and audit index. \\
Scene-enriched subject & A pair of polka-dotted sneakers standing side by side on rough concrete pavement, with the soles flat on the ground. & Adds physical support, contact evidence, and static environment context for the video prior. \\
Orbit-camera wrapper & Single continuous constant-radius 360-degree orbit; fixed long lens; no zoom or dolly; stable side-facing framing; rigid subject and stable ground. & Requests a source video whose motion can be estimated, regularized into a canonical orbit, and completed without confusing object motion for camera motion. \\
\bottomrule
\end{tabular}}
\caption{Prompt handling for source-video generation. The benchmark item is preserved as the semantic target; the video-model prompt adds scene support and camera-control language because the source video prior controls both appearance and motion through text.}
\label{tab:prompt-protocol}
\end{table}

This prompt protocol does not provide cameras, geometry, or a hidden 3D template. It only asks the frozen video model to produce a more reconstructible source clip. The cameras used by \ours are still estimated from the generated frames and then regularized into the canonical orbit in Appendix~\ref{app:canonical-orbit}. For qualitative examples whose prompts are already written with this full camera and rigidity protocol, the prompt editor leaves them unchanged. For baselines with different input protocols, we use their native prompt formats rather than forcing this orbit-camera instruction onto methods that do not consume text-to-video prompts.

The camera and rigidity wrapper is therefore part of the input-generation protocol for \ours, not an additional supervision signal or a baseline-specific advantage. It induces a reconstructible source trajectory from a generic text-to-video prior, after which all cameras and support masks are recovered from the generated frames. We do not claim prompt-normalized superiority over every method family; the claim is that, under this frozen source-video protocol, \ours{} converts a generic generated video into a closed-orbit Gaussian Splatting scene and improves the originally unsupported orbit interval.

\FloatBarrier

\section{Canonical Orbit Construction and Coverage-Aware Completion}
\label{app:canonical-orbit}

\subsection{Source Camera Estimation and Look-at Center}

\taskorbit{The source video is first assigned per-frame cameras with VGGT. These cameras are not the final cameras used for reconstruction of the completed orbit; they provide a geometric reference for the source trajectory. Let $\hat{\mathbf{o}}_t\in\mathbb{R}^3$ be the estimated camera center and $\hat{\mathbf{d}}_t\in\mathbb{R}^3$ its unit forward ray. We estimate a single look-at point by least-squares ray intersection,}
\begin{equation}
  \mathbf{x}_{\mathrm{look}}
  =
  \arg\min_{\mathbf{x}}
  \sum_{t=1}^{T}
  \left\|
  \left(I-\hat{\mathbf{d}}_t\hat{\mathbf{d}}_t^\top\right)
  \left(\mathbf{x}-\hat{\mathbf{o}}_t\right)
  \right\|_2^2 .
  \label{eq:lookat-fit}
\end{equation}
\taskorbit{This point is used as the common orientation target for the canonical cameras. Source-camera failures or outliers are handled by using the common successful camera set for the robust trajectory fit.}

\begin{figure}[t]
\centering
\includegraphics[width=0.95\linewidth]{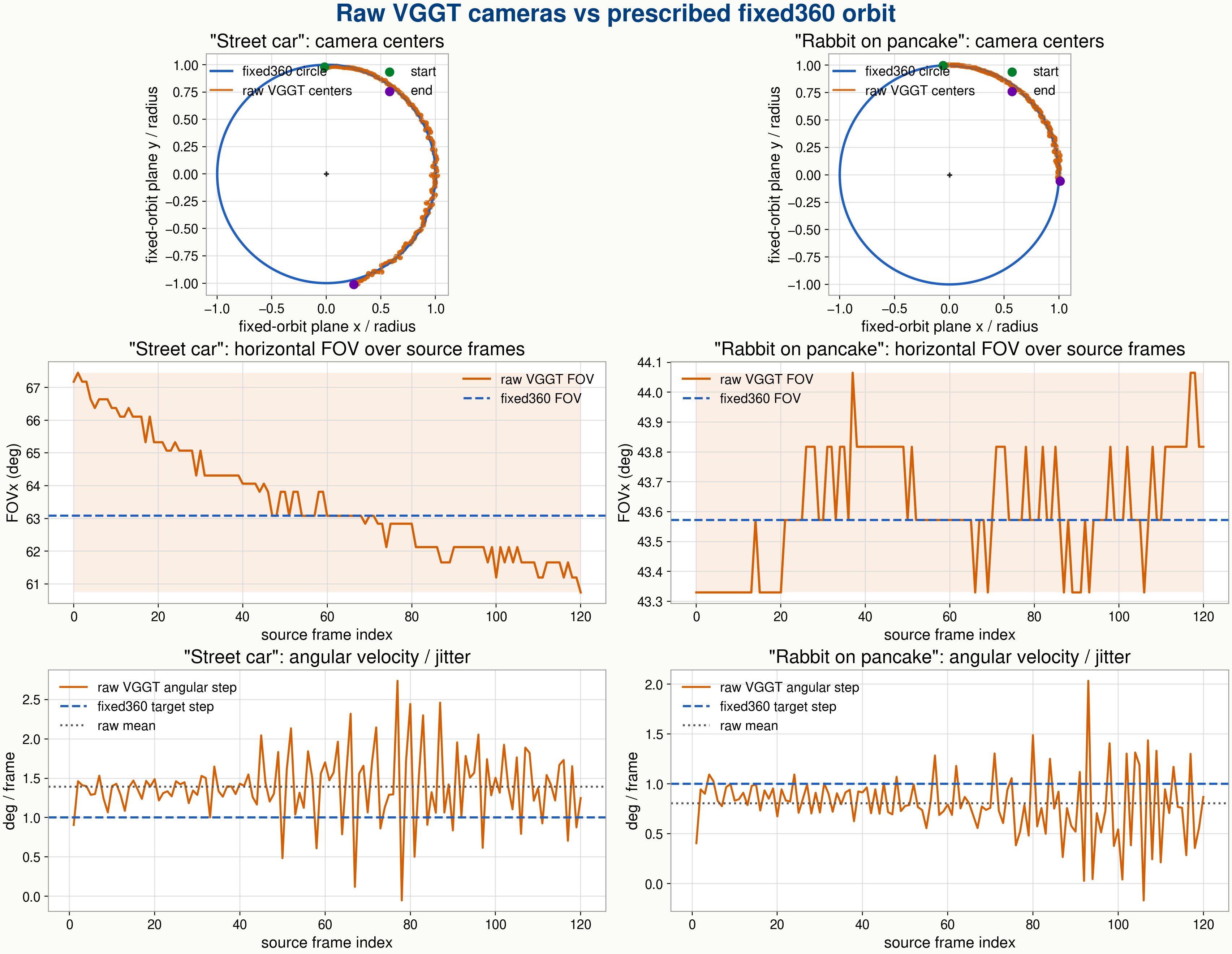}
\caption{\taskorbit{Source camera estimates and fitted canonical orbit. The recovered source cameras provide an irregular partial trajectory; the canonical orbit regularizes this trajectory into a fixed 360-degree camera system while preserving the recovered scene center and orbit scale.}}
\label{fig:app-camera-fit}
\end{figure}

\subsection{\taskrobust{VGGT Fitting Checks and Robustness Boundary}}

\taskrobust{The canonical orbit fit uses VGGT as a trajectory reference, not as the final camera system. The implementation evaluates multiple camera-axis conventions and selects the one with the smallest mean look-at ray-intersection error. It then fits the orbit plane by SVD over successful source camera centers, chooses the normal sign to agree with the average source-camera up direction, fits the in-plane circle by least squares, and uses median source intrinsics after scaling to the canonical render resolution. These choices regularize noisy generated-video cameras into one deterministic orbit instead of preserving frame-to-frame camera shake, roll drift, or focal-length jitter.}

\taskrobust{The robustness boundary is also explicit. If source cameras are too unreliable, the fitted orbit can inherit an incorrect scene center, wrong scale, or nonmonotone source-to-orbit mapping. \ours{} does not hide this uncertainty by treating completed views as source-observed evidence: $S_{\mathrm{obs}}$ records exact source-camera support, $S_{\mathrm{arc}}$ records the continuous first-reconstruction-supported arc, and $U_{\mathrm{gap}}$ is completion-supplied support. Failures in the source-camera fit therefore appear as visible orbit misalignment or completion-gap drift, which are represented in the failure cases rather than counted as direct source support.}

\taskfitstats{Table~\ref{tab:vggt-orbit-robustness} audits the existing camera-fitting records for the \ours{} runs on the frozen 300-prompt \tthreebench{}-derived audit set. The table is a fitting diagnostic, not a new baseline result: it checks that the source-camera stage produced a valid canonical orbit for each prompt and reports how much of the orbit was source-supported before completion.}

\begin{table}[t]
\centering
\small
\begin{tabular}{lccp{2.9cm}}
\toprule
Diagnostic & Median / count & P90 / range & Unit \\
\midrule
Prompts with branch summary & 300 / 300 &  &  \\
Prompts with valid 361-view orbit & 300 / 300 &  &  \\
Successful source cameras & 121 median & 121--121 & source frames \\
Source-covered azimuth span & 128.7 & 178.4 & degrees \\
Circle-fit RMSE & 0.0096 & 0.0143 & scene units \\
Planar radius std. & 0.0096 & 0.0143 & scene units \\
Height std. from fitted plane & 0.0039 & 0.0091 & scene units \\
Look-at convention error & 0.38 & 0.99 & mean degrees \\
Observed canonical bins & 105 & 117 & of 361 \\
Completion-gap bins & 232 & 282 & of 361 \\
Selected forward convention & col2: 299, row2: 1 &  & count \\
\bottomrule
\end{tabular}
\caption{\taskfitstats{VGGT-to-canonical-orbit fitting diagnostics for \ours{} on the frozen 300-prompt \tthreebench{}-derived audit set. The 361-view orbit is prescribed after fitting; the source-covered span and completion-gap rows quantify why coverage-aware completion is needed rather than claiming that the full orbit is source-observed.}}
\label{tab:vggt-orbit-robustness}
\end{table}

\subsection{Orbit Plane and Circle Fitting}

\taskorbit{Given successful source camera centers, we estimate the trajectory plane by centering the cameras and applying SVD. Let}
\begin{equation}
  \bar{\mathbf{o}}=\frac{1}{T}\sum_{t=1}^{T}\hat{\mathbf{o}}_t,\qquad
  X=
  \begin{bmatrix}
    (\hat{\mathbf{o}}_1-\bar{\mathbf{o}})^\top\\
    \cdots\\
    (\hat{\mathbf{o}}_T-\bar{\mathbf{o}})^\top
  \end{bmatrix}.
\end{equation}
\taskorbit{The two dominant right-singular directions define the orbit-plane basis $(\mathbf{e}_1,\mathbf{e}_2)$, and the smallest direction defines the plane normal $\mathbf{n}$. The normal sign is chosen to agree with the average source-camera up direction. We project the camera centers into this plane:}
\begin{equation}
  u_t=(\hat{\mathbf{o}}_t-\bar{\mathbf{o}})^\top\mathbf{e}_1,\qquad
  v_t=(\hat{\mathbf{o}}_t-\bar{\mathbf{o}})^\top\mathbf{e}_2 .
\end{equation}
\taskorbit{The circle center $(a,b)$ and radius $r$ are fit in 2D by the linear least-squares system}
\begin{equation}
  2a u_t + 2b v_t + \gamma \approx u_t^2+v_t^2,\qquad
  r=\sqrt{a^2+b^2+\gamma}.
  \label{eq:circle-fit}
\end{equation}
\taskorbit{The fitted orbit center in 3D is}
\begin{equation}
  \mathbf{c}_{\mathrm{orb}}=\bar{\mathbf{o}}+a\mathbf{e}_1+b\mathbf{e}_2 .
\end{equation}

\subsection{Canonical Azimuth, Elevation, and Intrinsics}

\taskorbit{The canonical orbit uses the source trajectory only to determine a stable plane, center, radius, and start angle. The start angle is the polar angle of the first source camera in the fitted 2D circle coordinates,}
\begin{equation}
  \theta_0=\operatorname{atan2}(v_1-b,u_1-a).
\end{equation}
\taskorbit{For $N=361$, the sampled canonical camera centers are}
\begin{equation}
  \theta_i=\theta_0+\frac{2\pi i}{N-1},\qquad
  \mathbf{o}_i=
  \mathbf{c}_{\mathrm{orb}}
  +r\cos\theta_i\,\mathbf{e}_1
  +r\sin\theta_i\,\mathbf{e}_2,
  \qquad i=0,\ldots,N-1 .
  \label{eq:canonical-camera-centers}
\end{equation}
\taskorbit{The endpoint $i=N-1$ duplicates the first azimuth and is used only for closed-loop export and loop diagnostics. The reported experiments do not let the camera height or elevation drift per frame; the recovered source-camera height variation is treated as trajectory noise rather than as a degree of freedom for orbit completion. The orbit radius is the fitted circle radius, and the intrinsics use the median source focal length after scaling to the render resolution.}

\subsection{Look-at Orientation and Roll}

\taskorbit{Each canonical camera is oriented by a look-at frame aimed at $\mathbf{x}_{\mathrm{look}}$. Its forward vector is}
\begin{equation}
  \mathbf{f}_i=
  \frac{\mathbf{x}_{\mathrm{look}}-\mathbf{o}_i}
       {\|\mathbf{x}_{\mathrm{look}}-\mathbf{o}_i\|_2}.
\end{equation}
\taskorbit{The orbit normal provides the up reference, with the sign aligned to the source-camera up direction. We do not re-estimate roll or per-view elevation from the completed video. Keeping roll fixed and using one shared look-at target makes the completion and second reconstruction operate in one explicit camera system rather than in cameras re-estimated from already completed frames.}

\subsection{Source-Supported Views and Completion Gap}

\taskorbit{Each source camera is assigned to the closest canonical bin by its relative azimuth on the fitted orbit plane,}
\begin{equation}
  \begin{aligned}
  \theta^{\mathrm{src}}_t
    &=\operatorname{atan2}(v_t-b,u_t-a),\\
  \Delta\theta_t
    &=\operatorname{wrap}_{[0,2\pi)}(\theta^{\mathrm{src}}_t-\theta_0),\\
  k_t
    &=\operatorname{clip}\left(
      \operatorname{round}\left(\frac{\Delta\theta_t}{2\pi}(N-1)\right),
      0,N-1\right),\\
  S_{\mathrm{obs}}
    &=\{k_t\}_{t=1}^{T}.
  \end{aligned}
\end{equation}
\taskorbit{$S_{\mathrm{obs}}$ contains exact source-camera correspondences, but the first reconstruction can render a continuous local arc around them. We therefore preserve the wrapped source-covered arc $S_{\mathrm{arc}}$ from the first to the last source-mapped bin in source-frame order and complete only its complement $U_{\mathrm{gap}}$. We never treat $U_{\mathrm{gap}}$ as source-observed support; it is completion-supplied support that is later fused by the second reconstruction.}

\begin{figure}[t]
\centering
\includegraphics[width=0.62\linewidth]{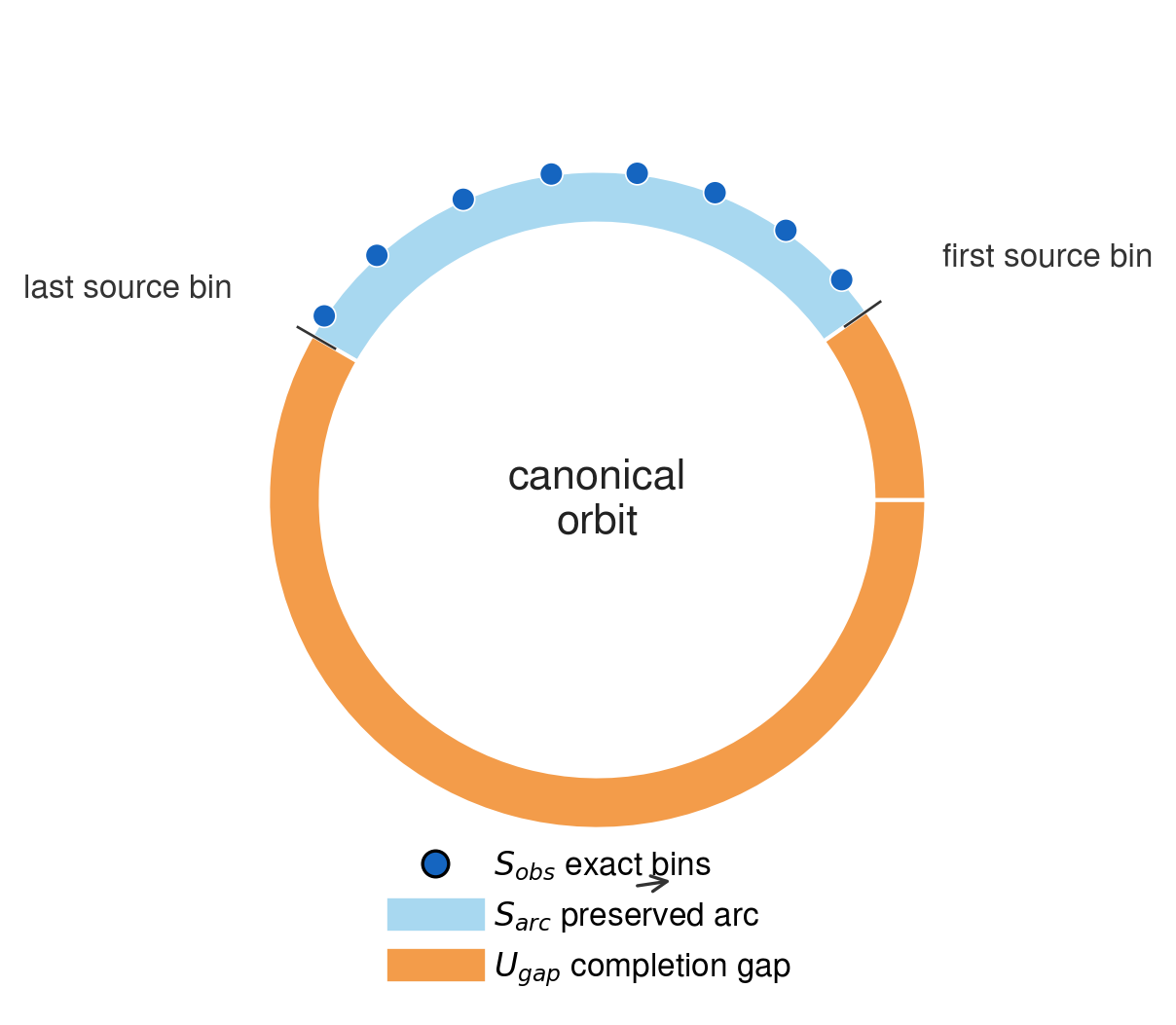}
\caption{\taskorbit{View-support mask on the canonical orbit. Exact source-observed bins $S_{\mathrm{obs}}$ define sparse correspondences, the wrapped source-covered arc $S_{\mathrm{arc}}$ is preserved from the first reconstruction, and the complementary gap $U_{\mathrm{gap}}$ is the only interval synthesized by orbit completion.}}
\label{fig:app-support-mask}
\end{figure}

\subsection{Endpoint-Window Completion Layout}

\taskorbit{The completion sequence is not given to the video model as an open-ended extrapolation. Instead, the preserved source-covered arc is split around the unsupported interval and arranged as an endpoint-window sequence: a known tail, the unknown gap, and a known head. The model is therefore asked to fill a missing interval whose two endpoints are already known. After generation, frames are restored to canonical order, and the final assembled sequence keeps the preserved arc from $\Rzero$ while replacing only $U_{\mathrm{gap}}$ with completed views.}

\subsection{Design Ablations}

\taskorbit{The endpoint-window layout, source-angle-aligned sparse anchors, and anchor stride are evaluated by visual strips and trajectory diagnostics. Non-endpoint layouts increase loop inconsistency; target-half anchor placement misaligns source views with their canonical azimuths; and overly sparse or overly dense anchors can introduce stalls or reversals.}

\begin{figure}[t]
\centering
\includegraphics[width=0.95\linewidth]{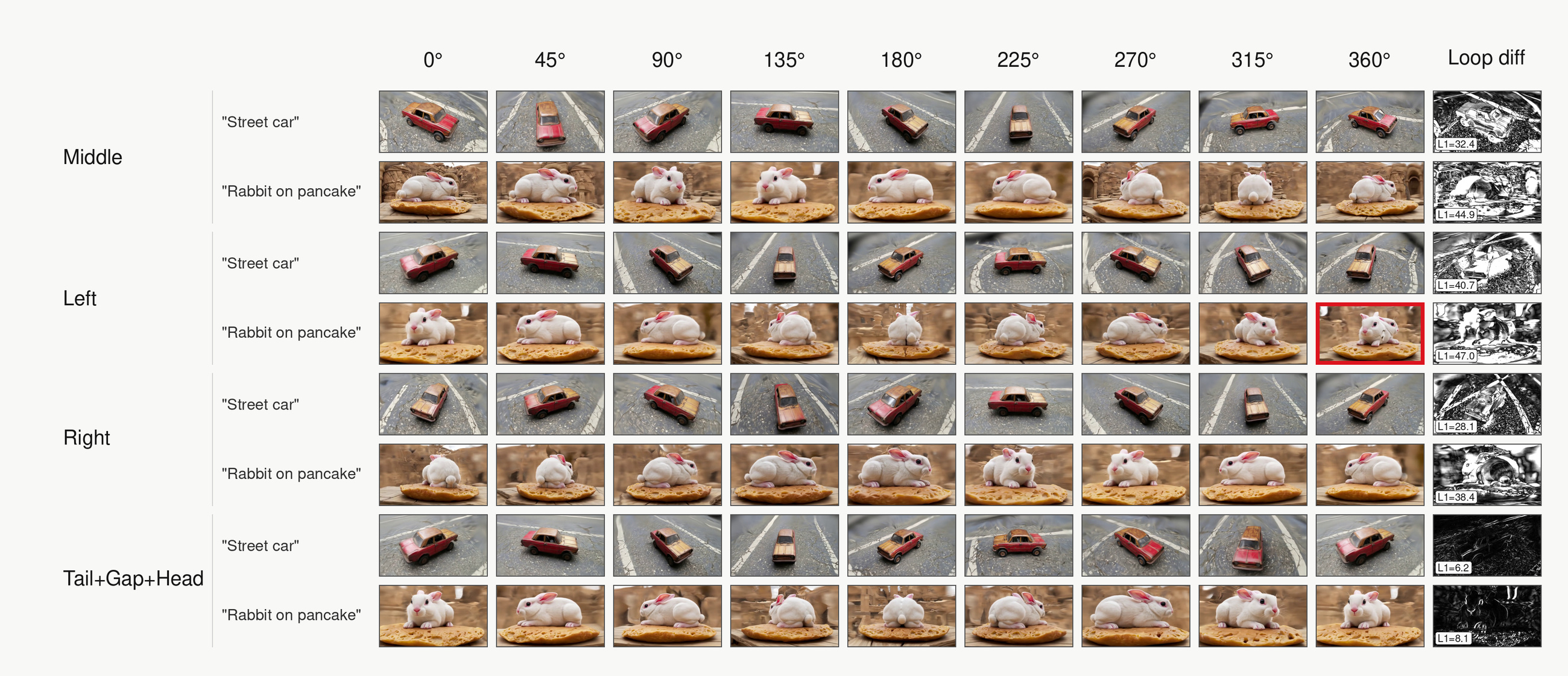}
\caption{\taskorbit{Endpoint-window placement ablation. Completing the unknown interval between two known orbit endpoints gives substantially lower endpoint discrepancy than layouts that make the model extrapolate or place the gap away from the orbit closure.}}
\label{fig:app-completion-layout}
\end{figure}

\begin{figure}[t]
\centering
\includegraphics[width=0.95\linewidth]{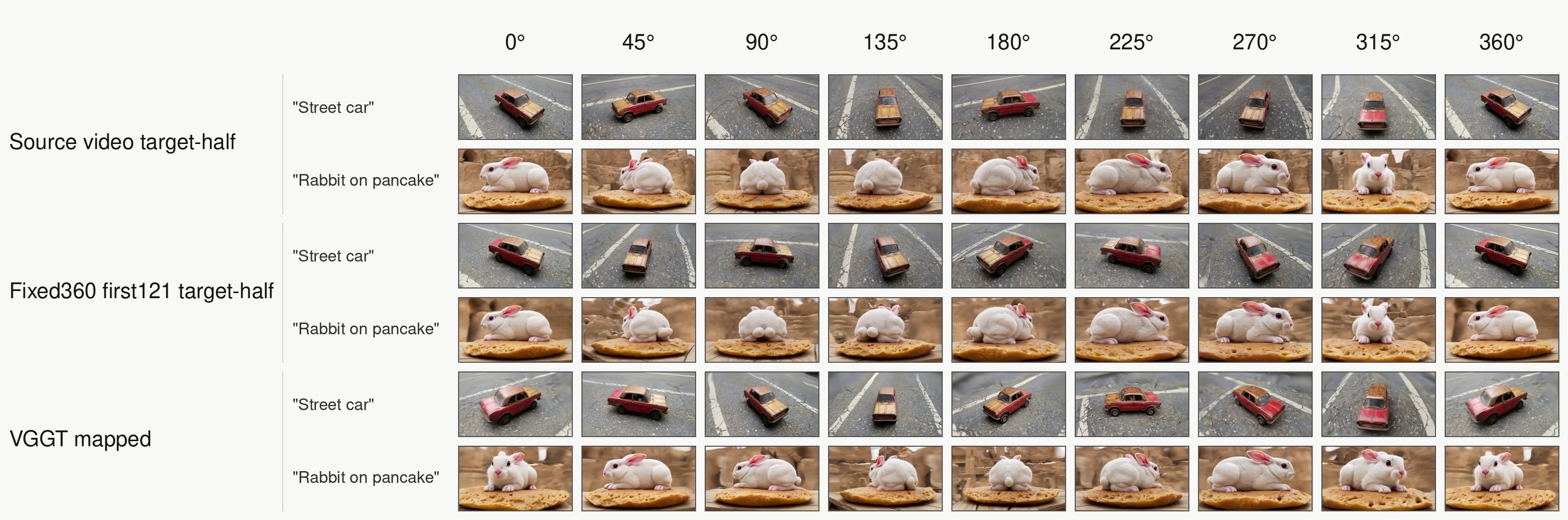}
\caption{\taskorbit{Sparse-anchor angle alignment ablation. Anchors aligned by source-to-canonical azimuth preserve where the source evidence belongs on the orbit; naive target-half placement can put reliable source views at the wrong canonical angles.}}
\label{fig:app-angle-alignment}
\end{figure}

\begin{figure}[t]
\centering
\includegraphics[width=0.95\linewidth]{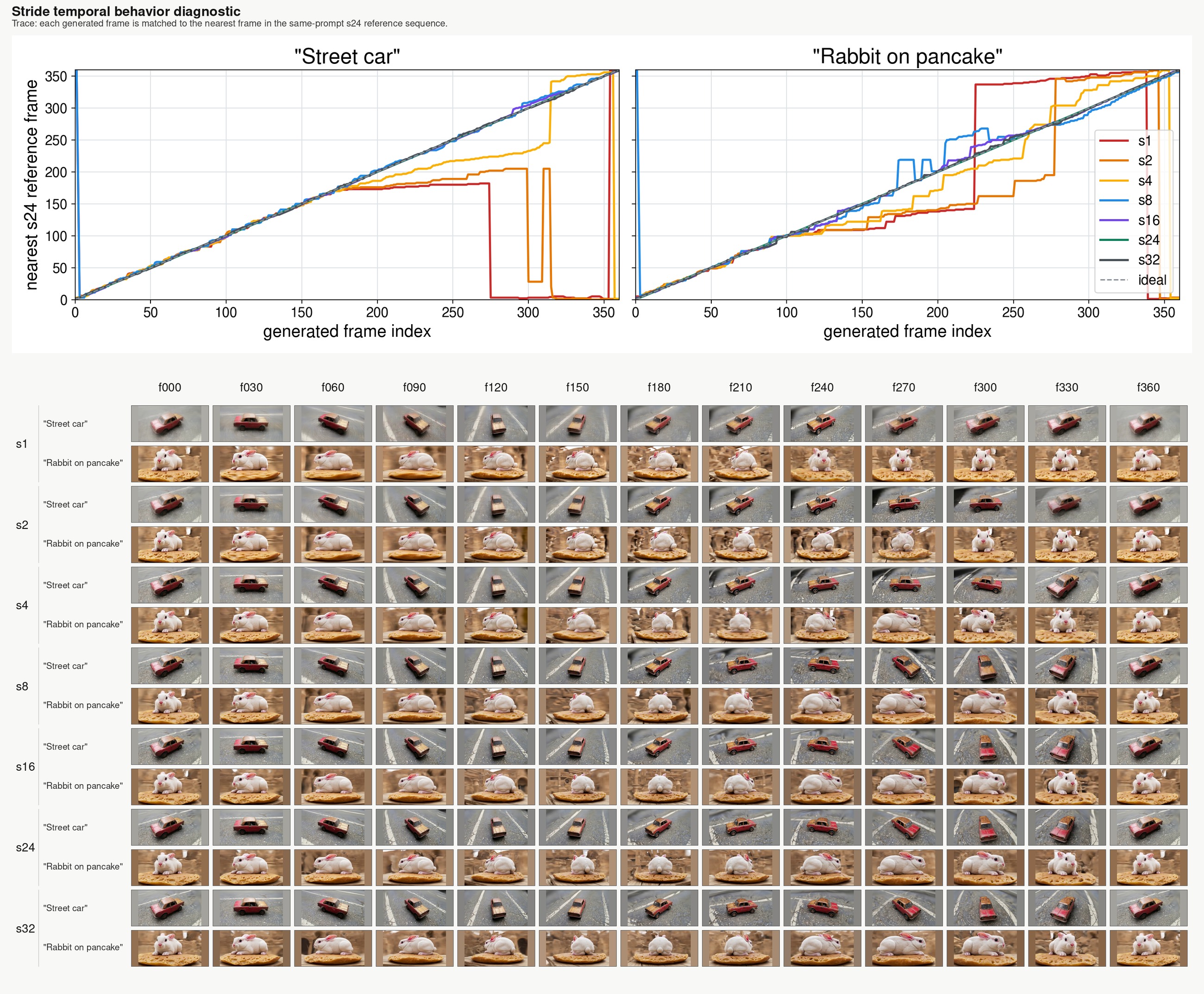}
\caption{\taskorbit{Anchor-stride ablation. The selected stride balances endpoint constraints with temporal progress, avoiding both weak conditioning and excessive anchor-induced stalls.}}
\label{fig:app-stride}
\end{figure}

\begin{table}[t]
\caption{\taskorbit{Coverage-aware completion design summary. Loop endpoint L1 measures generated canonical $I_{360}$ versus $I_0$; anchor index error measures whether sparse keyframes are placed at their canonical source indices; the stride diagnostic checks video-time monotonicity because endpoint distance alone misses stalls and reversals.}}
\label{tab:completion-summary}
\centering
\small
\resizebox{\linewidth}{!}{%
\begin{tabular}{llll}
\toprule
Design choice & \ours{} setting & Diagnostic & Evidence \\
\midrule
Window placement & Endpoint-window arrangement & Loop endpoint L1 $\downarrow$ & $6.20/8.13$ vs. $28.08$--$46.98$ for non-endpoint layouts \\
Angle mapping & source-angle-aligned sparse anchors & Anchor index error $\downarrow$ & Mean $0.06$, max $1$ frame vs. target-half mean $12$, max $24$ \\
Anchor stride & Stride $24$ & Temporal trace and frame strip & Fewer stalls and reversals than other tested strides \\
\bottomrule
\end{tabular}}
\end{table}

\begin{table}[t]
\centering
\small
\resizebox{\linewidth}{!}{%
\begin{tabular}{llrrr}
\toprule
Prompt & Variant & Loop endpoint L1 $\downarrow$ & Mean index error $\downarrow$ & Max index error $\downarrow$ \\
\midrule
\emph{Street car} & Source-video target-half & 11.97 & 12.00 & 24.00 \\
\emph{Street car} & Canonical-orbit target-half & 9.15 & 12.00 & 24.00 \\
\emph{Street car} & Source-angle aligned & 6.20 & 0.11 & 1.00 \\
\emph{Rabbit on pancake} & Source-video target-half & 11.48 & 12.00 & 24.00 \\
\emph{Rabbit on pancake} & Canonical-orbit target-half & 6.96 & 12.00 & 24.00 \\
\emph{Rabbit on pancake} & Source-angle aligned & 8.13 & 0.00 & 0.00 \\
\bottomrule
\end{tabular}}
\caption{\taskcompapp{Angle-alignment diagnostics for sparse anchors. Loop endpoint L1 measures generated canonical $I_{360}$ versus $I_0$, while index error measures whether each sparse keyframe is placed at the canonical-orbit index of its source view. A control can close the endpoint while still placing anchors at the wrong angles, so index error is the direct alignment diagnostic.}}
\label{tab:coverage-completion-angle-metrics}
\end{table}

\begin{table}[t]
\centering
\small
\resizebox{0.8\linewidth}{!}{%
\begin{tabular}{llrrrr}
\toprule
Prompt & Stride & Abs. turns $\downarrow$ & Net turns & Stall frac. $\downarrow$ & Reversals $\downarrow$ \\
\midrule
\emph{Street car} & 1 & 2.02 & -1.00 & 0.62 & 2 \\
\emph{Street car} & 2 & 2.10 & 1.00 & 0.59 & 3 \\
\emph{Street car} & 4 & 0.99 & 0.99 & 0.51 & 0 \\
\emph{Street car} & 8 & 1.98 & 0.00 & 0.53 & 1 \\
\emph{Street car} & 16 & 0.99 & 0.99 & 0.12 & 0 \\
\emph{Street car} & 24 & 0.99 & 0.99 & 0.02 & 0 \\
\emph{Street car} & 32 & 0.99 & 0.99 & 0.14 & 0 \\
\emph{Rabbit on pancake} & 1 & 0.92 & 0.00 & 0.75 & 2 \\
\emph{Rabbit on pancake} & 2 & 1.02 & 1.00 & 0.71 & 1 \\
\emph{Rabbit on pancake} & 4 & 1.00 & 1.00 & 0.65 & 0 \\
\emph{Rabbit on pancake} & 8 & 2.36 & -0.01 & 0.57 & 9 \\
\emph{Rabbit on pancake} & 16 & 0.99 & 0.99 & 0.30 & 0 \\
\emph{Rabbit on pancake} & 24 & 0.99 & 0.99 & 0.02 & 0 \\
\emph{Rabbit on pancake} & 32 & 0.99 & 0.99 & 0.26 & 0 \\
\bottomrule
\end{tabular}}
\caption{\taskcompapp{Temporal trace diagnostics for the anchor-stride sweep in Figure~\ref{fig:app-stride}. Absolute turns sum circular reference-frame movement through the generated sequence; stall fraction counts adjacent generated frames with less than one reference-frame step; reversals count sign changes in non-trivial movement. Non-default strides can score well under endpoint L1 while producing non-monotone or stalled video motion.}}
\label{tab:coverage-completion-stride-temporal-metrics}
\end{table}

\section{Evaluation Protocol and Additional Quantitative Diagnostics}
\label{app:additional-quant}

\taskappcite{This section expands the main coverage-first evaluation table. It separates local-view diagnostics, coverage-qualified lower-tail quality, and $\Rzero$/$\Rone$ support-stratified diagnostics so that each automatic score is read under the correct orbit protocol.}

\subsection{Frozen 300-Prompt Benchmark Set and Secondary Diagnostics}

\taskorbit{Table~\ref{tab:t3bench-main} reports diagnostics on the same 300-prompt \tthreebench{}-derived audit set. We treat these scores as diagnostics rather than as a complete full-orbit task score: CLIP and ImageReward measure prompt alignment and single-view visual quality, while adjacent LPIPS and adjacent DINO measure local smoothness over sampled views. The main coverage metrics provide the complementary full-orbit protocol.}

\taskdnear{The closed-success column in Table~\ref{tab:closed-orbit-coverage-audit} is deliberately not a hidden feature-threshold test. The reported audit has two auditable geometric requirements: sufficiently large span and monotone orbit progress. Feature loop metrics remain useful for diagnosing visual loop mismatch, but they are not used to promote or demote rows in the coverage table unless they are explicitly reported alongside the threshold. This avoids claiming a DINO-gated success criterion without exposing the measured feature distances.}

\begin{table}[t]
\caption{\taskorbit{Quantitative diagnostics on the frozen 300-prompt \tthreebench{}-derived audit set, pooling the single-object, object-in-context, and multi-object/action categories. Every row uses the same prompts and reports mean $\pm$ standard deviation over prompts.}}
\label{tab:t3bench-main}
\centering
\scriptsize
\resizebox{\linewidth}{!}{%
\begin{tabular}{lrrrrr}
\toprule
Method & N & CLIP $\uparrow$ & ImageReward $\uparrow$ & Adj. LPIPS $\downarrow$ & Adj. DINO $\downarrow$ \\
\midrule
DreamFusion~\citep{poole2022dreamfusion} & 300 & $19.76{\pm}3.70$ & $10.05{\pm}9.12$ & $0.085{\pm}0.048$ & $0.072{\pm}0.032$ \\
Latent-NeRF~\citep{metzer2023latent} & 300 & $22.13{\pm}2.97$ & $18.11{\pm}13.04$ & $0.095{\pm}0.052$ & $0.079{\pm}0.034$ \\
Magic3D~\citep{lin2023magic3d} & 300 & $18.26{\pm}4.04$ & $8.37{\pm}8.11$ & $0.088{\pm}0.047$ & $0.074{\pm}0.031$ \\
SJC~\citep{wang2023score} & 300 & $20.78{\pm}2.84$ & $12.14{\pm}9.09$ & $0.128{\pm}0.049$ & $0.113{\pm}0.039$ \\
SV3D~\citep{voleti2024sv3d} & 300 & $23.06{\pm}3.71$ & $24.13{\pm}19.66$ & $0.193{\pm}0.068$ & $0.140{\pm}0.054$ \\
VideoMV~\citep{zuo2024videomv} & 300 & $23.86{\pm}3.10$ & $30.52{\pm}19.48$ & $0.211{\pm}0.083$ & $0.126{\pm}0.043$ \\
ViewCrafter~\citep{yu2024viewcrafter} & 300 & $25.09{\pm}3.07$ & $43.68{\pm}21.15$ & $0.066{\pm}0.041$ & $0.023{\pm}0.025$ \\
MedianGS-only $\Rzero$ (ours) & 300 & $22.14{\pm}3.19$ & $23.72{\pm}14.38$ & $0.290{\pm}0.067$ & $0.156{\pm}0.053$ \\
\ours{} $\Rone$ (ours) & 300 & $23.83{\pm}3.37$ & $28.52{\pm}18.06$ & $0.268{\pm}0.060$ & $0.134{\pm}0.054$ \\
\ours{} cleanup (ours) & 300 & $23.00{\pm}3.40$ & $25.49{\pm}17.27$ & $0.267{\pm}0.051$ & $0.131{\pm}0.037$ \\
\bottomrule
\end{tabular}}
\end{table}

\subsection{\taskview{ViewCrafter Local-Trajectory Audit}}
\label{app:viewcrafter-local-diagnostic}

\taskview{ViewCrafter is evaluated as a local novel-view baseline, so we audit its camera range before interpreting any visual-quality score. For the benchmark outputs, ViewCrafter exposes a declared 25-frame camera path through its single-view trajectory interface. We reproduce the same interpolation used to render its conditioning sequence: yaw controls are interpolated to 25 frames, interpreted as horizontal azimuths, wrapped onto the circle, and scored with Eq.~\ref{eq:span}. Monotone progress is computed from the signed adjacent azimuth increments by choosing the dominant direction and reporting the fraction of steps that follow it.}

\taskview{This produces a measured horizontal span of $18.88^\circ$ with a $341.12^\circ$ uncovered circular gap. The path also returns near the input view, so its dominant-direction progress is $0.54$ rather than a monotone orbit. Under the same closed-orbit success rule used in the main table, ViewCrafter is therefore not coverage-qualified.}

\begin{table}[t]
\caption{\taskview{ViewCrafter trajectory coverage audit on the frozen 300-prompt \tthreebench{}-derived audit set. The measured azimuth range comes from the declared local camera path used for its outputs, not from frame count or ImageReward.}}
\label{tab:viewcrafter-trajectory-audit}
\centering
\small
\resizebox{\linewidth}{!}{%
\begin{tabular}{lrrrrrrr}
\toprule
Method & N & Frames & Azimuth span $\uparrow$ & Max gap $\downarrow$ & Monotone progress $\uparrow$ & Closed success $\uparrow$ & Coverage-qualified \\
\midrule
ViewCrafter~\citep{yu2024viewcrafter} & 300 & 25 & 18.88 & 341.12 & 0.54 & 0\% & No \\
\bottomrule
\end{tabular}}
\end{table}

\begin{table}[t]
\caption{\taskview{Local-view quality diagnostic, not a full-orbit ranking. ViewCrafter's high ImageReward reflects a short local novel-view trajectory; OrbitForge is scored on a full canonical orbit with originally unsupported views.}}
\label{tab:viewcrafter-local-quality}
\centering
\small
\resizebox{\linewidth}{!}{%
\begin{tabular}{llrrrr}
\toprule
Method & Trajectory type & Coverage-qualified & N & Q10 ImageReward $\uparrow$ & Mean ImageReward $\uparrow$ \\
\midrule
ViewCrafter~\citep{yu2024viewcrafter} & local novel-view video & No & 300 & $37.50{\pm}22.07$ & 43.68 \\
\ours{} $\Rone$ & full canonical orbit & Yes & 300 & $16.79{\pm}16.10$ & 28.52 \\
\bottomrule
\end{tabular}}
\end{table}

\subsection{\taskanchor{Anchor-to-View DINOv2 Appearance Consistency}}
\label{app:anchor-appearance-consistency}

\taskanchor{Table~\ref{tab:anchor-appearance-consistency} reports an anchor-to-view DINOv2 appearance-consistency diagnostic on the same frozen 300 prompts and sampled views used for lower-tail quality. For each method and prompt, the first sampled view is treated as an appearance anchor, the remaining sampled views are encoded with DINOv2-small, cosine similarities are computed against the anchor, and the per-prompt 10th percentile is averaged over prompts. This is still a diagnostic rather than an identity judgment, but it is a stronger semantic appearance check than raw color statistics.}

\begin{table}[t]
\centering
\small
\resizebox{0.82\linewidth}{!}{%
\begin{tabular}{lrrrr}
\toprule
Method & N & Views compared & Anchor-DINO Q10 $\uparrow$ & Anchor-DINO mean $\uparrow$ \\
\midrule
SV3D~\citep{voleti2024sv3d} & 300 & 11 & $0.510{\pm}0.209$ & 0.697 \\
VideoMV~\citep{zuo2024videomv} & 300 & 11 & $0.568{\pm}0.180$ & 0.683 \\
MedianGS-only $\Rzero$ (ours) & 300 & 11 & $0.311{\pm}0.207$ & 0.576 \\
\ours{} $\Rone$ (ours) & 300 & 11 & $0.564{\pm}0.166$ & 0.717 \\
\bottomrule
\end{tabular}}
\caption{\taskanchor{Anchor-to-view DINOv2 appearance-consistency diagnostic. The metric compares sampled orbit views to the first sampled view using DINOv2-small image embeddings. It is not an identity score, but it checks whether weak orbit views remain semantically close to the anchor view. \ours{} substantially improves the lower-tail DINO similarity over MedianGS-only $\Rzero$, consistent with completion reducing severe unsupported-view drift.}}
\label{tab:anchor-appearance-consistency}
\end{table}

\subsection{Coverage-Qualified Lower-Tail Quality}

\begin{figure}[t]
\centering
\includegraphics[width=0.82\linewidth]{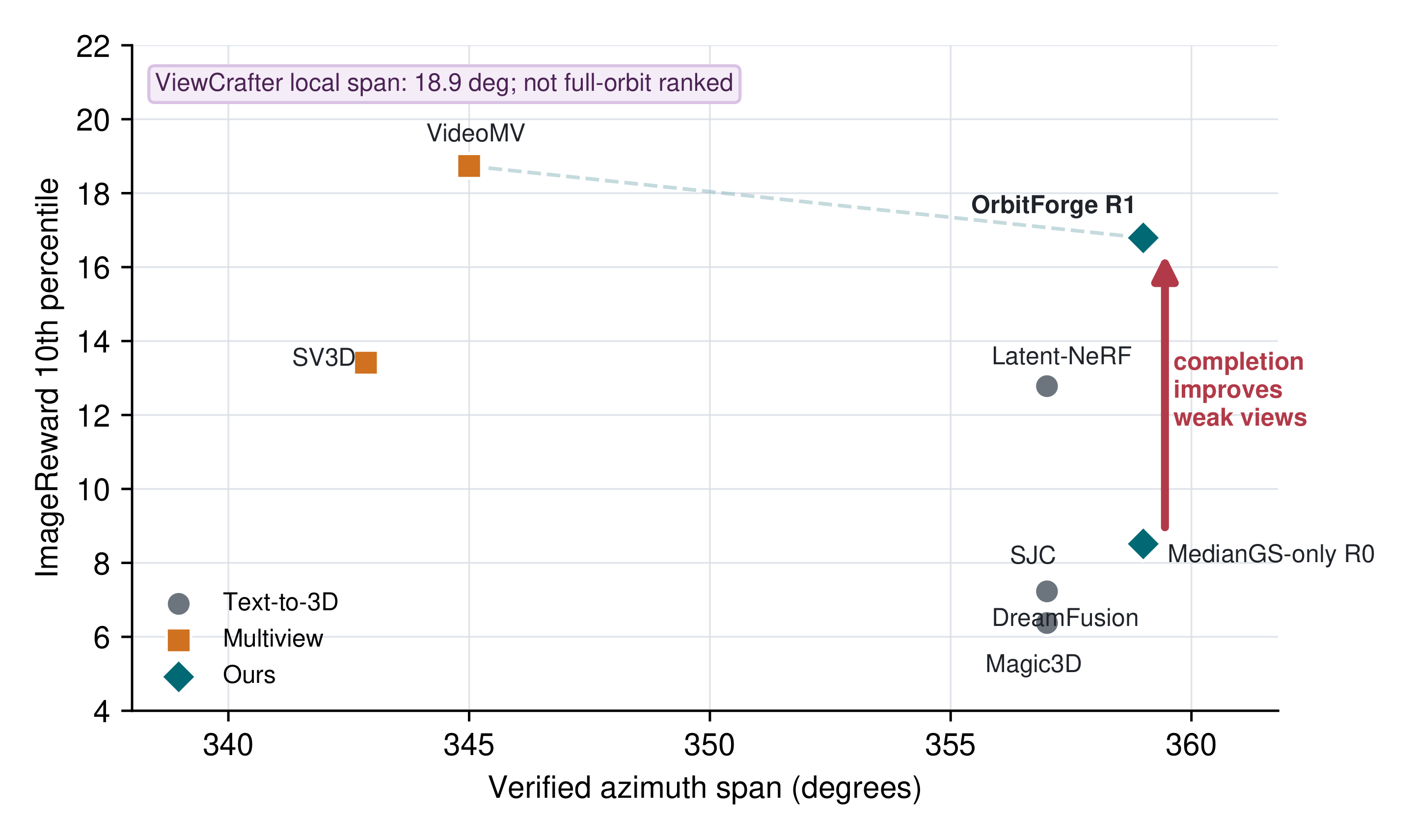}
\caption{\taskview{Coverage-quality Pareto audit for coverage-qualified full-orbit outputs. The horizontal axis is measured median azimuth span and the vertical axis is prompt-averaged 10th-percentile ImageReward over sampled orbit views. The emphasized arrow shows that coverage-aware completion lifts the weak-view tail from MedianGS-only $\Rzero$ to \ours{} $\Rone$ while preserving the measured canonical-orbit span. ViewCrafter's local trajectory span is measured in Table~\ref{tab:viewcrafter-trajectory-audit}, but it is not ranked on this full-orbit quality frontier.}}
\label{fig:coverage-pareto}
\end{figure}

\begin{table}[t]
\caption{\taskview{Coverage-qualified lower-tail visual-quality table underlying the main-paper quality panel. Q10 is the 10th percentile of per-view ImageReward over 12 uniformly sampled views for each prompt, then averaged over prompts. ViewCrafter is excluded from this ranking because Appendix~\ref{app:viewcrafter-local-diagnostic} measures it as a local trajectory rather than a closed orbit.}}
\label{tab:lower-tail-quality-app}
\centering
\scriptsize
\resizebox{\linewidth}{!}{%
\begin{tabular}{lrrrrr}
\toprule
Method & N & Views & Q10 ImageReward $\uparrow$ & Min ImageReward $\uparrow$ & Mean ImageReward $\uparrow$ \\
\midrule
DreamFusion~\citep{poole2022dreamfusion} & 300 & 12 & $7.29{\pm}6.91$ & 6.52 & 10.05 \\
Latent-NeRF~\citep{metzer2023latent} & 300 & 12 & $12.78{\pm}10.99$ & 10.79 & 18.11 \\
Magic3D~\citep{lin2023magic3d} & 300 & 12 & $6.37{\pm}6.21$ & 5.82 & 8.37 \\
SJC~\citep{wang2023score} & 300 & 12 & $7.23{\pm}5.50$ & 6.36 & 12.14 \\
SV3D~\citep{voleti2024sv3d} & 300 & 12 & $13.41{\pm}16.97$ & 10.83 & 24.13 \\
VideoMV~\citep{zuo2024videomv} & 300 & 12 & $18.74{\pm}18.10$ & 14.48 & 30.52 \\
MedianGS-only $\Rzero$ (ours) & 300 & 12 & $8.51{\pm}8.98$ & 7.14 & 23.72 \\
\ours{} $\Rone$ & 300 & 12 & $16.79{\pm}16.10$ & 13.20 & 28.52 \\
\bottomrule
\end{tabular}}
\end{table}

\subsection{\texorpdfstring{$\Rzero$/$\Rone$ Support Diagnostics}{R0/R1 Support Diagnostics}}

\begin{table}[t]
\caption{\taskorbit{$\Rzero$ versus $\Rone$ input-availability diagnostics on the same 361-view canonical camera path. $\Rzero$ source availability counts target views directly reached by the source-to-canonical coverage mask. $\Rone$ completed availability means that the second reconstruction receives one input frame at every canonical camera after completion; it is not direct source-observed support.}}
\label{tab:r0-r1-support}
\centering
\small
\resizebox{\linewidth}{!}{%
\begin{tabular}{lrrrrrr}
\toprule
Prompt & \multicolumn{2}{c}{Input frames} & \multicolumn{2}{c}{Missing frames} & \multicolumn{2}{c}{Adj. L1} \\
\cmidrule(lr){2-3}\cmidrule(lr){4-5}\cmidrule(lr){6-7}
 & $\Rzero$ source & $\Rone$ completed & $\Rzero$ source & $\Rone$ completed & $\Rzero$ & $\Rone$ \\
\midrule
\emph{Street car} & 117 (32.4\%) & 361 (100.0\%) & 244 & 0 & 6.09 & 6.55 \\
\emph{Rabbit on pancake} & 94 (26.0\%) & 361 (100.0\%) & 267 & 0 & 4.30 & 6.15 \\
\emph{Quartz crystal sculpture} & 117 (32.4\%) & 361 (100.0\%) & 244 & 0 & 6.52 & 6.67 \\
\emph{Wooden chair} & 117 (32.4\%) & 361 (100.0\%) & 244 & 0 & 5.61 & 6.58 \\
\bottomrule
\end{tabular}}
\end{table}

\begin{table}[t]
\caption{\taskorbit{Full support-stratified ImageReward table. Regions are defined by the $\Rzero$ coverage mask and reused for both rows, so $\Rone$ is evaluated on the same source-supported and originally unsupported canonical bins.}}
\label{tab:support-stratified-quality-full}
\centering
\scriptsize
\resizebox{\linewidth}{!}{%
\begin{tabular}{llrrrrr}
\toprule
Method & Region & N & Views & Q10 ImageReward $\uparrow$ & Min ImageReward $\uparrow$ & Mean ImageReward $\uparrow$ \\
\midrule
MedianGS-only $\Rzero$ & source-supported bins & 300 & 4 & $30.43{\pm}21.95$ & 28.85 & 36.24 \\
MedianGS-only $\Rzero$ & originally unsupported bins & 300 & 8 & $8.07{\pm}8.39$ & 7.18 & 18.18 \\
\ours{} $\Rone$ & source-supported bins & 300 & 4 & $26.64{\pm}20.55$ & 24.94 & 32.84 \\
\ours{} $\Rone$ & originally unsupported bins & 300 & 8 & $16.36{\pm}15.55$ & 13.60 & 26.61 \\
\bottomrule
\end{tabular}}
\end{table}

\FloatBarrier

\section{Qualitative Results}
\label{app:qualitative-results}

\taskqualapp{This section expands the qualitative evidence behind Figures~\ref{fig:orbitforge-effect-main} and~\ref{fig:qualitative-main}. The goal is not to replace the quantitative coverage audit, but to make the visual failure modes behind that audit inspectable: whether a method exposes the full orbit, whether the object remains identifiable across front, side, and backside views, whether the support surface and background remain coherent, and whether high local image quality comes from a genuinely closed trajectory or from a short local path. We therefore include full per-prompt comparison grids, an OrbitForge-only gallery, a format-normalized VideoMV sanity check, additional $\Rzero$/$\Rone$ comparisons, and representative difficult cases.}

\subsection{Full Qualitative Grids}

\begin{figure}[t]
\centering
\includegraphics[width=\linewidth]{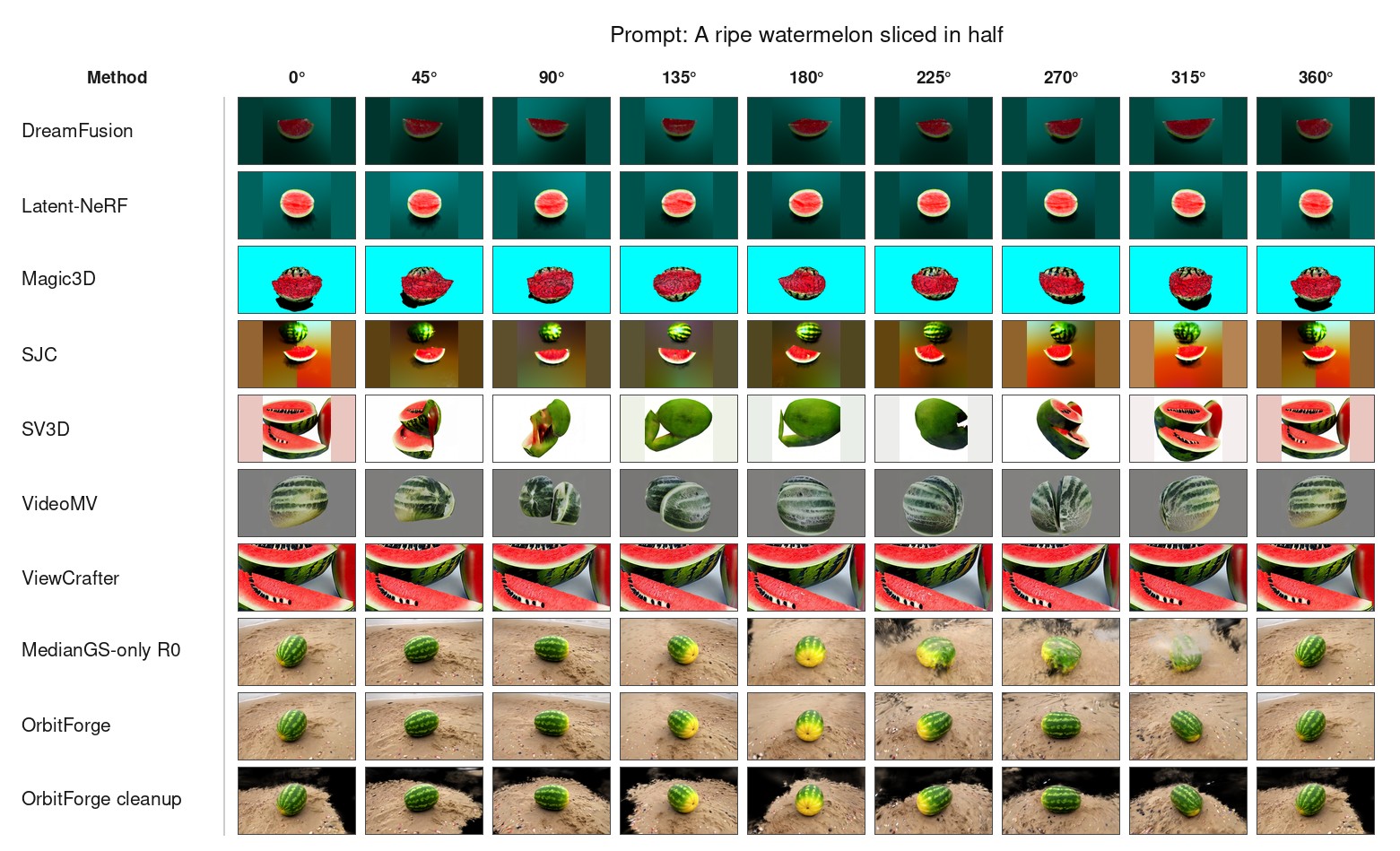}
\caption{\taskorbit{Full qualitative comparison for ``A ripe watermelon sliced in half.'' Rows show the evaluated methods and columns show uniformly sampled orbit views; this expanded grid preserves Latent-NeRF, SJC, MedianGS-only $\Rzero$, and optional cleanup rows that are omitted from the compressed main-paper comparison.}}
\label{fig:full-grid}
\end{figure}

\begin{figure}[t]
\centering
\includegraphics[width=\linewidth]{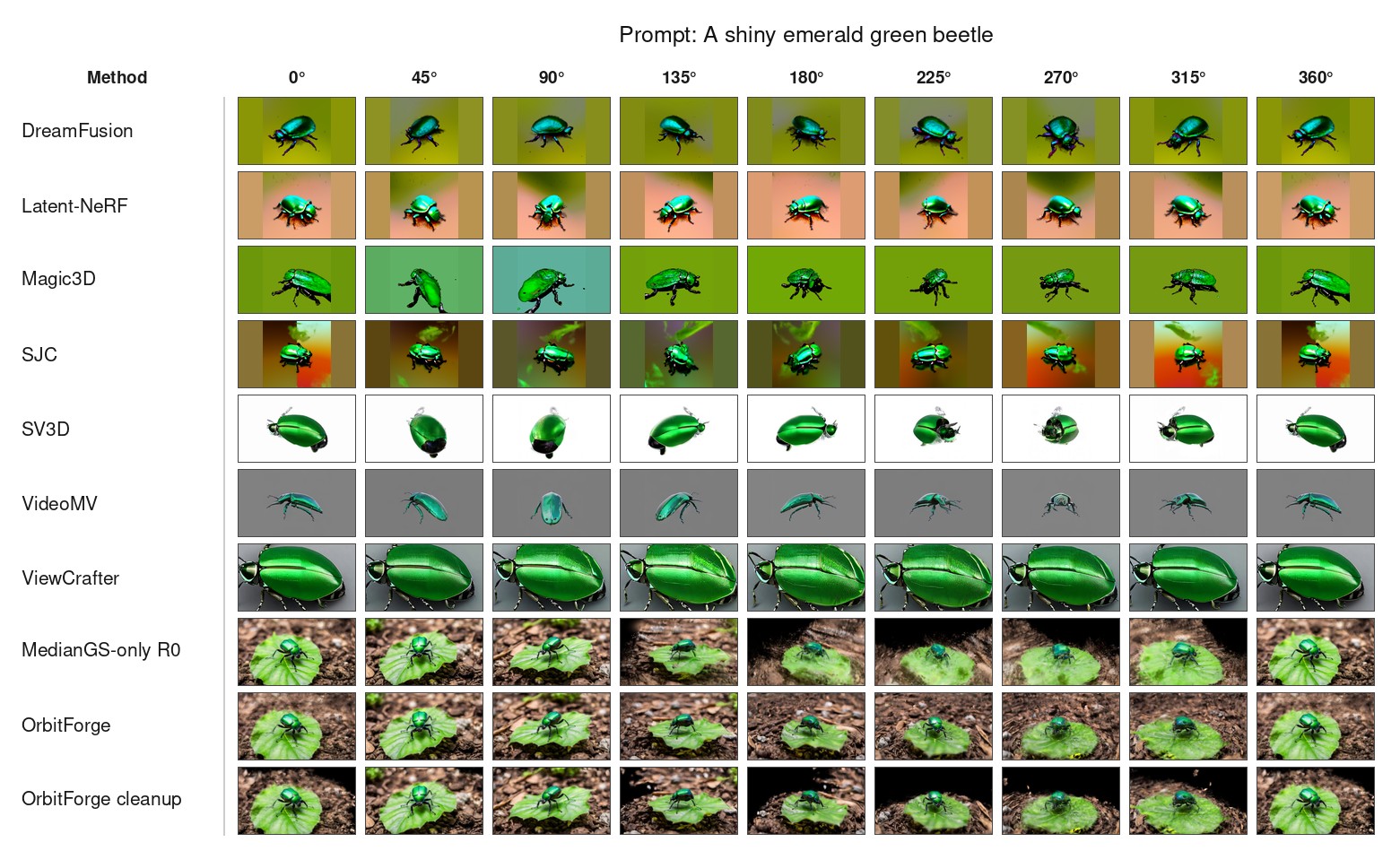}
\caption{\taskorbit{Full qualitative comparison for ``A shiny emerald green beetle.'' The grid uses the same method order and orbit-angle samples as Figure~\ref{fig:full-grid}, allowing local image quality and full-orbit behavior to be inspected together.}}
\label{fig:full-grid-beetle}
\end{figure}

\begin{figure}[t]
\centering
\includegraphics[width=\linewidth]{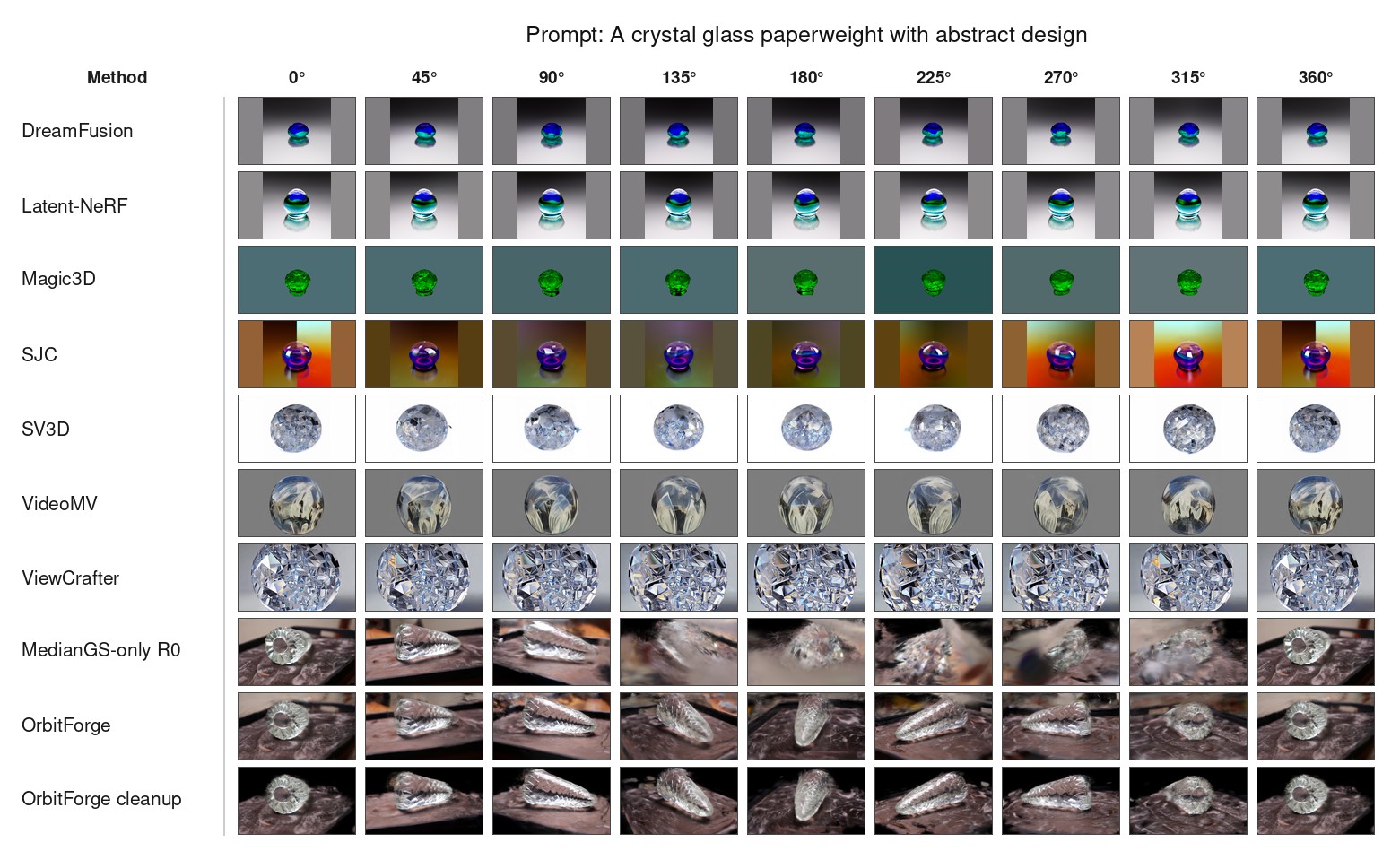}
\caption{\taskorbit{Full qualitative comparison for ``A crystal glass paperweight with abstract design.'' This prompt stresses transparent material and shape consistency across the orbit.}}
\label{fig:full-grid-paperweight}
\end{figure}

\begin{figure}[t]
\centering
\includegraphics[width=\linewidth]{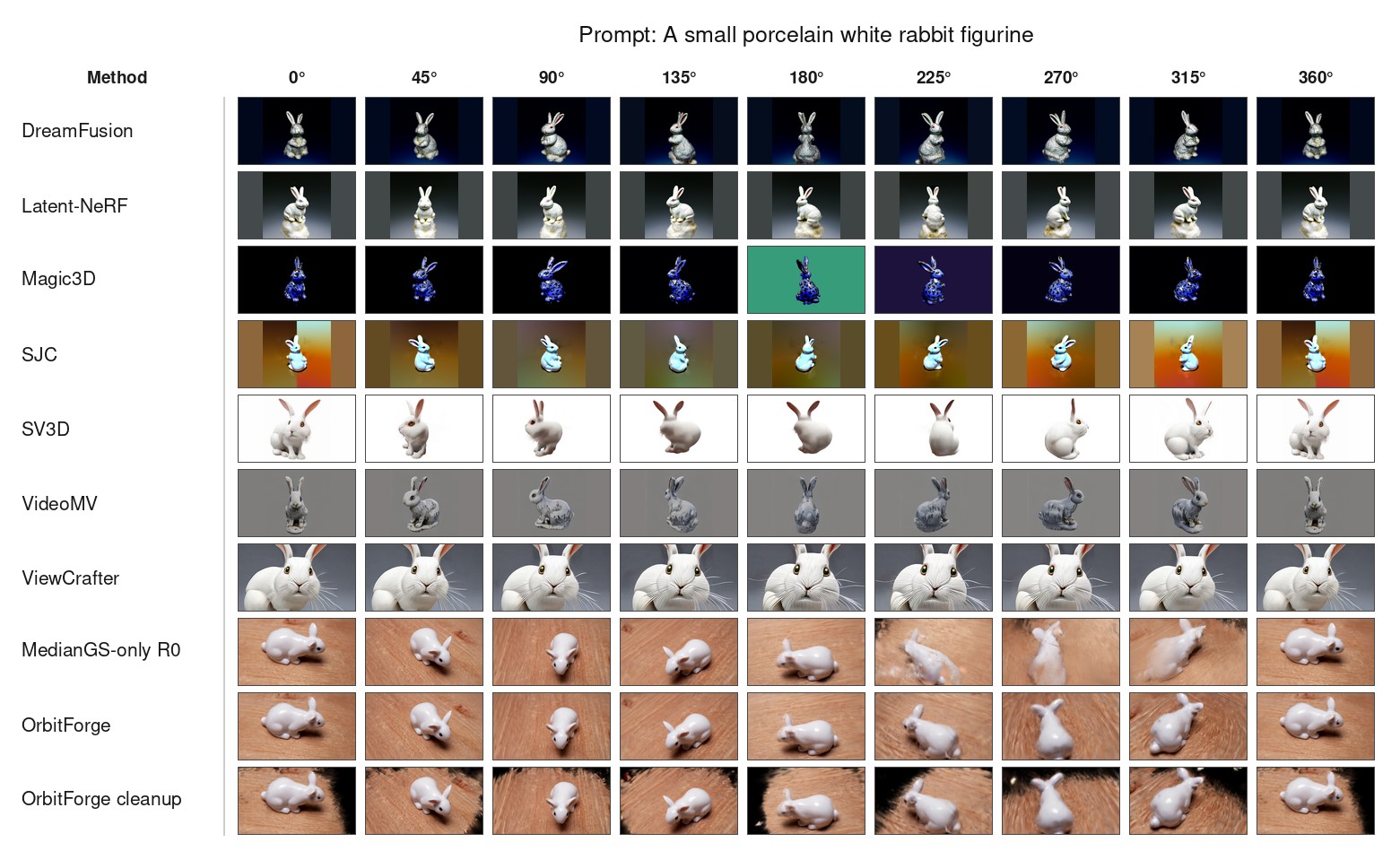}
\caption{\taskorbit{Full qualitative comparison for ``A small porcelain white rabbit figurine.'' The comparison shows how object-centric and local-view methods differ from the completed canonical-orbit reconstruction under the same sampled view range.}}
\label{fig:full-grid-rabbit-figurine}
\end{figure}

\begin{figure}[t]
\centering
\includegraphics[width=\linewidth]{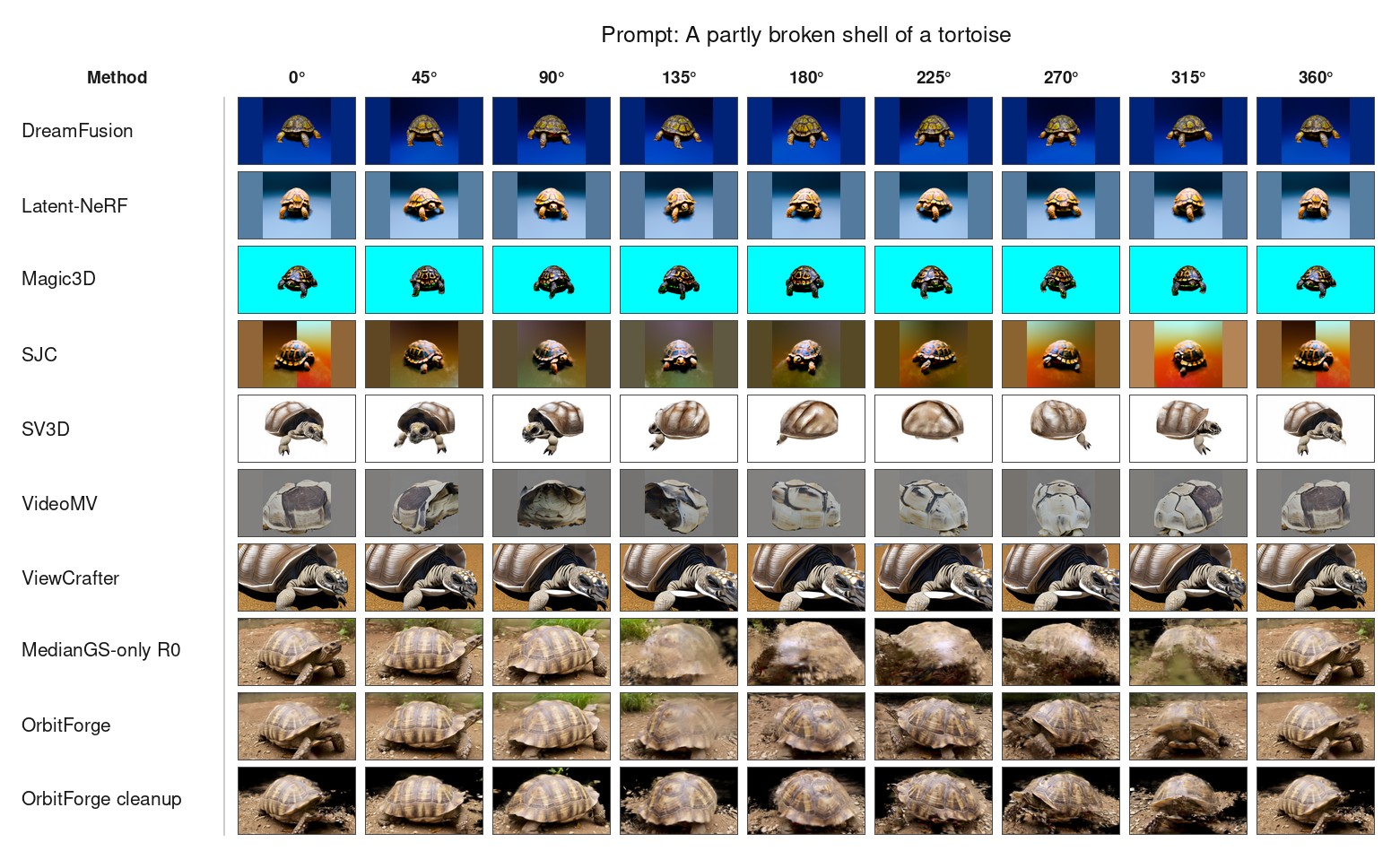}
\caption{\taskorbit{Full qualitative comparison for ``A partly broken shell of a tortoise.'' The grid includes both first-stage MedianGS-only rendering and the final OrbitForge reconstruction to expose the effect of coverage-aware completion.}}
\label{fig:full-grid-tortoise-shell}
\end{figure}

\begin{figure}[t]
\centering
\includegraphics[width=\linewidth]{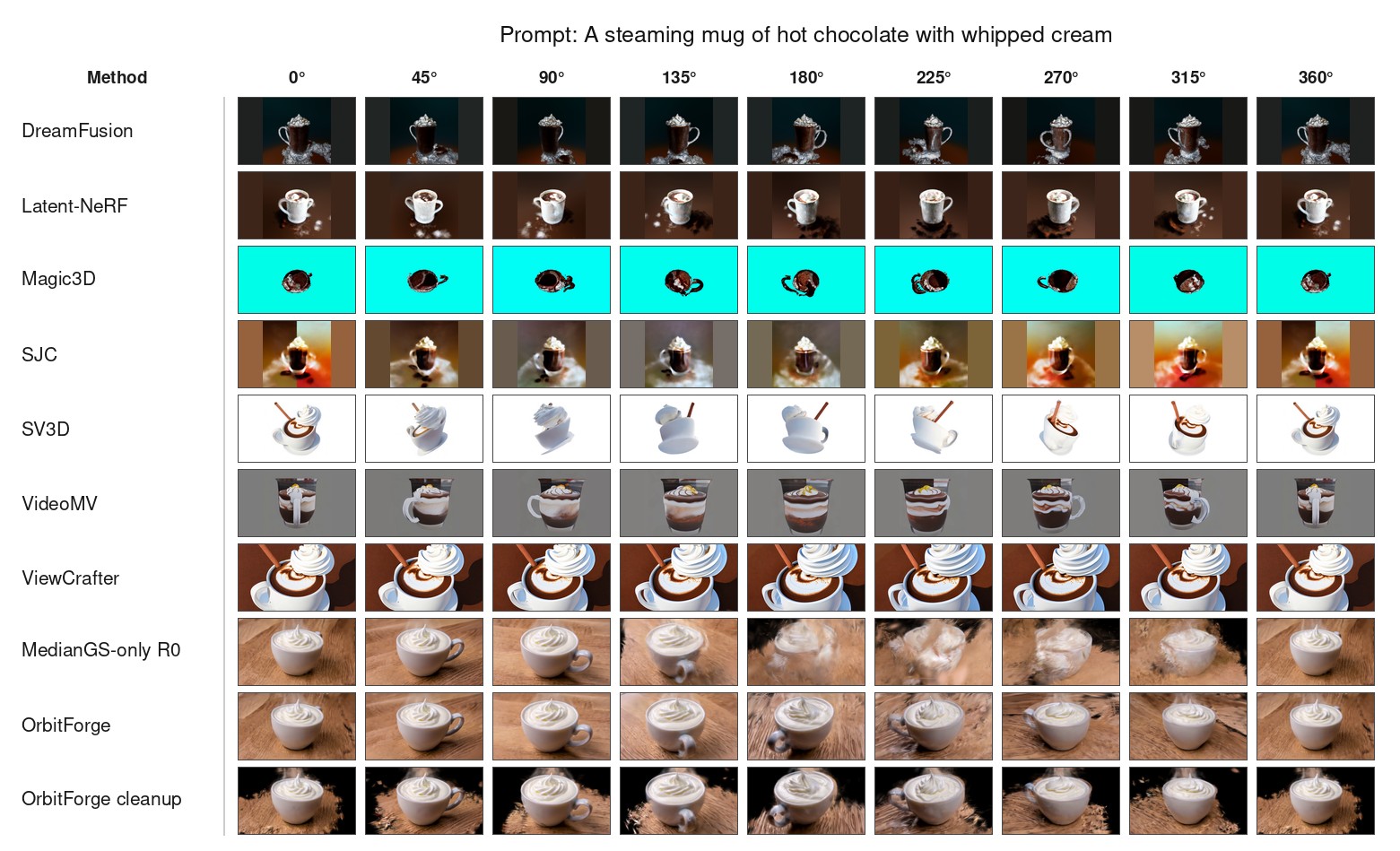}
\caption{\taskorbit{Full qualitative comparison for ``A steaming mug of hot chocolate with whipped cream.'' The prompt stresses local texture, top-view geometry, and the ability to maintain the mug and support surface around the orbit.}}
\label{fig:full-grid-hot-chocolate}
\end{figure}

\begin{figure}[t]
\centering
\includegraphics[width=\linewidth]{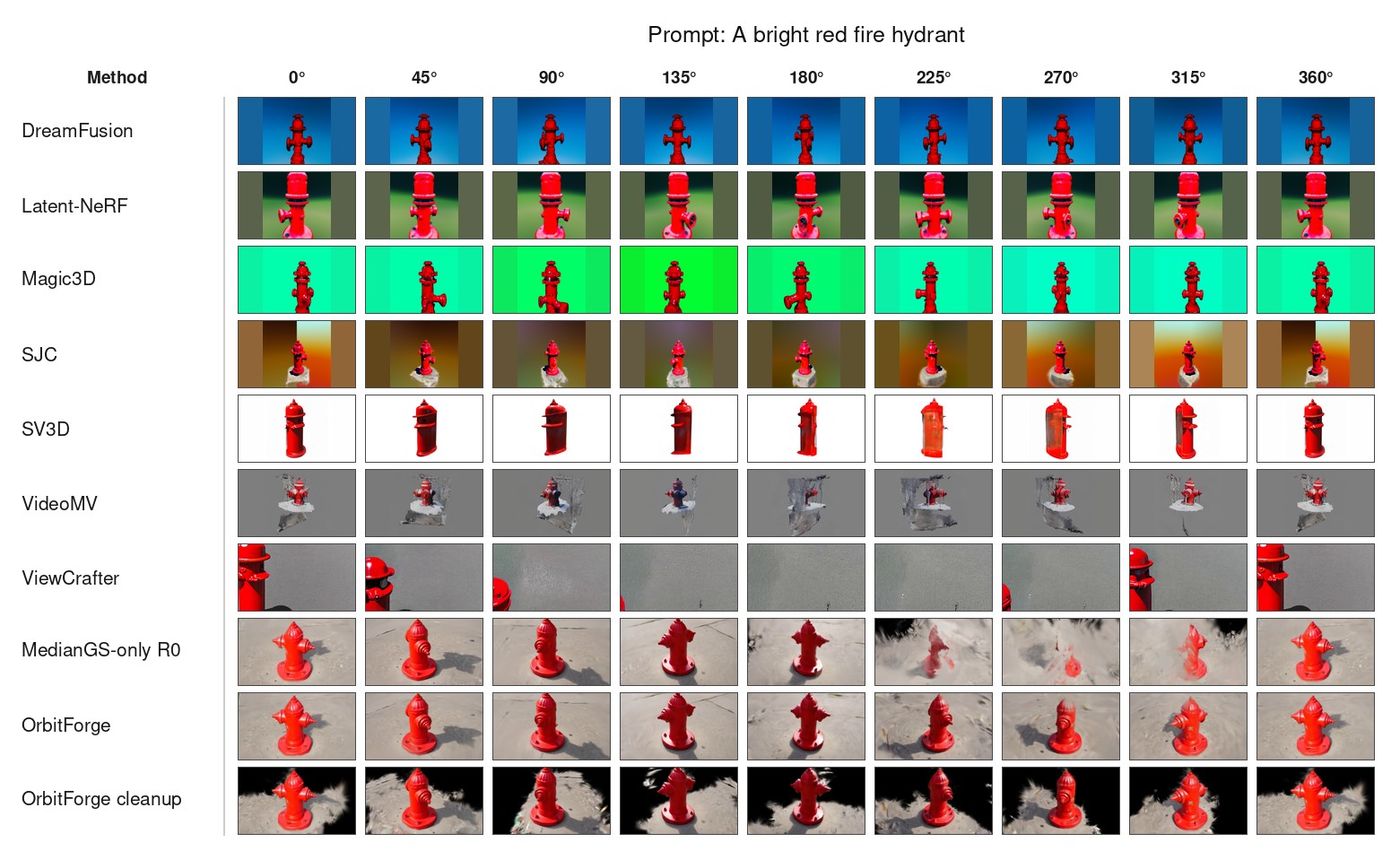}
\caption{\taskorbit{Full qualitative comparison for ``A bright red fire hydrant.'' The shared orbit-view samples make front, side, and backside behavior visible for every method.}}
\label{fig:full-grid-fire-hydrant}
\end{figure}

\begin{figure}[t]
\centering
\includegraphics[width=\linewidth]{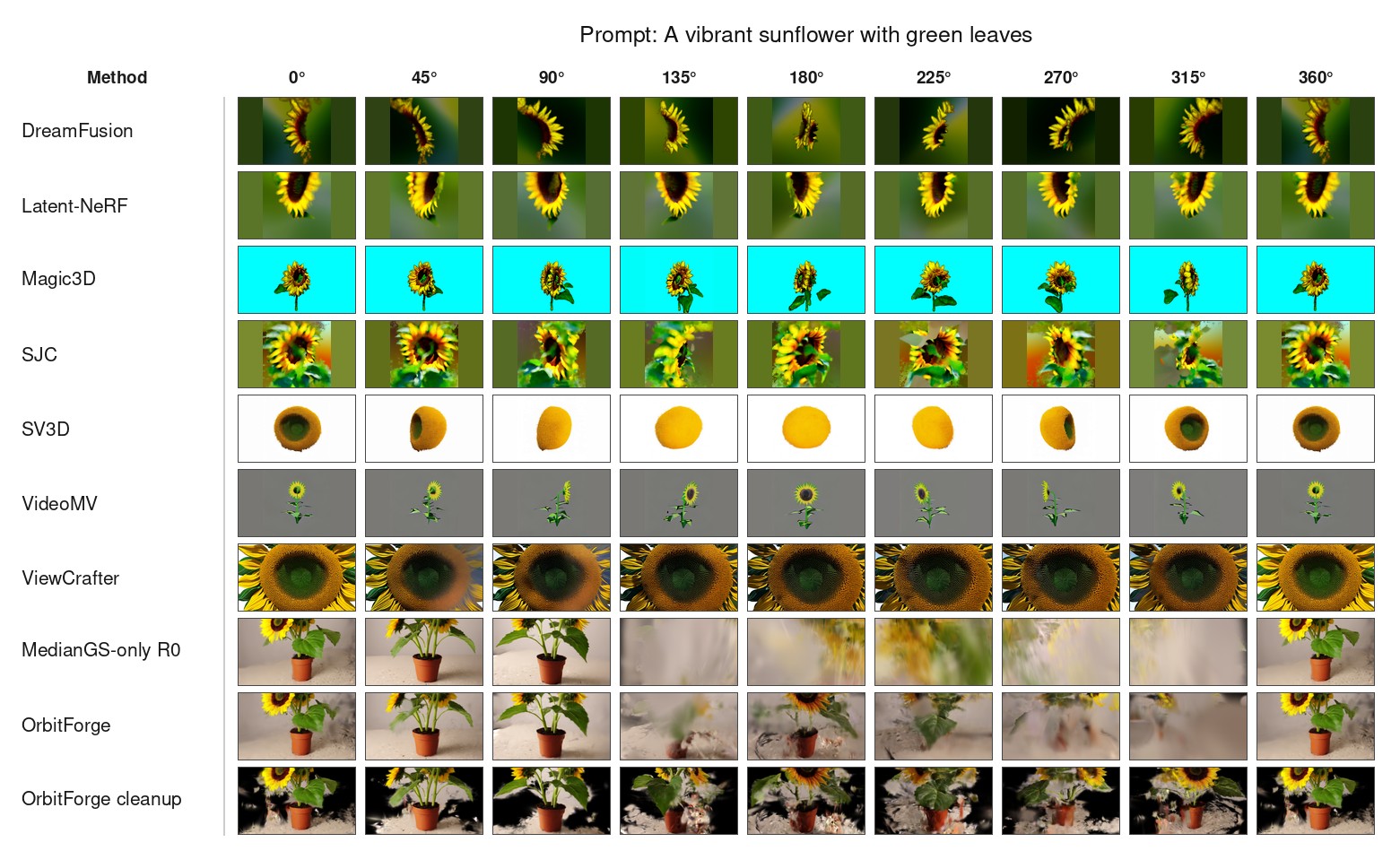}
\caption{\taskorbit{Full qualitative comparison for ``A vibrant sunflower with green leaves.'' Thin structures and repeated petal geometry make this a useful prompt for inspecting identity drift and unsupported-view hallucination.}}
\label{fig:full-grid-sunflower}
\end{figure}

\begin{figure}[t]
\centering
\includegraphics[width=\linewidth]{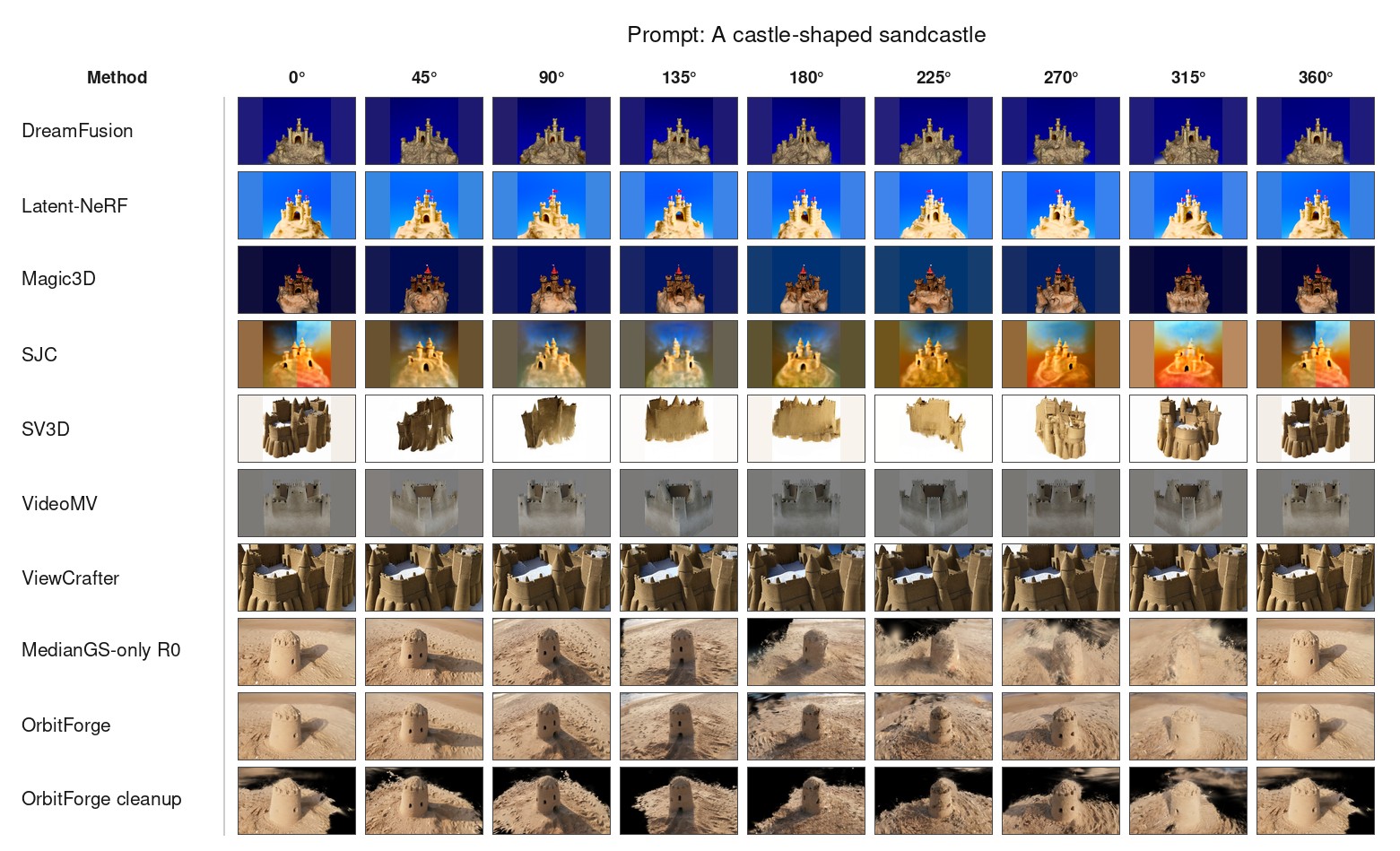}
\caption{\taskorbit{Full qualitative comparison for ``A castle-shaped sandcastle.'' The comparison highlights how methods handle structured geometry, ground contact, and scene context across the orbit.}}
\label{fig:full-grid-sandcastle}
\end{figure}

\begin{figure}[t]
\centering
\includegraphics[width=\linewidth]{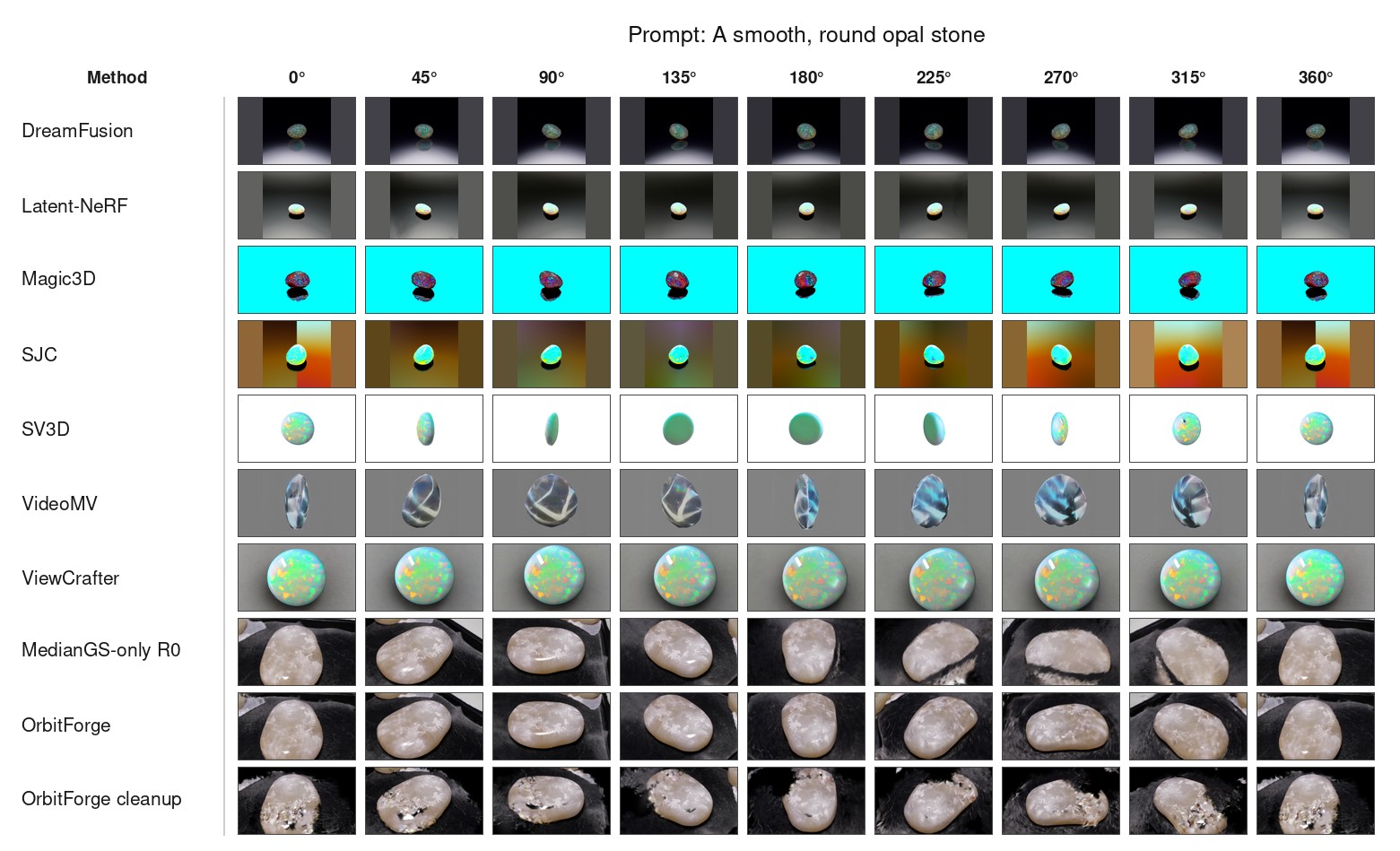}
\caption{\taskorbit{Full qualitative comparison for ``A smooth, round opal stone.'' This prompt emphasizes reflective or translucent appearance and backside consistency under full-orbit sampling.}}
\label{fig:full-grid-opal}
\end{figure}

\begin{figure}[t]
\centering
\includegraphics[width=\linewidth]{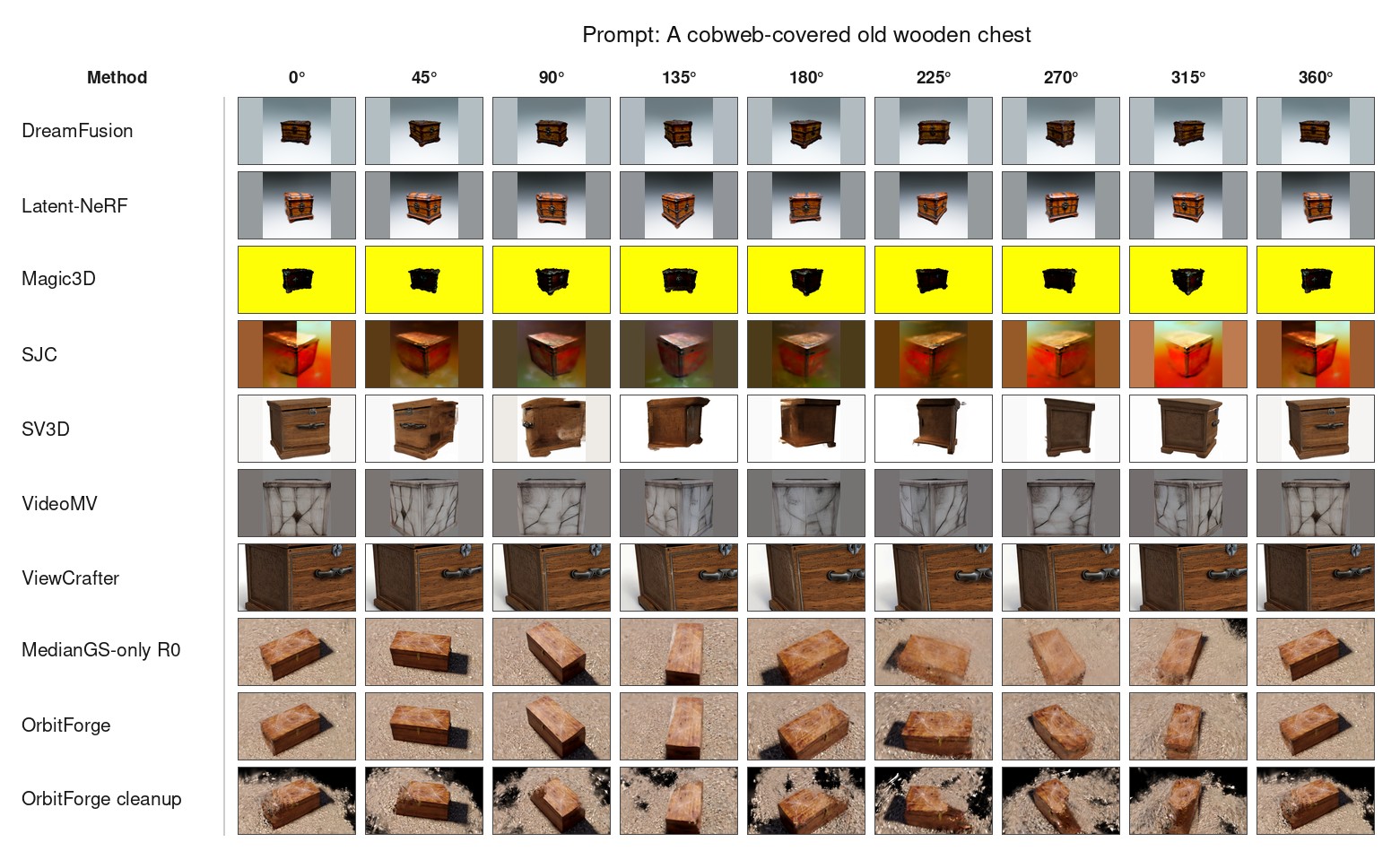}
\caption{\taskorbit{Full qualitative comparison for ``A cobweb-covered old wooden chest.'' The prompt tests box-like geometry, texture continuity, and whether the surrounding support remains coherent around the orbit.}}
\label{fig:full-grid-wooden-chest}
\end{figure}

\taskqualapp{Figures~\ref{fig:full-grid}--\ref{fig:full-grid-wooden-chest} split the full comparison into one prompt per page-scale grid. Each figure fixes one semantic target from the \tthreebench{}-derived prompt set and shows all available method variants over uniformly sampled orbit angles. This presentation is intentionally redundant with the compressed main-paper comparison: the main figure gives a fast overview, while the appendix grids let the reader check individual weak views, missing backside content, context collapse, and local-view baselines that remain visually strong without satisfying the closed-orbit protocol.}

\taskqualapp{The first five grids cover the prompts used in the broader method comparison: a sliced watermelon, beetle, glass paperweight, porcelain rabbit figurine, and broken tortoise shell. These examples stress material, shape, transparency, small-object geometry, and backside consistency. The additional six grids add hot chocolate, a fire hydrant, sunflower, sandcastle, opal stone, and wooden chest, which stress top-view structure, thin repeated parts, rigid geometry, reflective or translucent appearance, and texture continuity. Across all grids, \ours{} should be read as a scene-level closed-orbit reconstruction, whereas ViewCrafter is included as a local-view baseline and VideoMV as a specialized multiview baseline.}

\taskqualapp{These figures are qualitative audits rather than winner-take-all rankings. SDS-style methods can produce clean isolated objects but often remove scene context; specialized multiview methods can produce plausible object turntables while simplifying the surrounding environment; local view-completion methods can maintain high local fidelity over a short path. The visual question for \ours{} is different: whether the completed reconstruction preserves a usable object-environment relationship over a complete canonical orbit.}

\FloatBarrier

\begin{figure}[t]
\centering
\includegraphics[width=\linewidth,height=0.72\textheight,keepaspectratio]{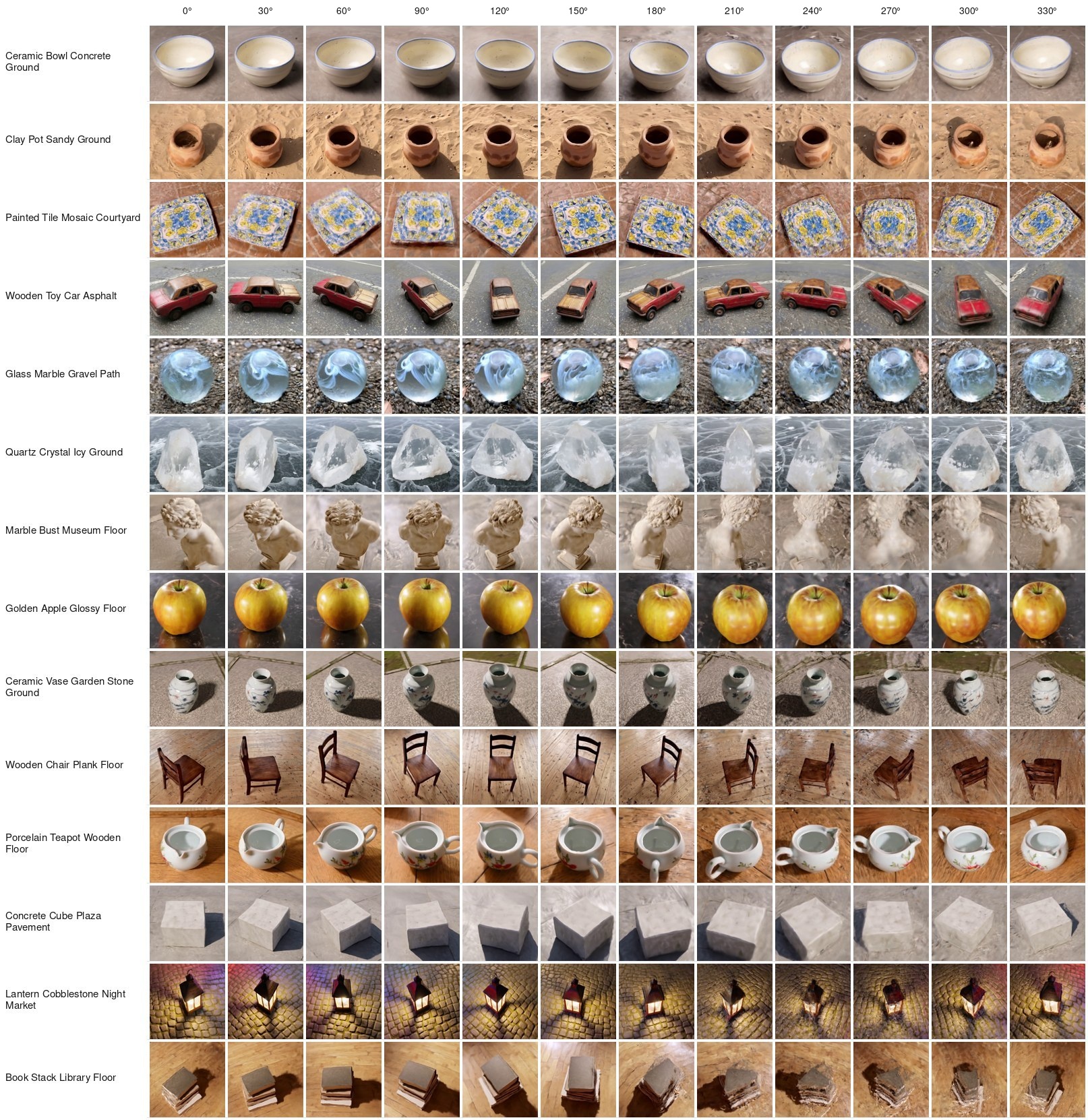}
\caption{\taskorbit{OrbitForge-only full-orbit gallery across 14 prompts. Each row samples the same canonical orbit at 30-degree intervals. The gallery shows that the method's closed-orbit behavior is not limited to the three prompts used in the main qualitative comparison.}}
\label{fig:orbitforge-gallery}
\end{figure}

\paragraph{OrbitForge-only gallery.}
\taskqualapp{Figure~\ref{fig:orbitforge-gallery} removes baselines and shows only the final \ours{} render across a broader set of prompts. Each row samples the same completed Gaussian Splatting scene at 30-degree intervals. This gallery is meant to check that the closed-orbit behavior in the main comparison is not restricted to a few selected prompts: the object should remain attached to a support surface, the surrounding context should move consistently with the orbit, and the sequence should expose both front and backside views rather than repeating a short local trajectory.}

\begin{figure}[t]
\centering
\includegraphics[width=\linewidth]{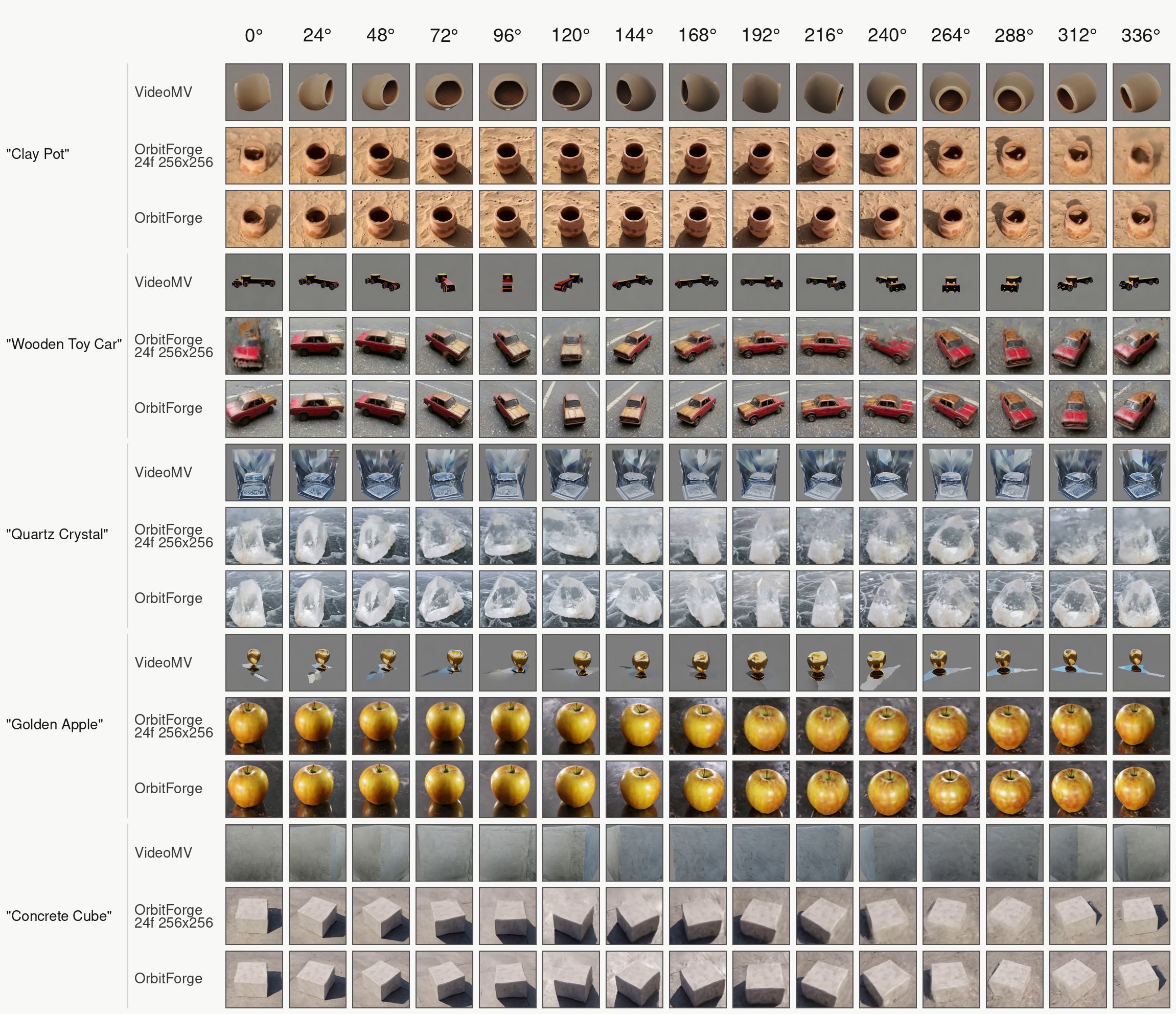}
\caption{\taskqualapp{Format-normalized qualitative sanity check against VideoMV. Each prompt compares native VideoMV, a 24-frame square OrbitForge variant, and the original OrbitForge canonical-orbit render. The middle row controls for native resolution and frame-count differences; the final row shows the full completed Gaussian Splatting scene render. The comparison is not used as a matched-camera metric benchmark because VideoMV retains its native camera path.}}
\label{fig:videomv-format-sanity}
\end{figure}

\paragraph{Format-normalized VideoMV comparison.}
\taskqualapp{Figure~\ref{fig:videomv-format-sanity} compares \ours{} with VideoMV under a format-normalized qualitative protocol. VideoMV natively produces a 24-frame, $256{\times}256$ multiview video, while the reported \ours{} output is a 361-frame, $896{\times}512$ canonical-orbit render. To separate output-format effects from method behavior, we include an intermediate OrbitForge variant with 24 square frames at $256{\times}256$. This variant samples the canonical-orbit render at 24 evenly spaced angles, converts the frames to the square low-resolution format, and reconstructs the corresponding 24-frame orbit before display.}

\taskqualapp{Each prompt block has three rows: native VideoMV, the 24-frame square OrbitForge variant, and the original OrbitForge canonical-orbit render. The columns show 15 orbit positions from \(0^\circ\) to \(336^\circ\). For VideoMV and the 24-frame OrbitForge variant, we display the nearest native frames to those angles; for the original OrbitForge row, we use the corresponding canonical-orbit frames. Because VideoMV cannot be rendered on the fitted \ours{} cameras without changing its native camera-conditioned generator, this is not a matched-camera reconstruction benchmark. It is a sanity check for whether the qualitative comparison is driven only by \ours{} having a denser, wider render. Visually, VideoMV often gives plausible object views, but it can reduce the ground-plane context or change the apparent support surface; \ours{} preserves a more continuous scene context because its views are rendered from a completed Gaussian Splatting scene.}

\begin{figure}[t]
\centering
\includegraphics[width=\linewidth]{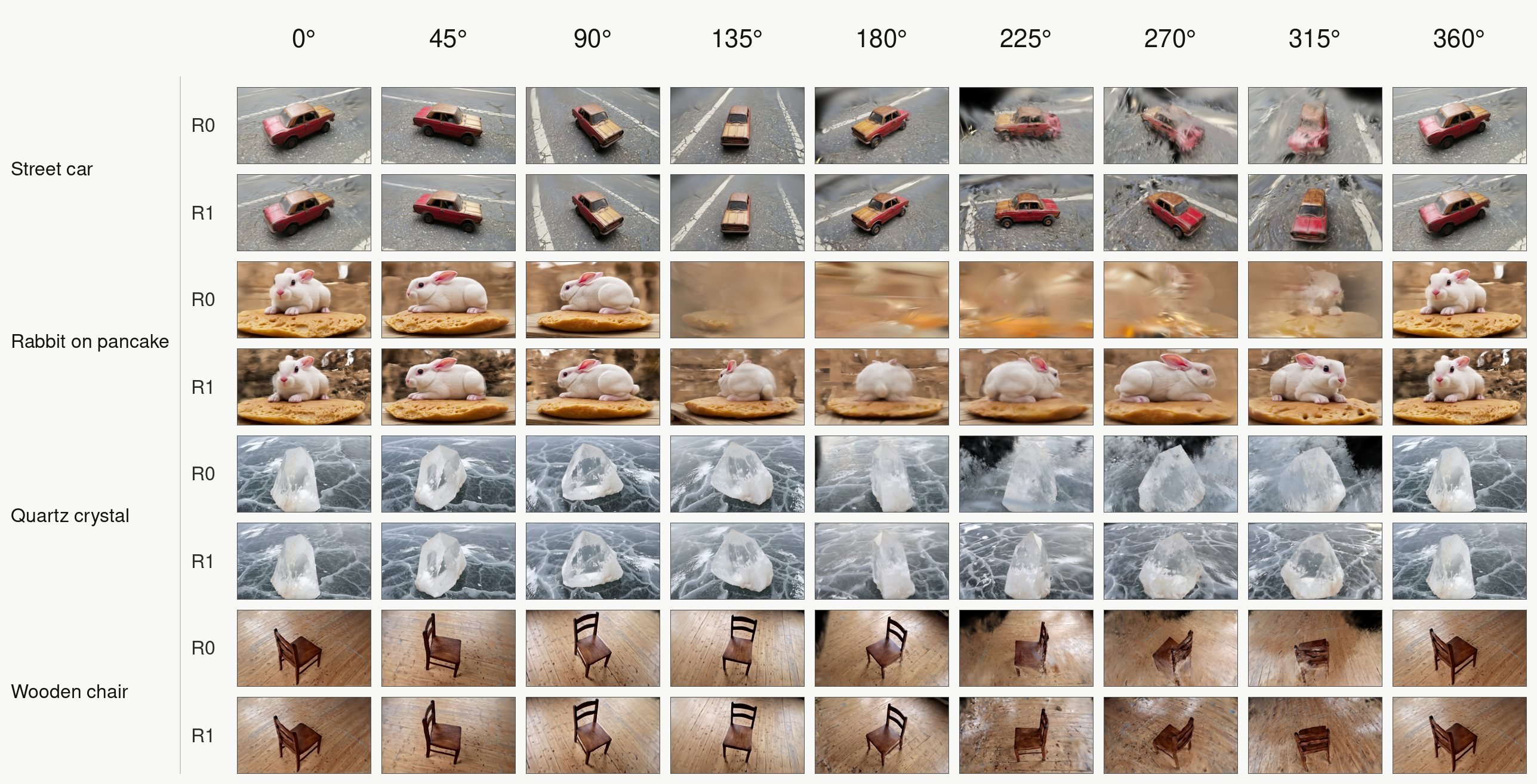}
\caption{\taskorbit{Additional $\Rzero$/$\Rone$ completion comparison on the same canonical cameras. The first reconstruction organizes the source video but remains weak in unsupported views; coverage-aware completion supplies those views before the second reconstruction.}}
\label{fig:r0-r1-app}
\end{figure}

\paragraph{Additional $\Rzero$/$\Rone$ comparison.}
\taskqualapp{Figure~\ref{fig:r0-r1-app} complements the main $\Rzero$/$\Rone$ figure by showing additional prompt blocks on the same canonical camera system. $\Rzero$ is useful because it organizes the source video into a fitted orbit and exposes which bins lack source support, but it is not expected to synthesize reliable backside content by itself. $\Rone$ is reconstructed after coverage-aware completion supplies the unsupported interval, so improvements should appear primarily in originally weak orbit regions rather than as a simple global sharpening operation.}

\begin{figure}[t]
\centering
\includegraphics[width=\linewidth]{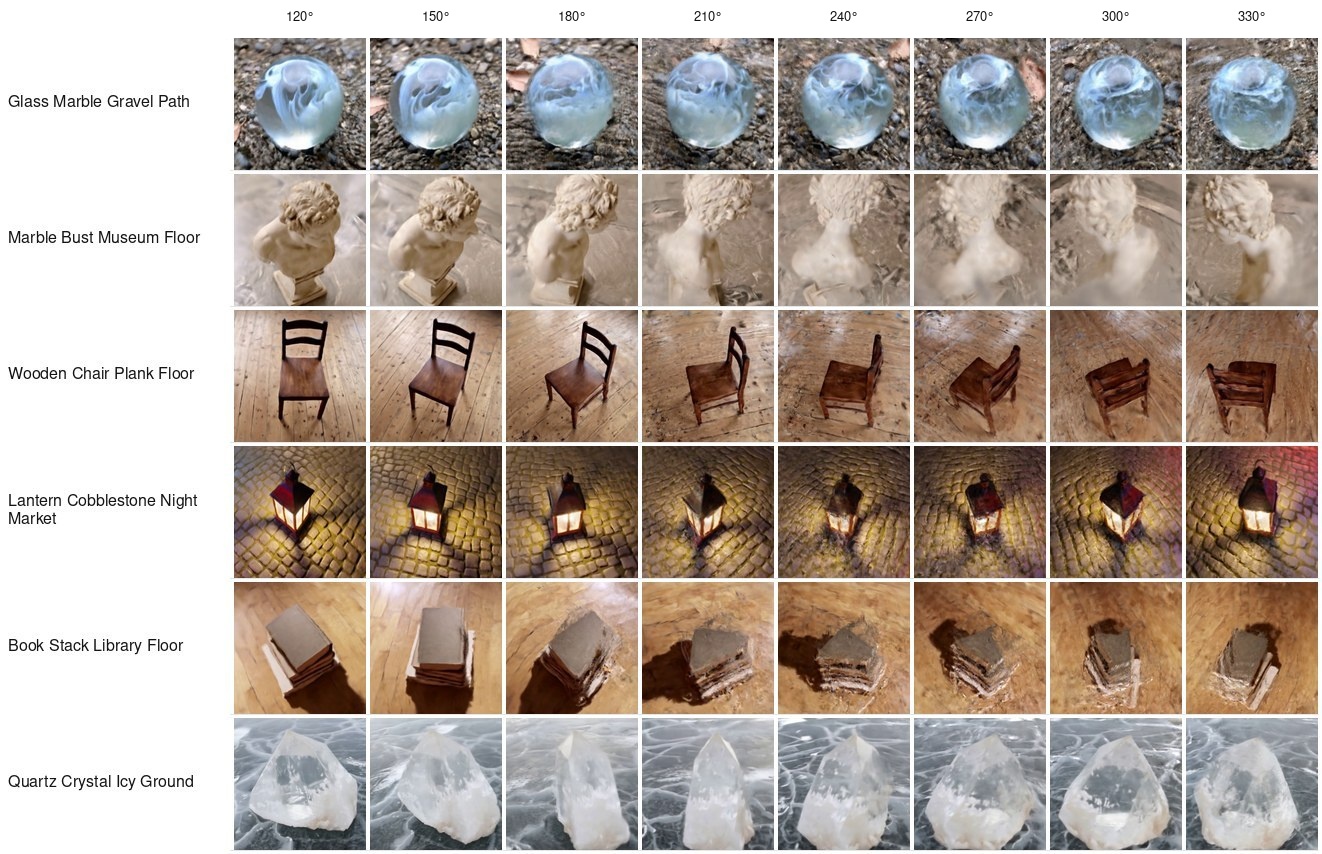}
\caption{\taskorbit{Representative difficult views. Transparent or reflective objects can become overly smooth, thin structures may bend or blur, and completion-supplied backside content can drift when the source arc provides weak structural evidence. These failures occur when the completion prior supplies unsupported content that the second reconstruction then solidifies.}}
\label{fig:failure-modes}
\end{figure}

\paragraph{Representative difficult cases.}
\taskqualapp{Figure~\ref{fig:failure-modes} shows difficult views that define the remaining limitations of the current method. These examples are included to clarify the boundary of the claim: completion-supplied backside content can drift when the source arc gives weak evidence, transparent or reflective objects can become overly smooth, thin structures can bend or blur, and the second reconstruction can solidify errors introduced by the completion prior.}

\FloatBarrier

\begin{figure}[t]
\centering
\includegraphics[width=\linewidth]{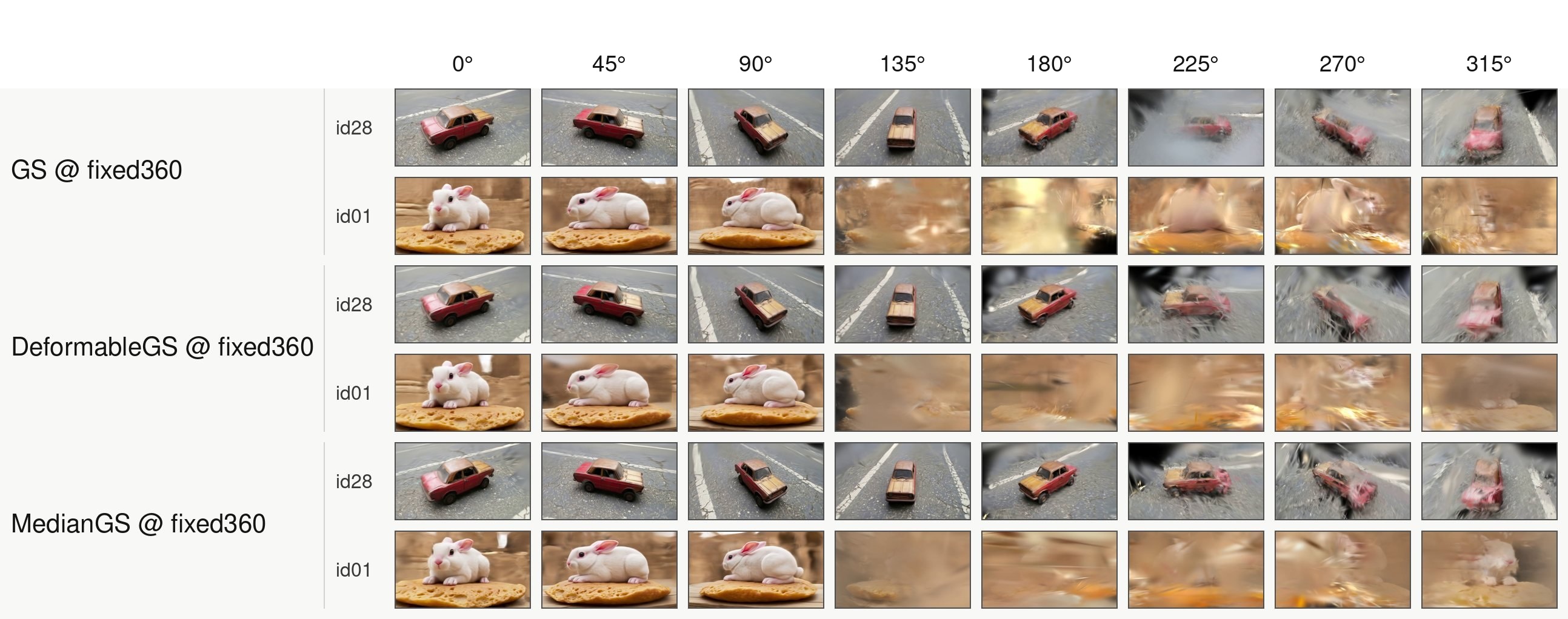}
\caption{\taskorbit{Source-video reconstruction ablation on the same canonical orbit. Static and frame-indexed reconstructions can fit observed views but carry background drag and pseudo-dynamic texture changes into unobserved angles. MedianGS is used as the robust first-stage static proxy before coverage completion.}}
\label{fig:first-reconstruction-app}
\end{figure}

\section{Reconstruction and Staticization Ablations}
\label{app:reconstruction-staticization}

\taskmedapp{This section expands the MedianGS evidence behind Section~3.3. The goal is not to claim that the first reconstruction is already a complete 3D result. Instead, the ablation asks which first-stage representation gives the most useful structural condition for coverage-aware completion. We use two diagnostic prompts, \emph{Street car} and \emph{Rabbit on pancake}, because they expose different failure modes: a rigid object on a structured ground plane and a soft subject with high-contrast local details. All variants are rendered on the same fitted 361-view canonical orbit, so differences come from the reconstruction/staticization choice rather than from camera sampling.}

\paragraph{First-stage reconstruction representation.}
The first experiment compares three ways of converting the source video into a canonical-orbit signal. The static 3DGS control fits one time-invariant Gaussian Splatting scene directly to the generated video. The frame-indexed DeformableGS control fits the same source video with a deformation field and renders the orbit using time-indexed deformation. MedianGS uses the DeformableGS fit during optimization, but removes frame-indexed deformation at render time by aggregating deformation offsets into a single static proxy. The purpose is to separate three effects: whether the representation can fit the observed source frames, whether it remains stable when rendered outside the source-covered arc, and whether it produces a condition sequence suitable for the later completion stage.

\taskmedapp{Figure~\ref{fig:first-reconstruction-app} shows the same canonical angles for the three first-stage representations. On \emph{Street car}, the direct static and frame-indexed reconstructions can preserve recognizable observed views, but later orbit positions show road dragging, car-body smear, and local dark borders. On \emph{Rabbit on pancake}, all variants become less reliable after leaving the observed arc, but MedianGS gives the most stable known-arc and transition-region condition. This is the intended role of $\Rzero$: it organizes source evidence and exposes the missing orbit interval, but it is not expected to solve unsupported backside views by itself.}

\begin{figure}[t]
\centering
\includegraphics[width=\linewidth]{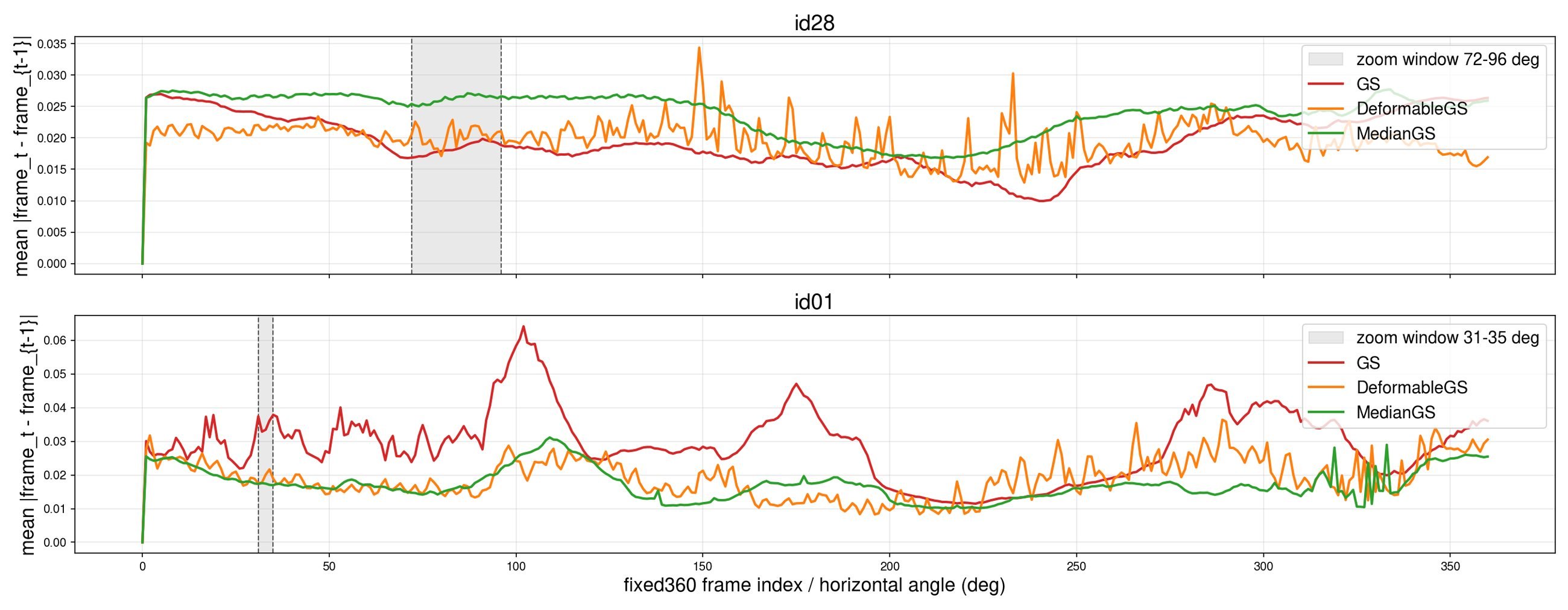}
\caption{\taskmedapp{Temporal fluctuation diagnostic for first-stage reconstruction variants. The curves are not absolute quality scores because a camera orbit naturally changes pixels; instead, they identify spike-like redraws and boundary flicker under the same canonical camera trajectory.}}
\label{fig:first-reconstruction-temporal}
\end{figure}

\paragraph{Temporal fluctuation diagnostic.}
The next diagnostic measures adjacent-frame fluctuation along the same canonical camera path. A raw adjacent-pixel curve is not an absolute visual-quality score, because normal orbit motion changes pixels even for a perfect reconstruction. It is used here as a controlled same-trajectory indicator of sudden redraws, boundary flicker, or frame-indexed shape changes. The shaded windows in the curve correspond to the zoomed crop figures that follow.

\taskmedapp{Figure~\ref{fig:first-reconstruction-temporal} localizes where the first-stage variants produce spike-like changes. For the street-car prompt, the later part of the orbit exposes unstable car-ground boundaries and background redraw. For the rabbit prompt, the diagnostic highlights a short region where small changes around the eye, ear, and background boundary become visually noticeable. The crop figures below make these artifacts explicit rather than relying only on a curve.}

\begin{figure}[t]
\centering
\includegraphics[width=\linewidth]{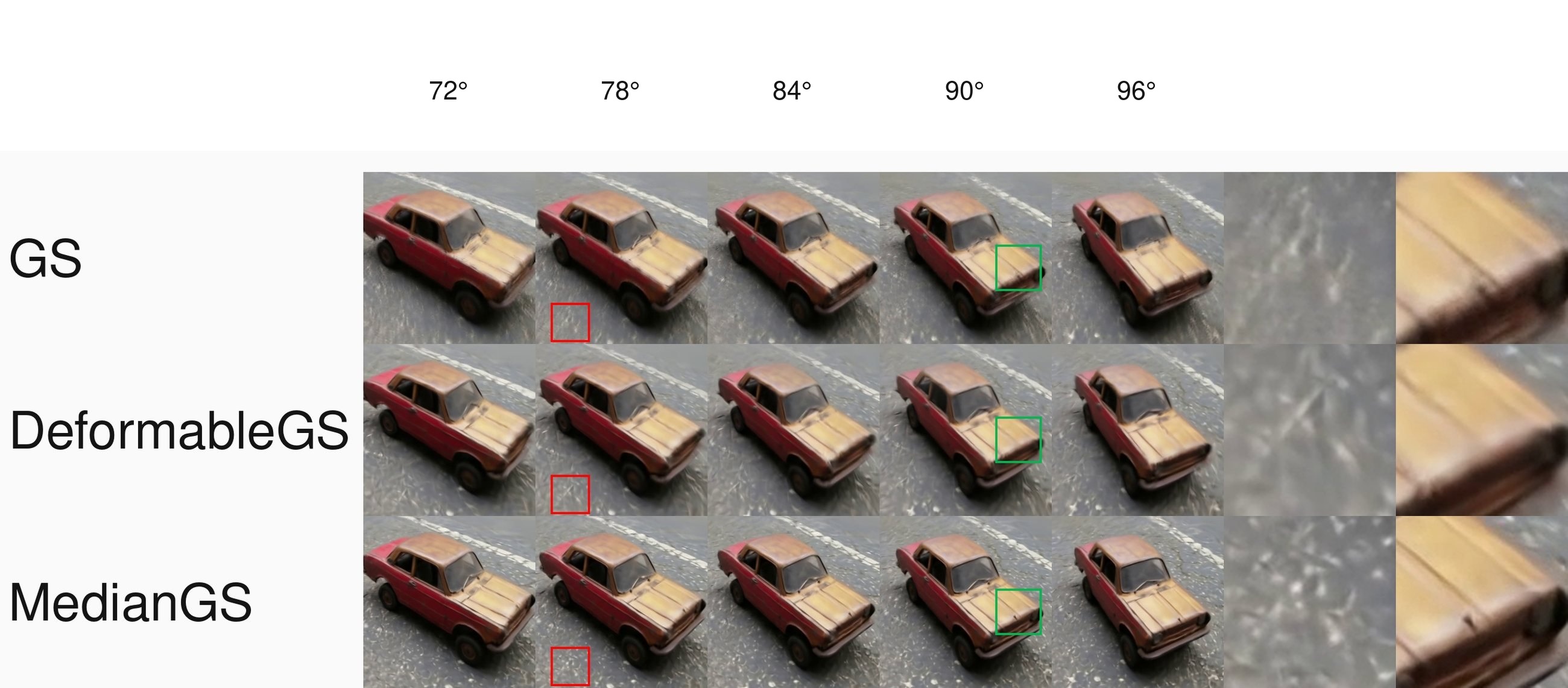}
\caption{\taskmedapp{Zoomed temporal crops for the street-car fluctuation window. The crops localize the boundary and texture redraws that produce spikes in Figure~\ref{fig:first-reconstruction-temporal}; MedianGS reduces the worst pseudo-dynamic artifacts while keeping the object attached to the support surface.}}
\label{fig:first-reconstruction-street-crops}
\end{figure}

\taskmedapp{In the street-car crop, the static control introduces unstable texture around the car-ground contact region, while the frame-indexed deformable render changes local shape as the orbit advances. MedianGS is not artifact-free, but it gives a more stable static condition because the render is no longer tied to a particular source-frame time index.}

\begin{figure}[t]
\centering
\includegraphics[width=\linewidth]{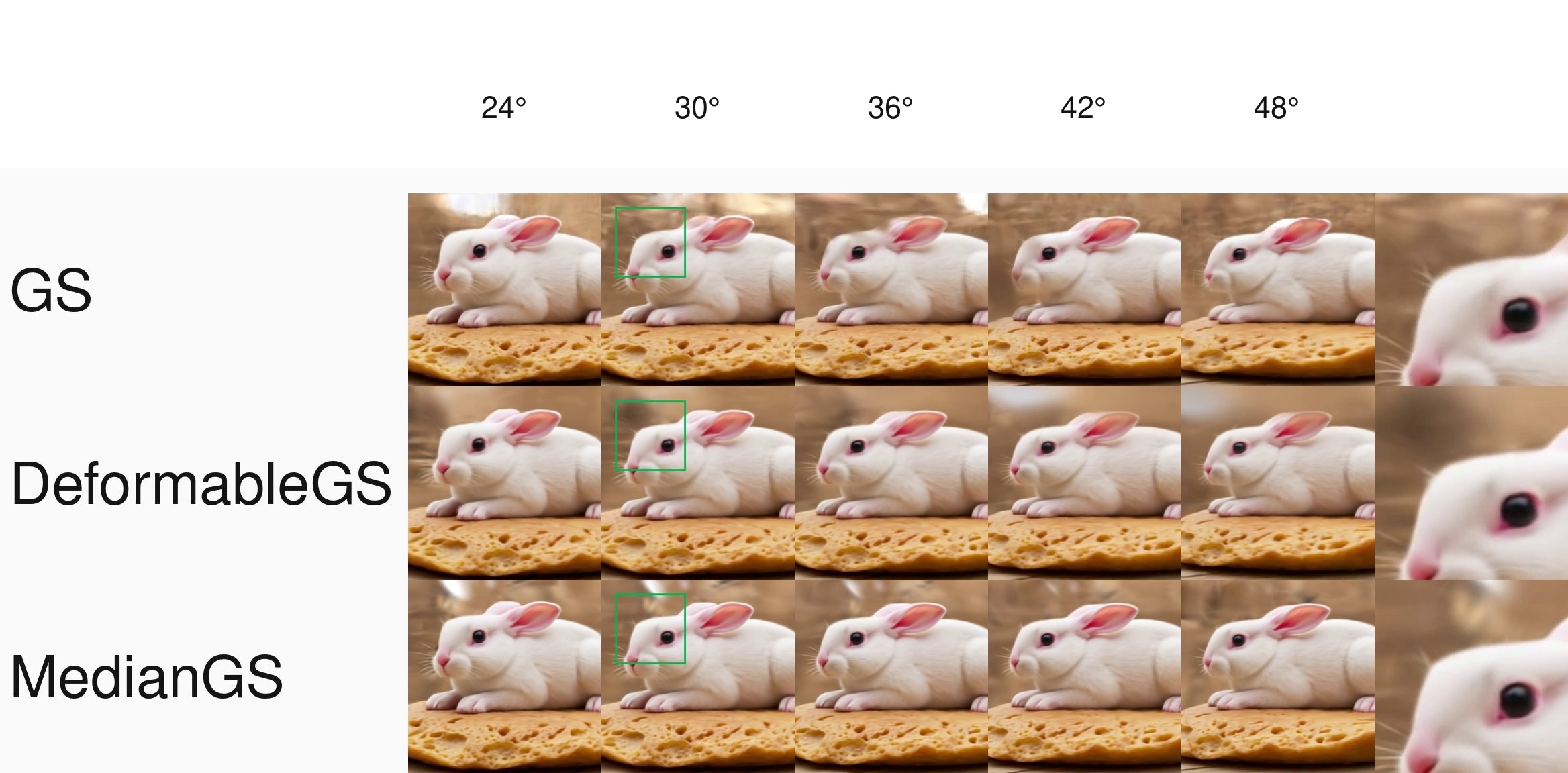}
\caption{\taskmedapp{Zoomed temporal crops for the rabbit-on-pancake fluctuation window. The comparison shows how frame-varying deformation can redraw local structure and background contact across nearby canonical views, while MedianGS produces a more stable static proxy for subsequent completion.}}
\label{fig:first-reconstruction-rabbit-crops}
\end{figure}

\taskmedapp{Figures~\ref{fig:first-reconstruction-street-crops} and~\ref{fig:first-reconstruction-rabbit-crops} show why the fluctuation diagnostic matters. In the rabbit crop, high-contrast eye and ear details are repeatedly redrawn by the direct controls. MedianGS again gives a more stable static condition, which is why the completion stage receives a static proxy rather than a frame-indexed deformable render.}

\begin{figure}[t]
\centering
\includegraphics[width=\linewidth]{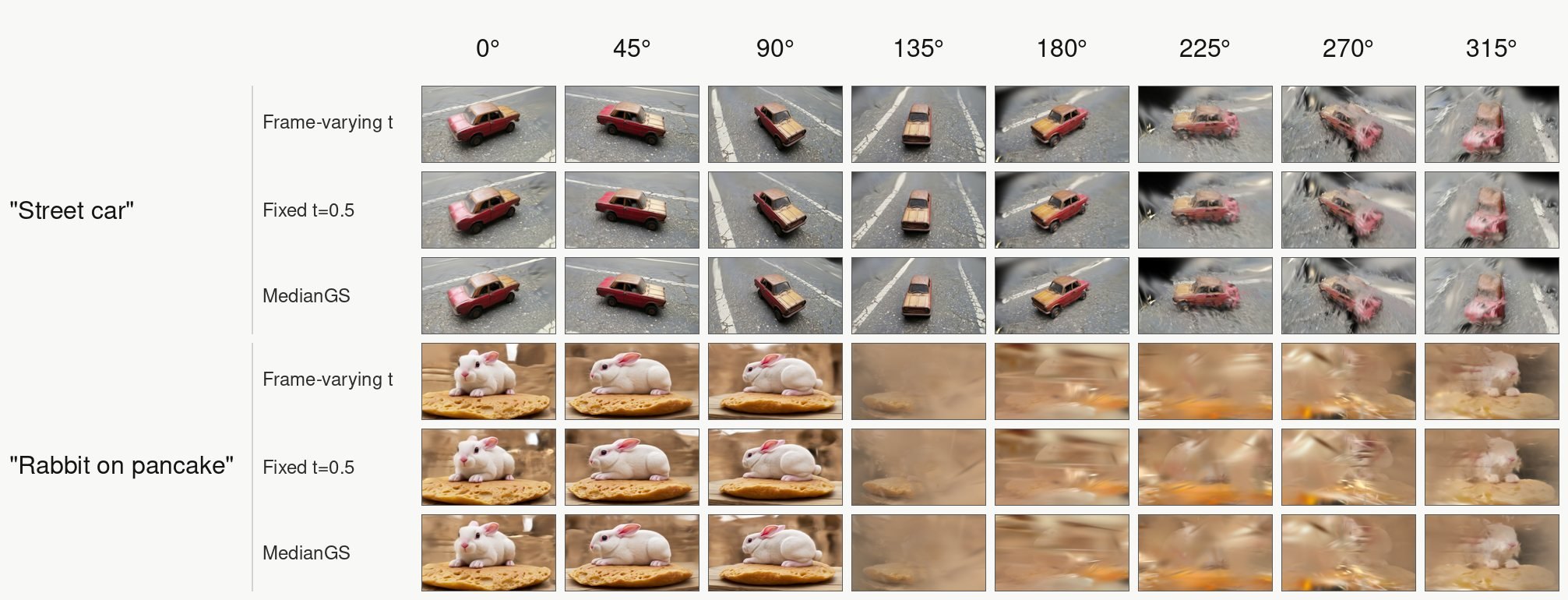}
\caption{\taskorbit{MedianGS static-proxy ablation on the same canonical 360-degree trajectory. Frame-varying deformation can preserve sharp observed views but carries pseudo-dynamic drift into the orbit; a fixed temporal slice is stable but arbitrary; MedianGS aggregates the shared deformation state through a robust temporal median.}}
\label{fig:mediangs-static-proxy-app}
\end{figure}

\paragraph{Staticization operator.}
We then isolate the staticization operator using the same DeformableGS checkpoint and the same canonical 360-degree trajectory. The three render-time variants are frame-varying deformation, a fixed temporal slice \(t=0.5\), and MedianGS. Frame-varying deformation can keep sharp details in views close to the source frames, but it also transports video-time deformation into a spatial orbit. The fixed temporal slice removes that time dependence but depends on an arbitrary choice of time. MedianGS uses all sampled deformation states and extracts the majority deformation pattern as a robust static proxy.

\begin{table}[t]
\centering
\small
\resizebox{\linewidth}{!}{%
\begin{tabular}{llccccccc}
\toprule
Prompt & Variant & Adj. L1 $\downarrow$ & L1 p95 $\downarrow$ & LPIPS p95 $\downarrow$ & DINO curv. p95 $\downarrow$ & Loop LPIPS $\downarrow$ & Loop DINO $\downarrow$ & Visual stability \\
\midrule
\emph{Street car} & Frame-varying $t$ & 6.39 & 7.76 & 0.061 & 0.467 & 0.230 & 0.057 & sharp but drifting \\
\emph{Street car} & Fixed $t=0.5$ & 5.20 & 5.97 & 0.045 & 0.365 & 0.167 & 0.032 & stable but arbitrary \\
\emph{Street car} & MedianGS & 6.09 & 6.91 & 0.048 & 0.356 & 0.168 & 0.027 & robust default \\
\emph{Rabbit on pancake} & Frame-varying $t$ & 4.91 & 7.34 & 0.087 & 0.828 & 0.303 & 0.055 & sharp but drifting \\
\emph{Rabbit on pancake} & Fixed $t=0.5$ & 4.53 & 7.06 & 0.069 & 0.629 & 0.217 & 0.023 & stable but arbitrary \\
\emph{Rabbit on pancake} & MedianGS & 4.30 & 6.58 & 0.060 & 0.562 & 0.206 & 0.027 & robust default \\
\bottomrule
\end{tabular}}
\caption{Static-proxy diagnostics over the full 361-frame canonical orbit. Frame-varying deformation can keep sharp observed views but carries pseudo-dynamic drift into the orbit; a fixed slice is stable but arbitrary; MedianGS uses all time samples through a robust coordinate-wise median. \taskrotabl{This ablation tests the staticization choice around the rotation-parameter aggregation; it should not be read as a comparison among intrinsic $\mathrm{SO}(3)$ rotation medians.}}
\label{tab:mediangs-app}
\end{table}

\taskrotabl{Table~\ref{tab:mediangs-app} summarizes the same comparison over the full 361-frame orbit. Adjacent L1 and L1 p95 measure pixel fluctuation; LPIPS p95 measures perceptual spikes; DINO curvature p95 measures abrupt second-order changes in semantic feature space; loop LPIPS and loop DINO compare near-loop cross-boundary pairs around \(0^\circ/360^\circ\), excluding the duplicated endpoint. The frame-varying render has the clearest drift pattern, especially on the rabbit prompt, where its LPIPS p95 and DINO curvature p95 are the largest. The fixed slice can reduce some pixel changes by suppressing motion more aggressively, but it is tied to an arbitrary source time. MedianGS stays in the stable perceptual cluster while retaining the all-sample interpretation, so we use it as the default first-stage proxy. This ablation isolates the practical staticization choice and should not be read as a claim that the component-wise rotation-parameter median is the unique optimal statistic on \(\mathrm{SO}(3)\). A dedicated chordal or geodesic rotation statistic would be an implementation replacement for Eq.~\eqref{eq:median}, not a change to the coverage-aware completion method.}

\FloatBarrier
\begin{figure}[t]
\centering
\includegraphics[width=\linewidth]{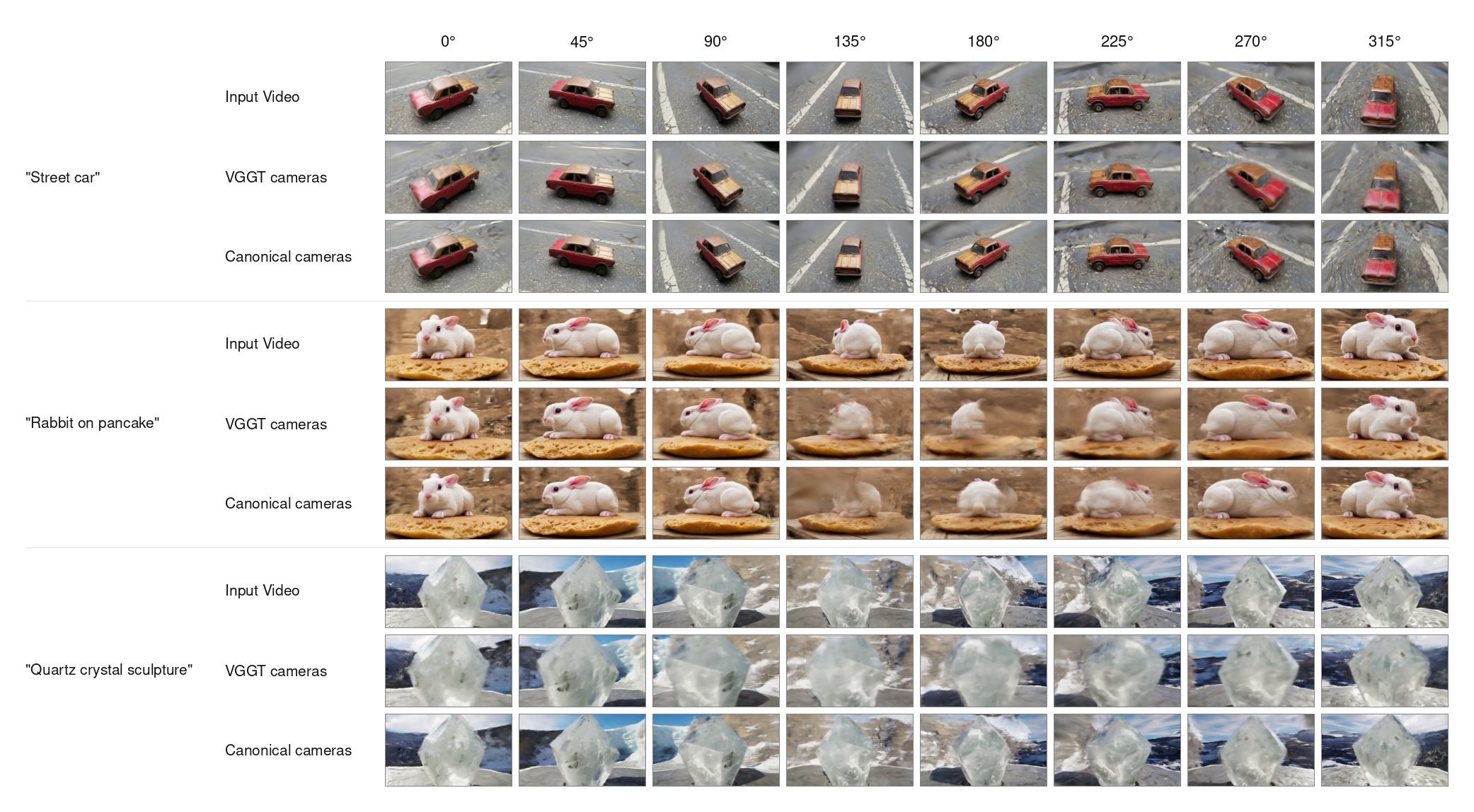}
\caption{\taskorbit{Canonical-camera versus re-estimated-camera second reconstruction. Both branches use the same completed frames and are evaluated on the same canonical cameras; inheriting the canonical cameras preserves the orbit coordinate system, while re-estimation can produce smoother adjacent motion but misaligned geometry.}}
\label{fig:canonical-camera-app}
\end{figure}

\section{Camera Continuity and Second Reconstruction Ablations}
\label{app:camera-continuity}

\taskorbit{This section tests whether the second reconstruction should keep the canonical cameras established before completion, or re-estimate a new camera system from the completed video. The question is important because the completed video may look temporally smooth even when its estimated cameras no longer agree with the canonical orbit used to define coverage. We therefore compare two branches under matched conditions: both use the same completed frames, the same reconstruction backend, the same training budget, and the same canonical evaluation cameras. The only changed variable is the camera set used during the second reconstruction. \ours trains with the prescribed canonical cameras, while the control re-estimates cameras from the completed video and aligns them back to the canonical coordinate system.}

\taskorbit{The purpose of this ablation is not to show that VGGT is unusable. VGGT is useful for estimating the initial source trajectory and constructing the canonical orbit, as detailed in Appendix~\ref{app:canonical-orbit}. The question here is narrower: after the missing interval has been completed in canonical order, should the method preserve that camera system or let a second camera-estimation pass redefine it? A lower adjacent-frame difference can be misleading in this setting, because an over-smoothed or misregistered reconstruction can be temporally smooth while failing to match the completed input frames under the intended cameras.}

\taskorbit{Figure~\ref{fig:canonical-camera-app} shows representative prompts from this controlled comparison. The re-estimated-camera branch can appear smoother locally, but it visibly drifts from the completed input sequence: the car body and road lose alignment, the rabbit and pancake smear in backside views, and faceted objects lose sharper structure. The canonical-camera branch better preserves the intended correspondence between frame index, orbit angle, and reconstructed content. This is why the main pipeline treats the canonical orbit as the coordinate system shared by completion and second reconstruction rather than as a visualization-only camera path.}

\begin{table}[t]
\centering
\small
\resizebox{\linewidth}{!}{%
\begin{tabular}{lrrrrrrrrr}
\toprule
Prompt & Canonical PSNR $\uparrow$ & VGGT PSNR $\uparrow$ & $\Delta$PSNR & Canonical SSIM $\uparrow$ & VGGT SSIM $\uparrow$ & Canonical MAE $\downarrow$ & VGGT MAE $\downarrow$ & Canonical Adj. L1 $\downarrow$ & VGGT Adj. L1 $\downarrow$ \\
\midrule
\emph{Street car} & 26.39 & 15.95 & +10.44 & 0.819 & 0.627 & 8.79 & 26.69 & 6.47 & 3.77 \\
\emph{Rabbit on pancake} & 20.40 & 15.89 & +4.52 & 0.801 & 0.666 & 15.77 & 28.39 & 5.48 & 5.32 \\
\emph{Quartz crystal sculpture} & 23.87 & 16.39 & +7.48 & 0.856 & 0.743 & 9.57 & 26.22 & 4.54 & 3.66 \\
\emph{Wooden chair} & 18.15 & 13.58 & +4.57 & 0.749 & 0.643 & 20.23 & 38.17 & 6.74 & 5.96 \\
\midrule
Mean & 22.20 & 15.45 & +6.75 & 0.806 & 0.669 & 13.59 & 29.87 & 5.81 & 4.68 \\
\bottomrule
\end{tabular}}
\caption{Fair canonical-camera versus aligned-VGGT-camera second reconstruction metrics over four prompts. PSNR, SSIM, and MAE compare each branch render against the same completed input video on the same canonical 360-degree evaluation cameras. Adjacent L1 is a smoothness diagnostic and should not be interpreted alone.}
\label{tab:canonical-camera-app}
\end{table}

\taskorbit{Table~\ref{tab:canonical-camera-app} quantifies the same effect on four prompts. The canonical-camera branch improves mean PSNR by \(6.75\) dB, improves mean SSIM from \(0.669\) to \(0.806\), and reduces mean MAE from \(29.87\) to \(13.59\) under the same canonical evaluation cameras. The re-estimated-camera branch has lower mean adjacent L1, but this is not evidence of a better reconstruction here; it reflects smoother renders that are less faithful to the completed canonical input. The ablation therefore supports keeping one camera system from canonical-orbit rendering through completion and second reconstruction.}

\section{Optional Cleanup and Outlier Filtering}
\label{app:optional-cleanup}

The optional cleanup branch can improve perceptual appearance by using $\Rone$ as a structural condition for a frozen video generator. It should remain visually useful but methodologically separate. Unless a downstream $V_1\rightarrow G_2$ reconstruction is performed and compared, cleanup should not be presented as the main 3D result. The main claim is the completed canonical-orbit Gaussian Splatting scene $\Gone$ and its render $\Rone$.

\taskcleanapp{The reported cleanup variants document condition-rendering choices only: no cleanup, pruning high-deviation Gaussian Splatting motion, pruning plus a statistical outlier filter, pruning plus a radius outlier filter, and the combined filter. These variants are not used to claim improvement of the primary $\Rone$ reconstruction.}

\begin{figure}[t]
\centering
\includegraphics[width=\linewidth]{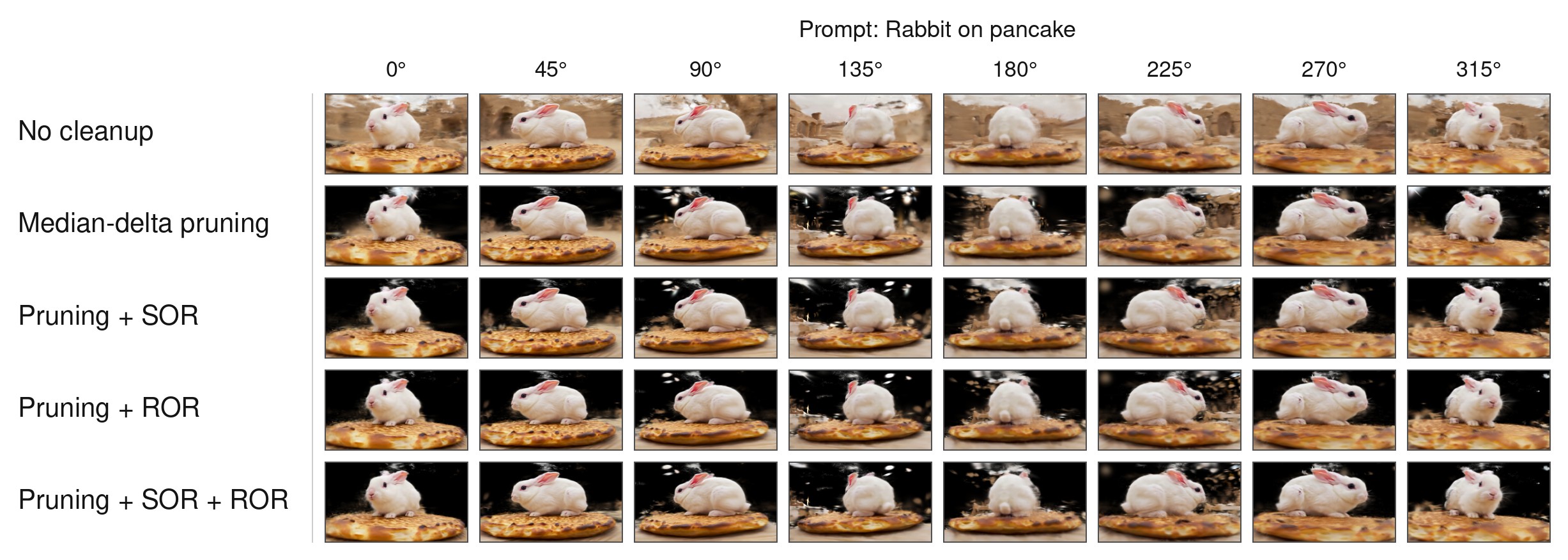}
\caption{\taskorbit{Optional Gaussian Splatting cleanup variants for condition rendering. The unfiltered render preserves the broadest context but can retain unstable background haze; pruning and spatial outlier filters suppress debris at the cost of removing some background support. These variants are condition-render candidates, not replacements for the reported reconstruction.}}
\label{fig:cleanup-app}
\end{figure}

\begin{figure}[t]
\centering
\includegraphics[width=\linewidth]{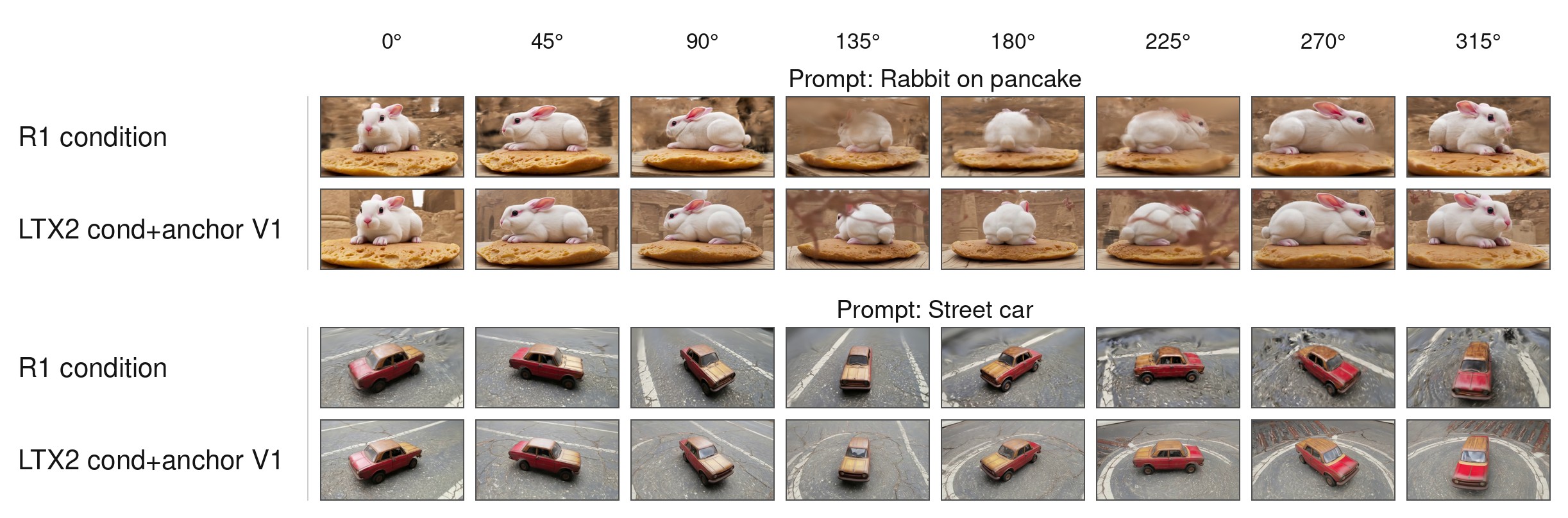}
\caption{\taskorbit{Optional condition-guided video refinement. The first row in each prompt block is the canonical-orbit condition render, and the second row is a visually refined video using the same orbit order and a first-frame appearance anchor. This is reported as visual refinement only; no downstream reconstruction from the refined video is claimed.}}
\label{fig:refinement-app}
\end{figure}

\paragraph{Condition-guided visual refinement.}
The second optional ablation asks what happens if the completed canonical render is used as a full-video structural condition for a frozen video generator. The input condition is the complete $\Rone$ sequence in canonical order, and the appearance anchor is the first source-video frame. This stage does not estimate cameras and does not change the canonical orbit metadata; it only produces a visually refined video aligned to the same frame order. Because no downstream reconstruction from the refined video is evaluated here, the output is treated as an auxiliary visualization rather than as the main 3D result.

\taskcleanapp{Figure~\ref{fig:refinement-app} shows the intended use of this refinement. On the rabbit prompt, the video prior can reduce render blur and smeared background artifacts while preserving the rabbit--pancake layout. On the street-car prompt, it can replace hazy road texture with a cleaner local appearance while following the same orbit order. The tradeoff is that the video generator may repaint local texture or background details according to its prior. This is why the paper reports the refined video as optional visual cleanup only; without a reconstructed and evaluated \(V_1\rightarrow G_2\) branch, it is outside the main 3D claim.}

\FloatBarrier

\section{Implementation Details}
\label{app:implementation-details}

\subsection{\taskimpl{Generative Backbones and Sampling}}

\taskimpl{The source video prior is HunyuanVideo-1.5 with the \texttt{480p\_t2v\_distilled} transformer. Unless otherwise stated, each prompt uses one source video sample with seed 42, a 121-frame 16:9 video at the model's 480p setting, and no manual best-of-many clip selection. The orbit-completion prior is the LTX-2.3 keyframe-interpolation pipeline using \texttt{ltx-2.3-22b-dev} with \texttt{ltx-2.3-22b-distilled-lora-384} and the Gemma-3-12B-it QAT Q4\_0 text encoder. Completion uses a 361-frame endpoint-window sequence at $896\times512$ orbit resolution, 25 fps, 12 denoising steps, and a negative prompt that suppresses blur, flicker, warping, geometry distortion, unstable cameras, texture repainting, deformation, and inconsistent perspective. Optional condition-guided visual refinement, which is not included in the main 3D result, uses the separate \texttt{ltx-2.3-22b-distilled} checkpoint with \texttt{ltx-2.3-22b-ic-lora-union-control-ref0.5}.}

\subsection{\taskimpl{Camera Estimation and Canonical-Orbit Constants}}

\taskimpl{VGGT estimates source cameras from the initial video. The camera fit uses the common successful source-camera set, the ray-intersection look-at center in Eq.~\eqref{eq:lookat-fit}, the SVD orbit plane, and the 2D circle fit in Eq.~\eqref{eq:circle-fit}. The canonical orbit contains $N=361$ views over $360^\circ$, where the final view duplicates the first azimuth for loop diagnostics. The render grid is $896\times512$, the orbit fps is 24, and the default elevation offset is $10^\circ$. Intrinsics use the median source focal length after scaling to the canonical render resolution.}

\subsection{\taskimpl{Gaussian Splatting Reconstruction Settings}}

\taskimpl{Both reconstruction stages use the same DeformableGS backend and the same training budget. The first stage fits the source video using estimated source cameras for 10{,}000 iterations. The second stage fits the assembled completed orbit video using the prescribed canonical cameras for 10{,}000 iterations. In controlled ablations, the initialization, completed frames, Gaussian Splatting backend, training budget, and evaluation cameras are held fixed; only the stated variable, such as second-stage camera source, is changed.}

\subsection{\taskimpl{MedianGS Staticization Settings}}

\taskimpl{MedianGS evaluates the deformation field at $T_m=128$ uniformly spaced time codes in $[0,1]$. It aggregates only geometry offsets: position, rotation-parameter offset, and scale offset. Opacity and spherical-harmonic appearance coefficients are copied from the canonical Gaussian Splatting state. The stored rotation is normalized after adding the aggregated rotation-parameter offset, and the static scale is converted back to log scale after preserving the magnitude of the activated scale.}

\subsection{\taskmedian{Rotation Aggregation in MedianGS}}

\taskmedian{The DeformableGS backend represents rotation changes as additive offsets in the renderer's rotation-parameter space. MedianGS therefore aggregates $\Delta q(\tau)$ component-wise before adding it to the canonical rotation parameter and normalizing the result. We do not interpret this operation as a Riemannian median on $\mathrm{SO}(3)$, and none of the paper's claims require that interpretation. Its role is a robust staticization heuristic: suppress frame-specific deformation offsets while preserving a single valid render state for the canonical orbit. A future implementation could substitute a hemisphere-aligned chordal mean, chordal median, or geodesic statistic at this point without changing the coverage-aware completion protocol.}

\subsection{\taskimpl{Coverage-Completion Settings}}

\taskimpl{The completion mask is a view-index mask rather than an image-space mask. Source-camera azimuths define exact observed bins $S_{\mathrm{obs}}$, the wrapped source-covered interval defines $S_{\mathrm{arc}}$, and only $U_{\mathrm{gap}}$ is synthesized. Sparse source anchors are selected from the VGGT-mapped source-to-canonical correspondence at stride 24. The source video has 121 frames, but the canonical completion sequence has 361 frame slots, one for each canonical camera. The endpoint-window sequence is arranged in synthetic order as known endpoint window, unsupported gap, known endpoint window; LTX-2.3 generates the 361-frame sequence, whose length already satisfies the model's $8m+1$ frame-count constraint. The output is then restored from synthetic order to canonical order. In the assembled training video for the second reconstruction, frames in $S_{\mathrm{arc}}$ are copied from the MedianGS render, and generated frames are used only for $U_{\mathrm{gap}}$.}

\subsection{\taskimpl{Metrics and Compute Accounting}}

\taskimpl{All automatic metrics use the same frozen 300-prompt \tthreebench{}-derived manifest. Coverage uses the prescribed or measured camera trajectory. Lower-tail quality samples 12 uniformly spaced views per prompt and scores them with ImageReward-v1.0; the reported Q10 is the per-prompt 10th percentile over sampled views averaged across prompts. Adjacent LPIPS and adjacent DINO are used only as smoothness diagnostics. ViewCrafter's native local-view trajectory is audited separately before its quality scores are interpreted.}
\taskconst{Table~\ref{tab:implementation-settings} reports the fixed model calls, frame counts, render grids, and optimization budgets that define the reproducible protocol. Table~\ref{tab:runtime-accounting} additionally reports measured wall-clock diagnostics from completed runs. These timings exclude queue wait and downstream metric computation, and should be read as practical compute accounting under shared GPU execution rather than as a controlled throughput benchmark, because GPU type, storage latency, and batching vary across the audit.}

\begin{table}[t]
\centering
\small
\resizebox{\linewidth}{!}{%
\begin{tabular}{lccp{6.3cm}}
\toprule
Component & Setting & Value & Notes \\
\midrule
Source video prior & HunyuanVideo-1.5, \texttt{480p\_t2v\_distilled} & 121 frames, 480p, 16:9 & Single sample per prompt under the frozen benchmark protocol. \\
Camera estimation & VGGT source cameras & successful-camera set & Used for orbit fitting and source-to-canonical correspondence. \\
Canonical orbit & prescribed render grid & 361 views, $896\times512$, 24 fps & Last view duplicates the first azimuth for loop diagnostics. \\
First reconstruction & DeformableGS & 10{,}000 iterations & Fits source video with estimated cameras. \\
MedianGS & deformation samples & $T_m=128$ & Aggregates geometry offsets only. \\
Orbit completion & LTX-2.3 keyframe interpolation, \texttt{22b-dev} + \texttt{distilled-lora-384} & 361 frames, stride-24 anchors, 12 steps & Synthesizes only the unsupported canonical interval. \\
Second reconstruction & DeformableGS & 10{,}000 iterations & Fits the assembled completed orbit with canonical cameras. \\
Quality metric & ImageReward-v1.0 & 12 views per prompt & Q10 is computed per prompt, then averaged over prompts. \\
Runtime diagnostics & wall-clock logs & see Table~\ref{tab:runtime-accounting} & Excludes queue wait and metric post-processing. \\
\bottomrule
\end{tabular}}
\caption{\taskconst{Implementation and compute-accounting constants used for the frozen 300-prompt \tthreebench{}-derived audit. The fixed settings define the reproducible protocol; measured wall-clock diagnostics are reported separately in Table~\ref{tab:runtime-accounting}.}}
\label{tab:implementation-settings}
\end{table}

\begin{table}[t]
\centering
\small
\resizebox{\linewidth}{!}{%
\begin{tabular}{lp{3.5cm}ccp{5.4cm}}
\toprule
Stage & Measurement set & Median time & Range or IQR & Notes \\
\midrule
Source text-to-video sampling & 19 representative prompt logs on H100-class nodes & 14.5 min & 11.8--14.8 min range & 121-frame 480p source video, one sample per prompt. \\
VGGT camera estimation & Same 19 representative prompt logs & 14.0 s & 11.2--20.5 s range & Source-camera inference only; later orbit fitting is negligible relative to reconstruction. \\
First DeformableGS reconstruction & Same 19 representative prompt logs & 4.9 min & 3.2--6.7 min range & 10{,}000 iterations on the source video with estimated cameras. \\
MedianGS canonical render & Same 19 representative prompt logs & $<0.2$ min & $<0.2$ min & Rendering and deformation aggregation are small compared with video sampling and Gaussian optimization. \\
Endpoint-window orbit completion & Same 19 representative prompt logs & 1.0 min & 0.8--1.1 min range & 361-frame $896\times512$ endpoint-window completion with sparse anchors. \\
Second DeformableGS reconstruction & Same 19 representative prompt logs & 8.3 min & 5.2--13.3 min range & 10{,}000 iterations on the completed canonical-orbit video. \\
Second DeformableGS reconstruction & 100 completed audit runs with stored timing summaries & 14.5 min & 11.0--19.6 min IQR & Same stage measured across the larger audit pool under mixed shared-GPU execution. \\
\bottomrule
\end{tabular}}
\caption{\taskconst{Measured wall-clock compute diagnostics. Times exclude scheduler queue wait and downstream metric computation. The first six rows summarize per-stage logs from representative complete runs; the last row reports the broader timing distribution available for the second reconstruction in the frozen \tthreebench{}-derived audit.}}
\label{tab:runtime-accounting}
\end{table}

\taskclaim{The claims in this paper are restricted to artifacts evaluated in the tables above. The optional cleanup branch and visual refinement figures are presented as auxiliary visual diagnostics, and local-view methods are not ranked in the coverage-qualified full-orbit quality table unless they satisfy the same orbit protocol.}

\FloatBarrier

\section{Additional Related-Work Boundary}
\label{app:related-work-boundary}

\taskrelwork{Table~\ref{tab:video-prior-related-boundary} separates \ours{} from the closest video-prior 3D and novel-view systems. The distinction is not whether a method uses video diffusion, but where the camera system and 3D representation enter the pipeline. \ours{} treats the generic text-generated video as an uncalibrated source observation, then builds a canonical orbit through reconstruction, camera fitting, and coverage-aware completion.} \taskboundary{We therefore discuss ViVid-1-to-3, V3D, Generative Gaussian Splatting, and G4Splat as closely related methods, but do not insert them into the frozen 300-prompt \tthreebench{}-derived audit: doing so would require adding image-conditioning, sparse-view, or model-training assumptions that are outside the text-video-to-closed-orbit protocol evaluated in the main tables.}

\begin{table}[t]
\centering
\scriptsize
\resizebox{\linewidth}{!}{%
\begin{tabular}{lp{2.4cm}p{3.0cm}p{2.8cm}p{3.8cm}p{4.3cm}}
\toprule
Method & Input & Use of video prior & Output protocol & Boundary relative to \ours & Why not ranked in the \tthreebench{}-derived audit \\
\midrule
ViVid-1-to-3~\citep{kwak2023vivid} & Single image and target-view path & Generates a scanning video to support novel-view synthesis & Target novel views along a camera path & Uses video diffusion for NVS; does not reconstruct a closed canonical-orbit Gaussian Splatting scene from a generic text-generated video. & \taskboundary{Image-conditioned NVS would require an extra image-anchor protocol, changing the input and error sources.} \\
V3D~\citep{chen2024v3d} & Single image & Fine-tunes video diffusion to synthesize 360-degree object-orbit frames & Object-centered orbit and reconstructed 3D asset & Specializes the video model for orbit generation; \ours{} keeps the video prior frozen and constructs the orbit after generation. & \taskboundary{Image-conditioned and fine-tuned orbit generation is outside the frozen generic text-to-video adapter setting.} \\
Generative Gaussian Splatting~\citep{schwarz2025ggs} & Dataset-trained generative setting & Integrates Gaussian Splatting feature fields into a latent video diffusion model & Generative 3D scene representation or multi-view images & Learns a model-internal 3D generative representation; \ours{} is an adapter around frozen video and reconstruction modules. & \taskboundary{Requires a trained generative 3D model rather than a per-prompt reconstruction adapter built from frozen priors.} \\
G4Splat~\citep{ni2025g4splat} & Sparse views or unposed video & Uses generative completion guided by explicit geometry and visibility reasoning & Geometry-guided scene reconstruction and completion & Starts from reconstruction evidence and geometry priors; \ours{} starts from a text-generated video and uses the first reconstruction primarily to define coordinates and coverage. & \taskboundary{Targets sparse-view or unposed-video reconstruction, not text-generated-video closed-orbit generation.} \\
\ours{} & Text-generated video & Uses frozen text-to-video and completion priors outside the reconstruction model & Scene-level closed-orbit Gaussian Splatting reconstruction & Converts an irregular generated-video trajectory into a canonical orbit without task-specific multiview fine-tuning or per-prompt SDS optimization. & \taskboundary{This is the protocol evaluated in the frozen 300-prompt \tthreebench{}-derived audit.} \\
\bottomrule
\end{tabular}}
\caption{\taskrelwork{Boundary to nearby video-prior 3D work. The comparison clarifies that \ours{} is not a new video diffusion generator, a fine-tuned orbit generator, or a geometry-guided sparse-view reconstructor; it is a reconstruction--completion adapter for generic text-generated videos.}}
\label{tab:video-prior-related-boundary}
\end{table}

\FloatBarrier
\newpage

\end{document}